\documentclass[]{article}

\usepackage[margin = 1.2 in]{geometry}
\usepackage{amsmath}
\usepackage{amssymb}
\usepackage[hidelinks]{hyperref}
\usepackage{graphicx}
\usepackage[dvipsnames]{xcolor}
\usepackage{float}
\usepackage{caption}
\usepackage{subcaption}
\usepackage[round]{natbib}
\usepackage{multicol}
\usepackage{array}
\usepackage[most]{tcolorbox}
\usepackage{booktabs}
\usepackage[framemethod=TikZ]{mdframed}
\usepackage{babel}
\usepackage[T1]{fontenc}
\usepackage{tabularx}
\usepackage{longtable}
\usepackage{enumitem}

\begin{document}

\title{Using Large Language Models for Qualitative Analysis can Introduce Serious Bias\thanks{The authors are grateful to the World Bank's Knowledge for Change Program, and the World Bank-UNHCR Joint Data Center on Forced Displacement for financial support. Sudarshan Aittreya provided valuable research assistance for the project.}}

\author{Julian Ashwin \\ Maastricht University \and Aditya Chhabra  \\ World Bank\and Vijayendra Rao\thanks{Corresponding author: vrao@worldbank.org} \\ World Bank}


\date{\today}

\maketitle

\begin{abstract}
Large Language Models (LLMs) are quickly becoming ubiquitous, but the implications for social science research are not yet well understood. This paper asks whether LLMs can help us analyse large-N qualitative data from open-ended interviews, with an application to transcripts of interviews with Rohingya refugees in Cox's Bazaar, Bangladesh. We find that a great deal of caution is needed in using LLMs to annotate text as there is a risk of introducing biases that can lead to misleading inferences. We here mean bias in the technical sense, that the errors that LLMs make in annotating interview transcripts are not random with respect to the characteristics of the interview subjects. Training simpler supervised models on high-quality human annotations with flexible coding leads to less measurement error and bias than LLM annotations. Therefore, given that some high quality annotations are necessary in order to asses whether an LLM introduces bias, we argue that it is probably preferable to train a bespoke model on these annotations than it is to use an LLM for annotation.
\end{abstract}

\begin{center}
\end{center}


\section{Introduction}

Large Language Models (LLMs) are increasingly being used in social science research to, among other things, analyze and annotate text data \citep{gilardi2023chatgpt}. As LLMs become more accessible and popular we can expect that there will be a temptation to use them to analyze open-ended interview data such as those used by qualitative researchers \citep{smallcalarco} who follow an interpretative  analytical approach. This relies on careful, nuanced, coding conducted by trained social scientists \citep{DeteringWatersFlexiblecoding}. Qualitative analysis of this kind lies at the core of fields like anthropology and sociology, and there is now a rapidly expanding literature on the use of Natural Language Processing (NLP) methods to analyze qualitative data in sociology \citep{bonikowski2022ends}, and qualitative analysis and NLP are also now being increasingly employed in more quantitative fields such as economics \citep{rao2022can}.

Data generated from open-ended, in-depth, interviews is potentially very different from the benchmark datasets often used in the NLP literature to validate modelling approaches such as English language tweets and news, or product reviews. This is because qualitative research is often conducted in a manner where the specific context matters in interpreting the data, and analyzed with codes that are "flexibly" developed that can be quite nuanced and complex. This is a particular problem in non-Western societies because LLMs have been shown to most resemble people from Western, Educated, Industrialized, Rich and Democratic (WEIRD) societies \citep{atari2023humans}, and our example application falls into this category. We have interviews on a very specific topic (children's aspirations) with a very specific population (Rohingya refugees and their hosts in Bangladesh) who are not well represented in the training data that LLMs are trained on (or in the data used in the NLP literature more broadly). 

We find that in such a context, using LLMs to annotate text is potentially dangerous. We test three different LLMs (ChatGPT and two versions of Meta's Llama 2) and find that the prediction errors they make in annotation are not random with respect to the characteristics on the interview subject. This can lead to misleading conclusions in later analysis, as we shown in Figure \ref{fig:coefs_ref_sex_small}. Statistical analysis based on LLM annotations can lead to estimated effects that are very different from those based on human expert annotations. It is therefore crucial to have some high quality expert annotations, even if it is just to assess whether the LLM is introducing bias or not. Given that some high quality annotations are needed to assess whether the LLM introduces bias, we argue that it is preferable to train a bespoke model on these annotations than it is to use an LLM.   
  
We show that iQual, a method we developed with others \citep{ashwin2022method} to analyze large-N qualitative data by training supervised models on small human annotated samples, not only performs better than LLMs in terms of out-of-sample prediction accuracy but also introduces much less bias. LLMs can possibly assist this process by generating larger training sets (i.e. data augmentation, as proposed by \cite{dai2023chataug}) but we only find evidence of marginal benefits in a few cases. This suggests a potential way in which to reconcile the nuance and "reflexive" qualities of interpretative qualitative analysis with large representative samples. Crucially, we see LLMs and other NLP methods as assisting and extending traditional qualitative analysis, not replacing it. In order to create a coding tree that captures important and interesting variation across documents in a nuanced and context-aware manner, there is no substitute for a careful reading on at least a subset of those documents.

Our application is based on open-ended interviews with Rohingya refugees and their Bangladeshi hosts in Cox's Bazaar, Bangladesh. These interviews focused on subjects aspirations and ambitions for their children \citep{callard2018aspiration} as well as their capacity to achieve those goals, i.e. their navigational capacity \citep{appadurai2004capacity}. They are analysed in detail in \cite{ashwin2022method}, so we will not discuss the detail of data collection or related social science literature here. The substance of these interviews is not critical to the methodological contribution of this paper, but it is important to note that while "ambition" can be captured well by structured questions that yield quantitative data, aspirations and navigational capacity are subtle and complex concepts not easily defined are captured in structured surveys. It is precisely when dealing with these sorts of concepts that open-ended interviews and interpretative qualitative analysis is valuable. The complexity and nuance of the concepts may play a role in explaining the poor performance of LLMs in annotating interviews compared with other studies where the annotation tasks were substantially more straightforward, e.g. \cite{mellon2022does}. 

Previous work has suggested that LLMs might outperform crowd-sourced human annotations \citep{gilardi2023chatgpt}, or even that a substantial proportion of workers on crowd-sourcing platform may be using LLMs in completing tasks \citep{veselovsky2023artificial}. Our results do not contradict these as for many annotation tasks LLMs may indeed perform very well and save researchers the expense and complication of crowd-sourcing. However, our results do suggest that researchers ought to be aware of the possibility of biases introduced by LLM annotation, particularly on data where a nuanced, contextual understanding of the documents is needed; LLMs, like other types of machine learning models, reflect the data they are trained on \citep{kearnsroth} and many of the contexts in which qualitative analysis adds value require an understanding of communities and concepts that may not be adequately represented in this training data. 

The paper is structured as follows. The remainder of this Section discusses this paper's contribution in the context of related literature. Section \ref{sec:data} then very briefly introduces our dataset of annotated interview transcripts. Section \ref{sec:methods} describes our approach to using LLMs for annotation (\ref{subsec:llm_class}) and the supervised NLP method introduced by \cite{ashwin2022method} which we refer to as iQual going forward (\ref{subsec:iqual}). Section \ref{sec:results} then describes LLM-based out-of-sample performance in comparison to iQual (\ref{subsec:results_perf}) and then shows that LLMs introduce more bias and illustrates this could cause researchers to draw incorrect conclusions (\ref{subsec:results_bias}). Section \ref{sec:discussion} then concludes.

\section{Data and Qualitative Analysis} \label{sec:data}

The interview transcripts, data collection and the qualitative coding process are explained in detail in \cite{ashwin2022method}, so we restrict ourselves to a very brief description here. The population we sample are Rohingya refugees based in the Cox's Bazzar camp and local Bangladeshi residents.  Along with a standard household survey including questions on demographics and economic conditions, the data include transcripts of 2,407 open-ended interviews with subjects on their aspirations for their eldest child. The interviews were conducted either in Bengali or in Rohingya which was then transcribed into Bengali, but we work with machine translations into English. The interviews take the form of an unstructured to-and-fro of question and answer (QA) pairs the interviewer and the subject. The interviews are on average 12.6 QA pairs long, with the average answer in each QA pair being 13.7 words long.

Based on a close reading of a subset of transcripts, and following a "flexible coding" process \citep{DeteringWatersFlexiblecoding}, a coding tree was developed including 25 potentially overlapping categories, 19 of which we focus on in this paper. A full description of each code along with examples are shown in Appendix \ref{app:prompts_classification}. Following \cite{callard2018aspiration} the distinction between aspiration and ambition was adapted within the context and nature of "dreams" parents expressed for their children. For example, concrete and measurable dreams for child (e.g wishing a child would become a doctor, teacher, entrepreneur, or specific educational goals) was used as a definition for ambition while intangible, value oriented goals (e.g wishing the child to live with dignity or be a good human being) was classified as aspiration. Aspirations, were divided into "Religious" and "Secular" . Ambition was divided into five major categories – Education (further sub-coded into High, Low, Neutral and Religious), Job Secular, Marriage, Entrepreneurship, Migration, Vocational Training, and No Ambition. While ambition and aspiration came up at any point in an interview,  "capacity to aspire" or Navigational capacity was restricted to discussions of what have parents were planning or able to do to fulfill dreams for their children.  Navigational Capacity was coded into seven sub-codes – Low and High "Ability", Low and High "Budget", Low and High "Information Awareness", and Reliance on God.

Of our sample of 2,407 interview transcripts, 789 are manually annotated by trained sociologists (co-authors on the \cite{ashwin2022method} paper) according to this coding structure. The annotations are defined at the level of QA pairs, allowing us to represent each annotation as a binary classification problem at the QA level.

\begin{figure}[!ht]
    \caption{LLM instructions example}
    \label{fig:instruct_example}
    \centering
    \vspace*{-0.5cm}
\begin{mdframed}[roundcorner=5pt, backgroundcolor=black!5]
    \scriptsize
    \textit{\textcolor{ForestGreen}{Your task involves interpreting and analyzing transcripts of interviews conducted with parents, focusing on their aspirations concerning their children's future education and careers solely in the context of religion, without implying any monetary incentives. Specifically, your role is to discern whether the parent explicitly expresses a desire for their child to pursue a religious path.
    \\[-1.8ex]\\
    Assign the “Religious Aspirations” label only if the parent explicitly articulates aspirations for their child that include becoming a religious scholar, Hafiz, attending a madrassa, learning Arabic, Quran reading, Islamic covering, regular prayer, working in Islamic banks, or being a  Islamic scholar. It is crucial not to assign this label if the child is already engaged in any of the mentioned activities, as current religious activities are not considered future aspirations. If there is no such direct reference to religious aspirations in the parent's statement, assign the “Not Applicable” label to indicate that the topic of religious aspirations was not addressed.}
    \\
    \textcolor{MidnightBlue}{Here are some examples: }
    \\
    \textcolor{MidnightBlue}{\#\#\# Input: \\
    Interviewer: What are your aspirations for your son's future?
    Respondent: I hope that he will become a great Islamic scholar. I want him to learn Arabic and read the Quran regularly.} \\
    \textcolor{BrickRed}{\#\#\# Output: \\
    Interpretation: The parent explicitly expresses a desire for their child to pursue a path in religious education, specifically mentioning becoming an Islamic scholar, learning Arabic, and reading the Quran regularly. Therefore, this conversation aligns with the `Religious Aspirations' label. Label: Religious Aspirations }
    \\
    \textcolor{MidnightBlue}{\#\#\# Input: \\
    Interviewer: Do you have any specific dreams for your daughter?
    Respondent: I want her to excel in her studies and become a successful engineer. }\\
    \textcolor{BrickRed}{\#\#\# Output: \\
    Interpretation: While the parent does express aspirations for their child's future, these aspirations are not religious in nature. The parent hopes for their child to become an engineer, which doesn't involve any religious activities as per the instruction. Therefore, this conversation should be labeled as 'Not Applicable'. Label: Not Applicable}
    \\
    \textcolor{MidnightBlue}{\#\#\# Input: \\
    Interviewer: What do you envision for your child’s future?
    Respondent: I would like my son to become a Hafiz. It’s a great honor in our religion.} \\
    \textcolor{BrickRed}{\#\#\# Output:  \\
    Interpretation: The parent explicitly expresses a desire for their child to become a Hafiz, which is a religious aspiration as per the instruction. Therefore, this conversation aligns with the `Religious Aspiration' label. Label: Religious Aspirations}
    \\
    \textcolor{MidnightBlue}{\#\#\# Input: \\
    Interviewer: What are your hopes for your children’s future careers?
    Respondent: I want them to have successful careers. I hope my daughter becomes a doctor and my son becomes a lawyer.} \\
    \textcolor{BrickRed}{\#\#\# Output:  \\
    Interpretation: While the parent does express aspirations for their children’s future, these aspirations do not involve any religious activities or pursuits as per the instruction. Therefore, this conversation should be labeled as `Not Applicable'. Label: Not Applicable} \\
    }
\end{mdframed}
\vspace*{-0.5cm}
\begin{mdframed}[roundcorner=5pt, backgroundcolor=black!5]
    \scriptsize
    \textit{\#\#\# Input: \\
Interviewer: What is your dream about your son who is studying in class one?
Respondent: Still in Noorani. After reading Noorani, he will read Hafezi. By the grace of Allah, he will teach in a madrasa when he becomes a hafez. That is my dream.}
\end{mdframed}
\vspace*{-0.5cm}
\begin{mdframed}[roundcorner=5pt, backgroundcolor=black!5]
    \scriptsize
    \textit{\#\#\# Output: (ChatGPT) \\
Interpretation: The respondent expresses a clear religious aspiration for their child, which is to become a Hafiz and teach in a madrasa. There is no mention of any monetary incentives, so this response aligns with the criteria for "Religious Aspiration". Label: Religious Aspiration}
\end{mdframed}
\vspace*{-0.5cm}
\begin{mdframed}[roundcorner=5pt, backgroundcolor=black!5]
    \scriptsize
    \textit{\#\#\# Output: (Llama-2-13b) \\
Interpretation: The parent explicitly expresses a desire for their child to pursue a path in religious education, specifically mentioning becoming a Hafiz and teaching in a madrassa. Therefore, this conversation aligns with the "Religious Aspiration" label. Label: Religious Aspiration}
\end{mdframed}
\vspace*{-0.5cm}
\begin{mdframed}[roundcorner=5pt, backgroundcolor=black!5]
    \scriptsize
    \textit{\#\#\# Output: (Llama-2-13b-chat) \\
Interpretation: The parent explicitly expresses a desire for their child to pursue religious studies and become a Hafiz (one who has memorized the Quran). They also mention teaching in a madrasa, which is an Islamic school. Therefore, this conversation aligns with the "Religious Aspiration" label. Label: Religious Aspiration }
\end{mdframed}
\vspace*{-0.6cm}
    \flushleft
    \caption*{\footnotesize \textit{Note}: This Figure shows example instructions given to the LLMs for the "Aspirations: Religious" code. These results are given to the LLM as a single prompt, but for clarity we have grouped the instructions into three parts here, as indicated by the colors. The first section (in green) describes the task and gives the definition of the code. The labelled examples (in blue) help the LLM understand the task at hand, often referred to as "few shot learning". For each example, we also provide a reasoning for the annotation (in red) so that the LLM is also asked to explain why it applies a certain label, which is known as "chain of thought" prompting. The second box shows an example of a QA pair to be annotated and the subsequent boxes show the responses of the three LLMs to this prompt.}
    \vspace*{-0.35cm}
\end{figure}

\section{Methods} \label{sec:methods}

In this Section we first explain how we use LLMs to annotated our interview transcripts. We then briefly describe the iQual method which trains supervised models on our expert human annotations, as well as how we use LLMs for data augmentation in combination with iQual. We test three different LLMs- the closed-source ChatGPT (gpt-3.5-turbo) by OpenAI, as well as two open-source LLMs by Meta, the Llama-2 (13b) and its fine-tuned "chat" variant. \citep{touvron2023llama}.The base Llama-2 is pretrained on publicly available online data sources. The chat variant is then fine-tuned on publicly available instruction datasets and over 1 million human annotations. This fine-tuning is designed to make the model align with human preferences using techniques such as Reinforcement Learning with Human Feedback (RLHF). ChatGPT is also fine tuned using RLHFFor all three LLMs, our approach to prompting remains consistent.

\subsection{Annotation with LLMs} \label{subsec:llm_class}

We follow several well-established practices to improve the effectiveness of LLMs in annotating our interview transcripts. We provide a prompt that includes precise directives for the LLM, and employ "few-shot learning" \citep{brown2020language} as well as "chain of thought" prompting \citep{wei2022chain}, as explained below. For each code, we created detailed textual instructions, similar to those one would give to human annotators. These instructions include enough details to ensure that, in principle, the model is fully aware of the specific standards and definitions required for coding transcripts. Each code and each question-answer pair are annotated by the LLM independently. By incorporating both few-shot learning and CoTP, we are in line with best practices and give the LLMs a good chance at annotating accurately. Previous work has shown that using these techniques can help LLM out-perform crowd workers in text annotation tasks \citep{gilardi2023chatgpt}.

The choice of prompt given to an LLM when giving it a certain task can make a substantial difference to its performance. We give the models a thorough briefing of what each code represents and how to identify its presence in a conversation. This includes the context, certain specific terms or activities that could be indicators, and the need to distinguish between current circumstances and future aspirations, as shown in the green text of Figure \ref{fig:instruct_example}. These instructions provide a benchmark for the model to understand the coding system and thereby infer the respective codes from the interview transcripts. Instructions for each of the codes are shown in Appendix \ref{app:prompts_classification}.

Few-shot learning and chain of thought prompting (CoTP) are two powerful techniques that can be combined to improve the performance and interpretability of LLMs. Few-shot learning provides examples of a task to the model, which helps guide its behavior and understanding of the task at hand. We provide the LLM with four examples that follow the detailed instructions, as shown by the blue text in Figure \ref{fig:instruct_example}, to demonstrate correct behavior to the model. These examples are chosen to be instructive of the how the code should be applied and are similar to the examples one would use to explain a code in traditional qualitative analysis.

We also apply chain of thought prompting (CoTP) in these examples to nudge the model to generate an interpretation of the transcript and articulate its line of reasoning before assignment of the final code. It is beneficial in complex tasks where reasoning and interpretation play crucial roles, such as our coding task. The underlying idea is that by having the model outline its thinking process, we can encourage it to reason more deeply and accurately, while also producing outputs that are more interpretable and trustworthy. For our task, we have used both few-shot learning and CoTP by asking the model not only to provide a label for each transcript, but also to give an interpretation explaining why it chose that label.

An example of a full prompt for the 'Religious Aspirations' code are shown in Figure \ref{fig:instruct_example}, with the instructions and few shot examples for all codes shown in Appendix \ref{app:prompts_classification}.

\subsection{Training supervised models on interpretative annotations (iQual)} \label{subsec:iqual}

An alternative to using LLMs to annotate large corpora of text documents would be to create high quality annotations on a smaller sub-sample and then training supervised models to predict these annotations on the remainder of the documents. We thus train a separate classifier for each code on a numerical representation of the text at the QA level. As discussed in Appendix \ref{app:iqual}, there are many options for both the classifier we can use here (e.g. random forest, logistic regression, neural networks, SVM), as well as how to represent the text numerically (e.g. tf-idf ngram vectors, sentence embeddings, translations or transliterations). Using k-fold cross-validation we select the best performing model, the text representation and a variety of hyperparameters, so that the approach which performs best in out-of-sample prediction is selected. In each case, we hold out a test set of 200 interviews in order to assess out-of-sample performance. Details about this methodology are provided in \cite{ashwin2022method}, and it is implementable in an open source Python package.\footnote{\href{https://github.com/worldbank/iQual}{https://github.com/worldbank/iQual}} The crucial intuition though is simply that we use a subset of high quality expert annotations to train a small bespoke model for each code. These models rely only on the annotated training data, unlike the pre-trained LLMs which are trained on huge quantities of text from a huge range of contexts.

Rather than asking LLMs to directly annotate text, another potential use for them is for data augmentation in combination with a supervised model, such as those described above. Data augmentation is a common technique in machine learning to generate more variation in a training set while preserving the important signals. For example, when training a model on a labelled dataset of images of animals, one might generate extra variation in the training data by rotating the images by 90 degrees or transforming them into a mirror image of themselves. The idea is to generate more training observations where the noise in the data is different but the signals are the same. A good example of this from the natural language processing literature is back-translation, where text is translated into a different language and then back into the original, so that the exact phrasing and style of the text is different but the meaning is the same \citep{edunov2018understanding}. Using LLMs for data augmentation has been found to increasing prediction performance in some contexts, so we follow the approach set out in \cite{dai2023chataug} as an additional experiment here. The example prompts and further details on the augmentation are shown in Appendix \ref{app:prompts_augmentation}.

We thus test two different versions of iQual: first, training supervised models on the human annotations without the use of LLMs, and second, training the model on data augmented by the LLMs to generate more variation in the text while preserving the meaning.

\section{Results} \label{sec:results}

We assess the performance of LLMs in our annotation tasks along two dimensions. Firstly, we assess how accurate of the LLMs predictions relative to our expert human annotations, finding that performance is poor relative to our simpler supervised models. Secondly, and more importantly, we investigate whether the annotations provided by LLMs or iQual introduce bias. We here mean bias in the technical sense that the prediction errors which the models make are not random. 

\begin{figure}[H]
    \centering
    \caption{Out-of-sample prediction performance of different methods}
    \label{fig:perf}
    \includegraphics[width =  \linewidth]{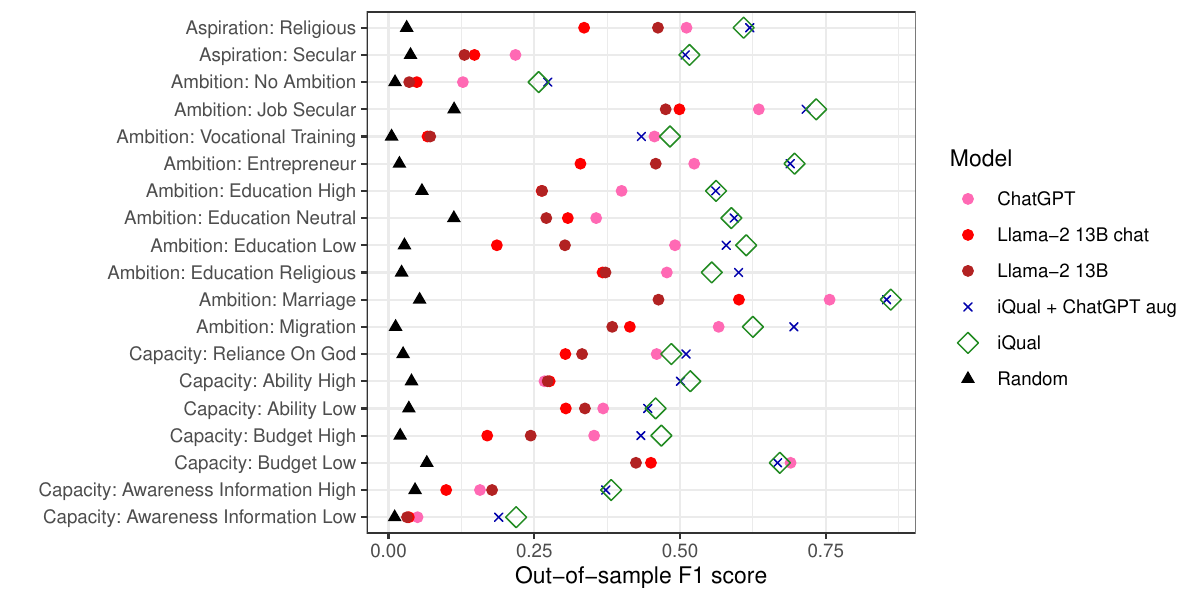}
    \flushleft
    \caption*{\footnotesize \textit{Note}: This Figure compares the out-of-sample prediction performance of LLM and supervised approaches, compared to the expert human annotations. Each code is shown along the vertical axis, and the test set F1 scores are shown on the horizontal axis. The F1 score that would be achieved by random guessing is shown as the black triangle and all models perform better than this. The performance of each model for each code is shown as a separate point with the color and shape of the point denoting the model. Averaging the F1 scores across all codes, iQual performs best with 0.542, followed by iQual + ChatGPT aug (0.541), ChatGPT (0.414), Llama-2 13B (0.290) and finally Llama-2B chat (0.274).}
\end{figure}

\subsection{Out-of-sample Performance} \label{subsec:results_perf}

Given that the interview transcripts are annotated with a series of binary variables at the QA level, we can assess LLM prediction accuracy with the out-of-sample F1 score for each code. We compare the performance of each LLM to the supervised models trained on annotated data, with and without augmentation.

Figure \ref{fig:perf} shows the results comparative performance of the different annotation approaches, as measured by the test set F1 score.\footnote{The F1 score is the harmonic mean of the precision and recall, where precision is the number of true positive divided by the sum of true positives and false positive, and recall is the number of true positive results divided by the sum of true positives and false negatives. It thus symmetrically represents both type 1 and type 2 errors.}  Given that many of the codes are very sparse, a useful comparison is the F1 score that random guessing would achieve, which is shown as black triangles. All models across all codes do better than random (i.e. have a higher F1 score than that which random guessing would achieve). In all but one case (Capacity: Awareness Information High) ChatGPT is the best performing LLM. However, in all but one case (Capacity: Budget Low) all LLMs perform worse than iQual in terms of these F1 scores. When used for augmentation, ChatGPT does improve performance slightly in some cases, but it slightly worsens performance just as often. If we measure performance in terms of accuracy (i.e. the proportion of observations that are correctly classified) rather than F1 score we get the same results. iQual achieves accuracy of 0.969. In contrast, ChatGPT only achieves 0.909, Llama-2 13B  0.854 abd Llama-2 13B chat 0.851. 

These results are of course specific to our context, and a different annotation structure on a different set of text data may lead to different results. However, in our case it is clear that LLMs generate less accurate annotations than training much smaller models on a subset of human annotations does.

\subsection{Bias} \label{subsec:results_bias}

If the annotations generated by LLMs are inaccurate, this is not necessarily a hugely consequential problem. If the mistakes they make are random, with a large enough sample we should still be able to come to correct conclusions. However, if the mistakes are not random, then using LLM annotations can lead to completely incorrect conclusions. In other words, if the LLMs errors are biased, then relying on these annotations could lead researchers to identify relationships in the data that are purely a result of these algorithmic biases rather than reality. 

We look at two ways in which the predicted annotations could be biased. Firstly, and most straightforwardly, we show that LLMs over-predict annotations that are very sparse (i.e. there are many more false positives than false negatives). Secondly, we show that in many cases LLM prediction errors are systematically associated with characteristics of the interview subject (e.g. refugee status, gender, education). 

\begin{figure}[H]
    \centering
    \caption{LLMs systematically over predict annotations}
    \label{fig:comp_bias_mean_alt}
    \includegraphics[width =  \linewidth]{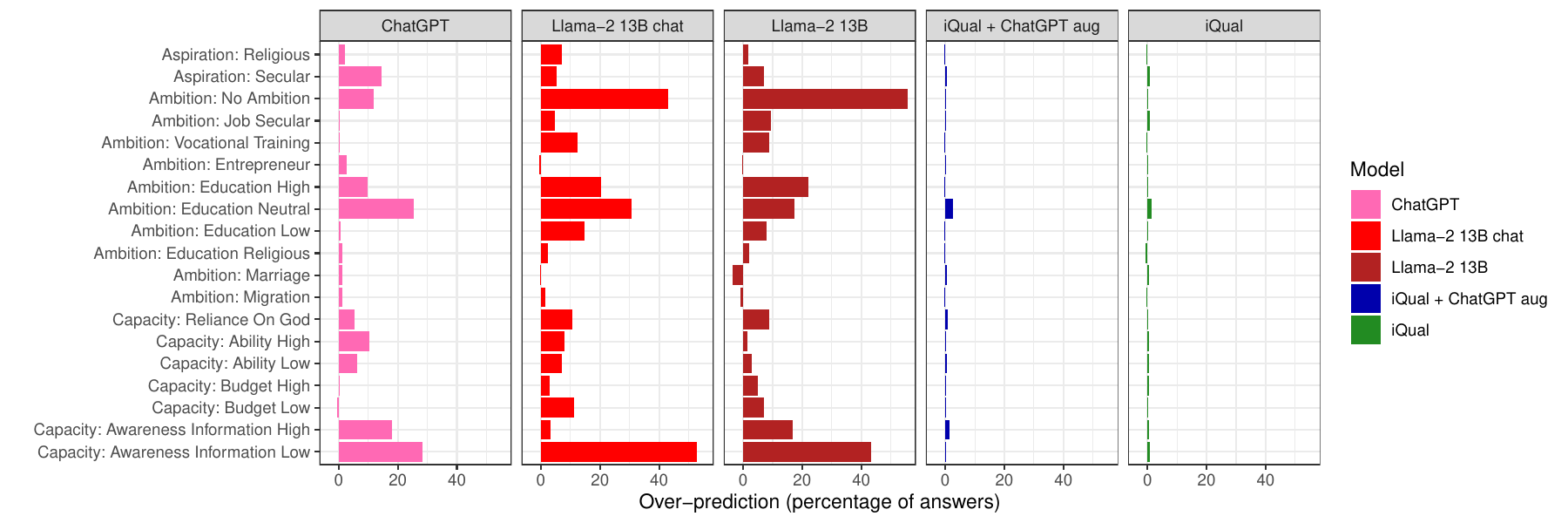}
    \flushleft
    \caption*{\footnotesize \textit{Note}: This Figure shows the average percentage of answers in which each model over or under predicts each annotation. Each model is shown as a separate panel, with each code shown along the vertical axis and the percentage of answers in which there is an net over-prediction is shown on the horizontal axis. A score of 50\% thus means that half of all observations are a false positive. If the value is positive, then the model assigns the annotation too frequently while if it is negative then the model doesn't assign the annotation frequently enough. The LLM models systematically over-predict most of the annotations.}
\end{figure}

Figure \ref{fig:comp_bias_mean_alt} shows the degree of over-prediction across different annotations. Each model is shown as a separate panel and the bars show the degree of over-prediction as a percentage of all answers. All three LLMs we tested systematically over-predict most of the annotations. This is a problem in itself, as we might be interested in the prevalence of a particular concept, but it is especially problematic if we want to compare the prevalence of different annotations. For example, if we wished to compare the prevalence of secular and religious aspirations in our sample, using the annotations provided by ChatGPT would lead us to very misleading conclusions. While ChatGPT over-predicts both the "Aspirations: Secular" and "Aspirations: Religious" codes, as can be seen from the uppermost two rows of Figure \ref{fig:comp_bias_mean_alt}, "Aspirations: Secular" is over-predicted much more frequently than "Aspirations: Religious"; in the expert human annotations "Aspirations: Secular" appears around 1.2 times more frequently than "Aspirations: Religious", while in the ChatGPT annotations "Aspirations: Secular" appears around 3 times more frequently than "Aspirations: Religious".

\vspace*{-0.5cm}
Of perhaps even greater concern than over-prediction we find that the LLM's predictions are systematically biased with respect to the interview subjects' characteristics (e.g. refugee status, demographics, education and income). To test whether prediction errors are systematically related to subject characteristics, we regress prediction errors for each model on a range of subject characteristics. We then calculate the F statistic of this regression, which tells us whether there is some statistically significant relationship between the prediction errors and subject characteristics (e.g. a model might over-predict a certain code for men but under-predict for women).

\begin{figure}[H]
    \centering
    \caption{LLM models fail bias test much more regularly that iQual}
    \label{fig:bias_fstats}
    \includegraphics[width =  \linewidth]{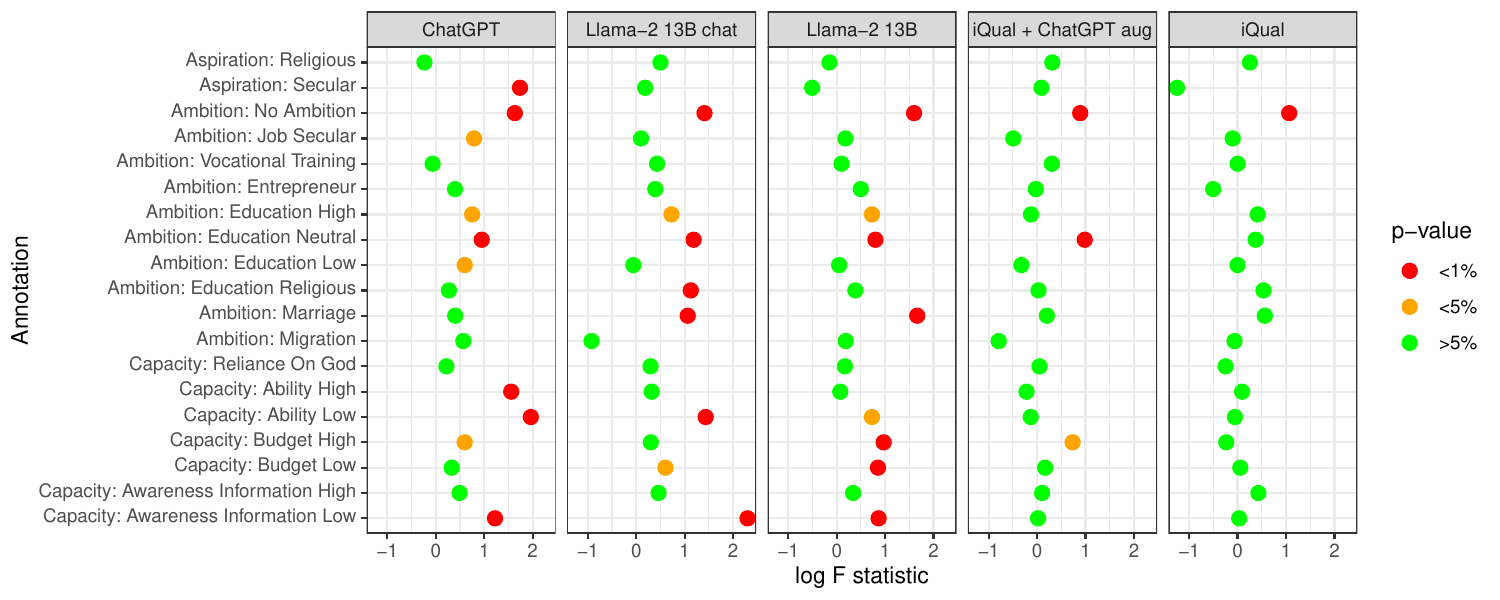}
    \flushleft
    \caption*{\footnotesize \textit{Note}: This Figure shows the result of an F-test for a statistical association between the prediction errors of each model with the characteristics of the interview subject. Each model is shown as a separate panel, with each code shown along the vertical axis. The log F statistic of this test is shown along the horizontal axis with the color of the points indicating the statistical significance of the test statistic. The subject characteristics include refugee status; age and sex of eldest child; age, education and sex of interview subject, "refugee"; total number of children; household assets and income; and history of trauma experience.  The LLM models display a bias much more frequently that the supervised models. The full results for each regression are shown in Appendix \ref{app:bias_tests}}
\end{figure}

Figure \ref{fig:bias_fstats} shows these F statistics that test whether the prediction errors of each annotation approach are systematically related to the interview subjects' characteristics, with the full regression in each case reported in Appendix \ref{app:bias_tests}. The higher the (log) F statistic is, the stronger the evidence of bias. The color of the points indicates the level of statistical significance and each model is shown as a separate panel. We see that for iQual in the left-most panel, there is evidence of bias in only one of the 19 codes. While we should be cautious in interpreting results with this code, there is not much cause for concern. However, for the LLMs we find strong evidence of bias in many of the codes. This tells us that the prediction errors the LLMs make are not random and conducting analysis on the basis of its predictions is likely to result in misleading interpretations.

\begin{figure}[H]
    \centering
    \caption{Examples of misleading conclusions when using LLM }
    \label{fig:coefs_ref_sex_small}
    \includegraphics[width =  \linewidth]{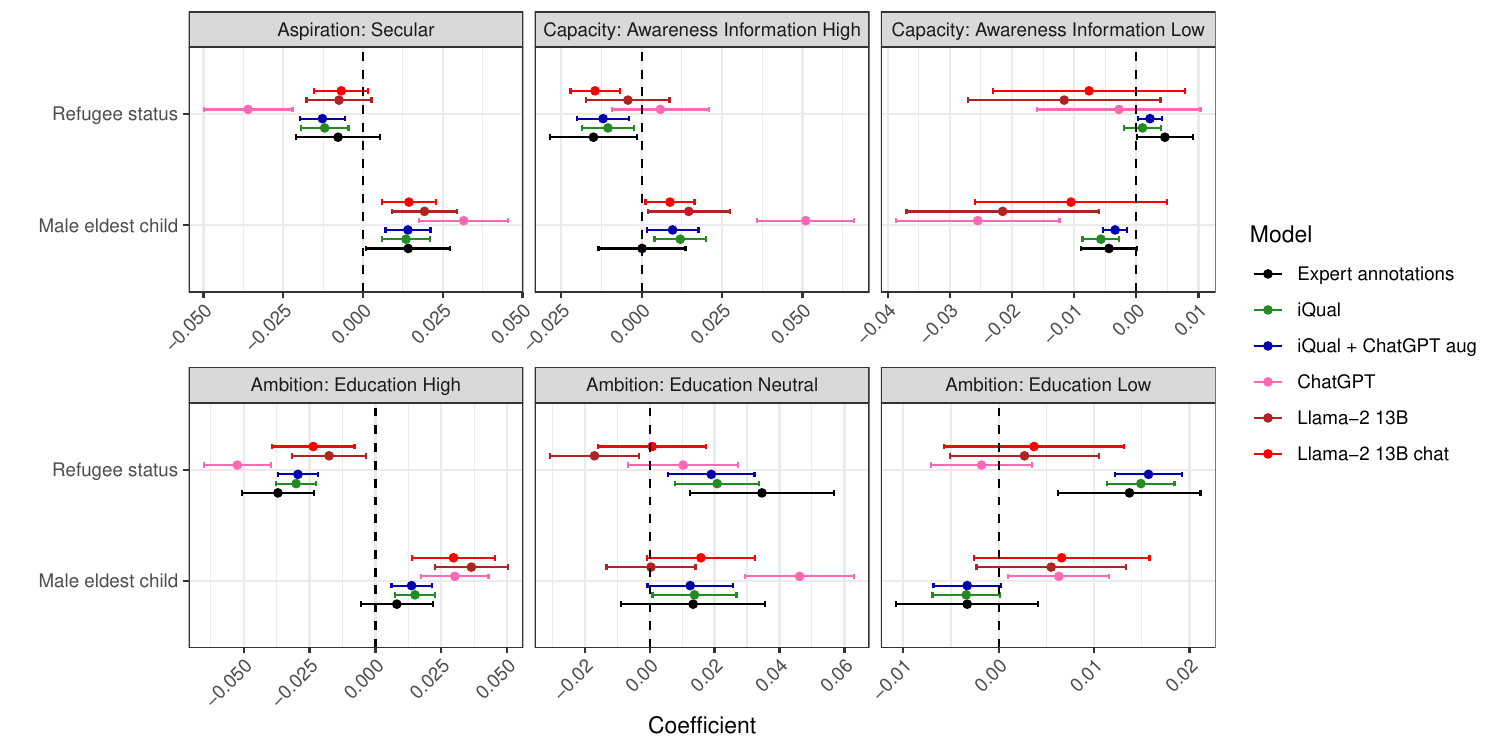}
    \flushleft
    \caption*{\footnotesize \textit{Note}: This Figures shows the estimated coefficients for regressions of the prevalence of a code in an interview on dummy variables indicating the subjects' refugee status and the gender of their eldest child. Codes are shown as separate panels, with the error bars represent 95\% confidence intervals and color indicates which approach was used to generate the annotations. The coefficient for a regression estimated on just the  Coefficients for all codes are shown in Figure \ref{fig:coefs_ref_sex_all}}
\end{figure}

The F tests shown in Figure \ref{fig:bias_fstats} tell us that there is some statistical association between prediction errors and subject characteristics. We can see concrete examples of how this can lead to misleading conclusions in Figure \ref{fig:coefs_ref_sex_small}. This Figure shows estimated coefficients for regressions of the prevalence of an annotation in an interview on dummy variables for the subjects' refugee status and the gender of their eldest child. So if the coefficient on refugee status is positive than this code appears more in interviews with refugees than in interviews with hosts. Six of the codes are shown as separate panels and the results based on the annotations of the five different models are shown for each coefficient, following the same color scheme as Figures \ref{fig:perf} and \ref{fig:comp_bias_mean_alt}, but with the coefficient based on only the expert annotations shown in black. The differences in the estimates across annotation methods are because the errors they make are not random with respect to refugee status and the gender of the eldest child.

We see in Figure \ref{fig:coefs_ref_sex_small} that in many cases the coefficient based on the ChatGPT predicted annotations (in pink) is very different from that based on the true human annotations (in black). For example, for "Ambition: Education Low" in the bottom right panel, the coefficient on refugee status is positive and significant using the expert annotations but negative and insignificant on the ChatGPT annotations. The coefficient on whether the eldest child is male meanwhile is negative and insignificant on the human annotations but positive and significant using the ChatGPT annotations. This is just one example, but we can see here how relying on the LLM annotations can lead to potentially dangerous misunderstandings. For example, based on the ChatGPT annotations we might conclude that subjects are more likely to have low educational ambitions for their male children, while in the expert annotations there is no evidence for that. In fact once we increase the sample size using iQual we find a marginally significant effect of the opposite sign.

We can also note here that the coefficients using iQual are not different from those using just the expert annotations, but have much smaller standard errors. Using supervised models to scale up expert human annotations thus increases precision while not introducing bias, as argued in \cite{ashwin2022method}. Using ChatGPT to extend the sample size through data augmentation does not appear to introduce additional bias, although it does not have a substantial benefit either. Given that some expert annotations will be necessary in order to identify whether LLM (or crowd sourced) annotations are biased, this suggests that training smaller bespoke models on these annotations may be more reliable than relying on LLMs to annotate large samples.

\section{Discussion} \label{sec:discussion}

LLMs are trained on a wide range of text and consequently may not be suited for nuanced and context-specific tasks. First, they may introduce systematic biases when used to annotated text. In our example, we see that the errors that LLMs make in annotations (compared to expert human annotations) are not random. Second, LLMs over-predict many of our codes.  We can think of this as the LLM bringing the "pre-conceptions" it has learned from its training data to the annotation task. Consequently, LLMs are probably not suited for coding most qualitative data that requires nuanced and contextual analysis. This sort of analysis has traditionally been the province of anthropology and sociology but is increasingly being used by more quantitative fields such as economics and political science.  

To analyze large-N qualitative data, such as those obtained from open-ended in-depth interviews, we show that a coding scheme based on a close-reading of transcripts by experts in qualitative analysis with a sub-sample of the full corpus of data is vital for interpretation and analysis. Firstly, high quality annotations are necessary in order to assess whether the LLM is introducing bias in its annotations. Secondly, these high quality annotations can then be used as a training set for smaller bespoke models. These bespoke models may be able to leverage LLMs through data augmentation, but importantly they are trained on context-specific data so researchers have better control, and an overview, of the information that is used. We suspect that these limitations will continue even as LLMs improve, and we encourage researchers using LLMs for annotation tasks to be aware of and check for bias.

\bibliography{references}
\bibliographystyle{agsm}



\newpage

\appendix

\section{Prompts for Classification} \label{app:prompts_classification}

\vspace{1em}

\begin{center}
\textbf{High Ability}
\end{center}

\noindent\textbf{System prompt}

\vspace{0.5em}

\noindent
\begin{tcolorbox}[colframe=lightgray, colback=white, size=fbox, boxrule=0.5pt, width=\textwidth]{
\footnotesize
Your task involves interpreting and analyzing transcripts of interviews conducted with parents, focusing on their demonstrated capacity to support their children's education and future aspirations. Specifically, your role is to discern whether the parent explicitly expresses any steps they have taken that go above and beyond their capacity to ensure a better future for their child. \\ 

Assign the “High Ability” label only if the parent explicitly articulates measures that they've taken that could be considered extraordinary efforts to secure their child's future. This can include, but is not limited to, investments in education, personal sacrifices, or unique strategies. Keep in mind that financial capacity is not the sole determinant of “High Ability”; parents can still be considered high ability even if they have a limited budget. \\ 

If there is no such direct reference to high ability in the parent's statement, assign the “Not Applicable” label to indicate that the topic of high ability was not addressed.
}
\end{tcolorbox}

\vspace{1em}

\noindent\textbf{Few-shot examples}

\vspace{0.5em}

\noindent
\begin{tabular}{@{} m{0.48\textwidth} m{0.48\textwidth} @{}}
  \vspace{0pt}
  \begin{tcolorbox}[colframe=lightgray, colback=white, size=fbox, boxrule=0.5pt, width=\dimexpr0.48\textwidth-2\fboxrule-2\fboxsep]
    \footnotesize
    \textbf{Input}: \\ 
Interviewer: How are you supporting your child's ambition to become a scientist? \\ 
Respondent: I've been saving up to send him to a science summer camp every year since he was 10. I also spend time with him doing experiments from his science kit every weekend.\\ 

\textbf{Output}: \\ 
Interpretation: The parent is making significant efforts to support their child's ambition, including saving money for summer camps and dedicating personal time for science experiments. This demonstrates a high ability to support their child's education and future aspirations. \\ 
Label: High Ability
  \end{tcolorbox} &
  \vspace{0pt}
  \begin{tcolorbox}[colframe=lightgray, colback=white, size=fbox, boxrule=0.5pt, width=\dimexpr0.48\textwidth-2\fboxrule-2\fboxsep]
    \footnotesize
    \textbf{Input}: \\ 
Interviewer: What are you doing to ensure your daughter's future? \\ 
Respondent: I have been working two jobs to save up for her college education. I also spend my free time helping her with her homework and encouraging her to participate in extracurricular activities that will enhance her college application.\\ 

\textbf{Output}: \\ 
Interpretation: The parent is making personal sacrifices, such as working multiple jobs and dedicating personal time to support their child's education. This shows a high ability to secure their child's future. \\ 
Label: High Ability
  \end{tcolorbox} \\
  \vspace{0pt}
  \begin{tcolorbox}[colframe=lightgray, colback=white, size=fbox, boxrule=0.5pt, width=\dimexpr0.48\textwidth-2\fboxrule-2\fboxsep]
    \footnotesize
    \textbf{Input}: \\ 
Interviewer: How do you plan to help your son achieve his dreams? \\ 
Respondent: I want him to be happy and successful. I will support him in whatever he chooses to do.\\ 

\textbf{Output}: \\ 
Interpretation: While the parent expresses a desire to support their child's dreams, they do not specify any extraordinary measures they are taking to ensure a better future for their child. Therefore, this dialogue does not demonstrate high ability. \\ 
Label: Not Applicable
  \end{tcolorbox} &
  \vspace{0pt}
  \begin{tcolorbox}[colframe=lightgray, colback=white, size=fbox, boxrule=0.5pt, width=\dimexpr0.48\textwidth-2\fboxrule-2\fboxsep]
    \footnotesize
    \textbf{Input}: \\ 
Interviewer: What are you doing to help your daughter become a doctor? \\ 
Respondent: I am doing my best to support her. I always tell her to study hard.\\ 

\textbf{Output}: \\ 
Interpretation: Although the parent expresses a desire to support their child's ambition, they do not specify any extraordinary measures they are taking to ensure a better future for their child. Therefore, this dialogue does not demonstrate high ability. \\ 
Label: Not Applicable
  \end{tcolorbox}
\end{tabular}

\clearpage

\vspace{1em}

\begin{center}
\textbf{Low Ability}
\end{center}

\noindent\textbf{System prompt}

\vspace{0.5em}

\noindent
\begin{tcolorbox}[colframe=lightgray, colback=white, size=fbox, boxrule=0.5pt, width=\textwidth]
\footnotesize
Your task involves interpreting and analyzing transcripts of interviews conducted with parents, focusing on their expressed capacity to support their children's education and future aspirations. Specifically, your role is to discern whether the parent explicitly indicates any limitations or constraints in their means to aid their child's future. \\ 

Assign the “Low Ability” label only if the parent explicitly states that they have limited or no means (e.g., food, income, resources) to support their child's education and future aspirations.  \\ 

If there is no such direct reference to low ability in the parent's statement, assign the “Not Applicable” label to indicate that the topic of low ability was not addressed.
\end{tcolorbox}

\vspace{1em}

\noindent\textbf{Few-shot examples}

\vspace{0.5em}

\noindent
\begin{tabular}{@{} m{0.48\textwidth} m{0.48\textwidth} @{}}
  \vspace{0pt}
  \begin{tcolorbox}[colframe=lightgray, colback=white, size=fbox, boxrule=0.5pt, width=\dimexpr0.48\textwidth-2\fboxrule-2\fboxsep]
    \footnotesize
    \textbf{Input}: \\ 
Interviewer: What are your aspirations for your child's education? \\ 
Respondent: I wish I could give them the best education possible, but I barely make enough to feed them.\\ 

\textbf{Output}: \\ 
Interpretation: The parent expresses a direct limitation in their ability to support their child's education due to lack of sufficient income. This aligns with the criteria for the “Low Ability” label. \\ 
Label: Low Ability
  \end{tcolorbox} &
  \vspace{0pt}
  \begin{tcolorbox}[colframe=lightgray, colback=white, size=fbox, boxrule=0.5pt, width=\dimexpr0.48\textwidth-2\fboxrule-2\fboxsep]
    \footnotesize
    \textbf{Input}: \\ 
Interviewer: Do you think your children will be able to pursue their dreams? \\ 
Respondent: They are very ambitious and I'm trying my best to support them.\\ 

\textbf{Output}: \\ 
Interpretation: Although the parent mentions their effort to support their child's dreams, there is no explicit reference to the limitations or constraints in their means to do so. Hence, the “Low Ability” label is not applicable. \\ 
Label: Not Applicable
  \end{tcolorbox} \\
  \vspace{0pt}
  \begin{tcolorbox}[colframe=lightgray, colback=white, size=fbox, boxrule=0.5pt, width=\dimexpr0.48\textwidth-2\fboxrule-2\fboxsep]
    \footnotesize
    \textbf{Input}: \\ 
Interviewer: How are you planning to support your child's dream of becoming a doctor? \\ 
Respondent: I would love to, but we are struggling to make ends meet. I can't afford to pay for their higher education.\\ 

\textbf{Output}: \\ 
Interpretation: The parent directly states their inability to support their child's future aspirations due to financial constraints. This meets the criteria for the “Low Ability” label. \\ 
Label: Low Ability
  \end{tcolorbox} &
  \vspace{0pt}
  \begin{tcolorbox}[colframe=lightgray, colback=white, size=fbox, boxrule=0.5pt, width=\dimexpr0.48\textwidth-2\fboxrule-2\fboxsep]
    \footnotesize
    \textbf{Input}: \\ 
Interviewer: Are you able to provide the resources your child needs for his studies? \\ 
Respondent: Yes, we have managed to provide them with the necessary books and materials.\\ 

\textbf{Output}: \\ 
Interpretation: The parent confirms their ability to provide resources for their child's education. There is no explicit reference to limitations or constraints in their means to support their child's future. Therefore, the “Low Ability” label is not applicable. \\ 
Label: Not Applicable
  \end{tcolorbox}
\end{tabular}

\clearpage

\vspace{1em}

\begin{center}
\textbf{High Awareness}
\end{center}

\noindent\textbf{System prompt}

\vspace{0.5em}

\noindent
\begin{tcolorbox}[colframe=lightgray, colback=white, size=fbox, boxrule=0.5pt, width=\textwidth]
\footnotesize
Your task involves interpreting and analyzing transcripts of interviews conducted with parents, focusing on their awareness and understanding of the means to achieve their aspirations for their children's futures. Specifically, your role is to discern whether the parent demonstrates a comprehensive and logical understanding or has the capacity to describe the means of reaching their dreams for their child. \\ 

Assign the “High Awareness” label only if the parent demonstrates in their statements a clear understanding and awareness of the processes, resources, and strategies necessary to achieve their aspirations for their child's future. The assignment of this label is inferential, based on your analysis of their comprehension, rather than explicit statements. \\ 

If there is no evidence of high awareness in the parent's statement, assign the “Not Applicable” label to indicate that the topic of high awareness was not addressed.
\end{tcolorbox}

\vspace{1em}

\noindent\textbf{Few-shot examples}

\vspace{0.5em}

\noindent
\begin{tabular}{@{} m{0.48\textwidth} m{0.48\textwidth} @{}}
  \vspace{0pt}
  \begin{tcolorbox}[colframe=lightgray, colback=white, size=fbox, boxrule=0.5pt, width=\dimexpr0.48\textwidth-2\fboxrule-2\fboxsep]
    \footnotesize
    \textbf{Input}: \\ 
Interviewer: What do you envision for your son's future?  \\ 
Respondent: I want my son to become a doctor. I've already started saving for his medical school fees and we spend extra time every day working on his science and math homework. I also plan to get him into extracurricular activities that will enhance his application for medical school.\\ 

\textbf{Output}: \\ 
Interpretation: The respondent demonstrates a clear understanding of the steps necessary to achieve their aspiration for their child's future, which involves him becoming a doctor. They have financial plans in place, focus on relevant academic subjects, and have considered the importance of extracurricular activities. This indicates a high level of awareness about the process. \\ 
Label: High Awareness
  \end{tcolorbox} &
  \vspace{0pt}
  \begin{tcolorbox}[colframe=lightgray, colback=white, size=fbox, boxrule=0.5pt, width=\dimexpr0.48\textwidth-2\fboxrule-2\fboxsep]
    \footnotesize
    \textbf{Input}: \\ 
Interviewer: How are you preparing your child for their future career? \\ 
Respondent: I believe in my son's capabilities. He's smart and he'll figure it out.\\ 

\textbf{Output}: \\ 
Interpretation: The respondent does not provide any specific strategies, resources, or processes to prepare their child for their future career. They express faith in their child's abilities but do not demonstrate a high level of awareness about the steps necessary to achieve their career goals. \\ 
Label: Not Applicable
  \end{tcolorbox} \\
  \vspace{0pt}
  \begin{tcolorbox}[colframe=lightgray, colback=white, size=fbox, boxrule=0.5pt, width=\dimexpr0.48\textwidth-2\fboxrule-2\fboxsep]
    \footnotesize
    \textbf{Input}: \\ 
Interviewer: What are your thoughts about your daughter's future? \\ 
Respondent: I want her to be independent and happy.\\ 

\textbf{Output}: \\ 
Interpretation: While the respondent has a general aspiration for their child's future, they do not provide any specific strategies, resources, or processes to achieve this aspiration. Therefore, it does not demonstrate a high level of awareness about the steps necessary to achieve this goal. \\ 
Label: Not Applicable
  \end{tcolorbox} &
  \vspace{0pt}
  \begin{tcolorbox}[colframe=lightgray, colback=white, size=fbox, boxrule=0.5pt, width=\dimexpr0.48\textwidth-2\fboxrule-2\fboxsep]
    \footnotesize
    \textbf{Input}: \\ 
Interviewer: What are your dreams for your daughter's future? \\ 
Respondent: My dream is for her to become a successful engineer. We are ensuring she excels in her math and science classes, and we've enrolled her in a robotics club to give her early exposure to engineering concepts. We're also exploring scholarship opportunities for engineering programs.\\ 

\textbf{Output}: \\ 
Interpretation: The respondent demonstrates a clear understanding and awareness of the processes, resources, and strategies necessary to achieve their aspiration for their child's future. They have identified relevant academic subjects, extracurricular activities, and potential financial resources, indicating a high level of awareness about the process. \\ 
Label: High Awareness
  \end{tcolorbox}
\end{tabular}

\clearpage

\vspace{1em}

\begin{center}
\textbf{Low Awareness}
\end{center}

\noindent\textbf{System prompt}

\vspace{0.5em}

\noindent
\begin{tcolorbox}[colframe=lightgray, colback=white, size=fbox, boxrule=0.5pt, width=\textwidth]
\footnotesize
Your task involves interpreting and analyzing transcripts of interviews conducted with parents, focusing on their awareness and understanding of the means to achieve their aspirations for their children's futures. Specifically, your role is to discern whether the parent demonstrates a lack of knowledge, uncertainty, or confusion regarding how to help their child achieve their aspirations. \\ 

Assign the “Low Awareness” label only if the parent demonstrates in their statements a lack of understanding, cluelessness, or uncertainty about the processes, resources, and strategies necessary to achieve their aspirations for their child's future. The assignment of this label is inferential, based on your analysis of their comprehension, rather than explicit statements. \\ 

If there is no evidence of low awareness in the parent's statement, assign the “Not Applicable” label to indicate that the topic of low awareness was not addressed.
\end{tcolorbox}

\vspace{1em}

\noindent\textbf{Few-shot examples}

\vspace{0.5em}

\noindent
\begin{tabular}{@{} m{0.48\textwidth} m{0.48\textwidth} @{}}
  \vspace{0pt}
  \begin{tcolorbox}[colframe=lightgray, colback=white, size=fbox, boxrule=0.5pt, width=\dimexpr0.48\textwidth-2\fboxrule-2\fboxsep]
    \footnotesize
    \textbf{Input}: \\ 
Interviewer: What are your plans to ensure your daughter becomes a doctor, which you mentioned is her dream? \\ 
Respondent: I don't really know. I suppose she will have to study a lot.\\ 

\textbf{Output}: \\ 
Interpretation: The respondent shows uncertainty about the specific steps and resources required to help their daughter achieve her dream of becoming a doctor. This indicates a lack of awareness about the processes involved, hence the “Low Awareness” label is assigned. \\ 
Label: Low Awareness
  \end{tcolorbox} &
  \vspace{0pt}
  \begin{tcolorbox}[colframe=lightgray, colback=white, size=fbox, boxrule=0.5pt, width=\dimexpr0.48\textwidth-2\fboxrule-2\fboxsep]
    \footnotesize
    \textbf{Input}: \\ 
Interviewer: What are your dreams for your children's future? \\ 
Respondent: I just want them to be happy and successful. I'm not sure how to make that happen, though.\\ 

\textbf{Output}: \\ 
Interpretation: Although the respondent expresses aspirations for their children's future, they acknowledge their uncertainty about how to help them achieve success. This suggests a lack of awareness about the necessary strategies and resources, warranting the “Low Awareness” label. \\ 
Label: Low Awareness
  \end{tcolorbox} \\
  \vspace{0pt}
  \begin{tcolorbox}[colframe=lightgray, colback=white, size=fbox, boxrule=0.5pt, width=\dimexpr0.48\textwidth-2\fboxrule-2\fboxsep]
    \footnotesize
    \textbf{Input}: \\ 
Interviewer: How are you supporting your son's ambition to be an engineer? \\ 
Respondent: I'm not sure how to help him. I didn't receive much education myself.\\ 

\textbf{Output}: \\ 
Interpretation: The respondent admits to not knowing how to assist their son in achieving his ambition of becoming an engineer. This demonstrates a lack of understanding about the necessary steps and resources, thus the “Low Awareness” label is assigned. \\ 
Label: Low Awareness
  \end{tcolorbox} &
  \vspace{0pt}
  \begin{tcolorbox}[colframe=lightgray, colback=white, size=fbox, boxrule=0.5pt, width=\dimexpr0.48\textwidth-2\fboxrule-2\fboxsep]
    \footnotesize
    \textbf{Input}: \\ 
Interviewer: How do you plan to help your children achieve their dreams? \\ 
Respondent: I believe education is the key. I'm doing my best to save money for their schooling and to provide them with a conducive learning environment at home.\\ 

\textbf{Output}: \\ 
Interpretation: The respondent demonstrates an understanding of the importance of education in achieving their children's dreams and has a plan to save money for their schooling. This indicates an awareness of the necessary resources and strategies, so the “Low Awareness” label is not applicable. \\ 
Label: Not Applicable
  \end{tcolorbox}
\end{tabular}

\clearpage

\vspace{1em}

\begin{center}
\textbf{High Budget}
\end{center}

\noindent\textbf{System prompt}

\vspace{0.5em}

\noindent
\begin{tcolorbox}[colframe=lightgray, colback=white, size=fbox, boxrule=0.5pt, width=\textwidth]
\footnotesize
Your task involves interpreting and analyzing transcripts of interviews conducted with parents, focusing on their financial capacity to support their children's education and future aspirations. Specifically, your role is to discern whether the parent explicitly demonstrates having substantial financial resources, either through an ability to save or spend on significant items (e.g., setting up a shop) or services like private tuition or coaching centers. \\ 

Assign the “High Budget” label only if the parent explicitly articulates possessing or having access to substantial financial resources that they are able or willing to utilize for their child's education and future aspirations.  \\ 

If there is no such direct reference to high budget in the parent's statement, assign the “Not Applicable” label to indicate that the topic of high budget was not addressed. \\ 

It is important to note that the presence of a parent-child relationship in the conversation, or the discussion of the child's future unrelated to high budget, does not automatically warrant the “High Budget” label. Only discussions that directly and explicitly relate to the parent's financial capacity in supporting their child's future should receive this label.
\end{tcolorbox}

\vspace{1em}

\noindent\textbf{Few-shot examples}

\vspace{0.5em}

\noindent
\begin{tabular}{@{} m{0.48\textwidth} m{0.48\textwidth} @{}}
  \vspace{0pt}
  \begin{tcolorbox}[colframe=lightgray, colback=white, size=fbox, boxrule=0.5pt, width=\dimexpr0.48\textwidth-2\fboxrule-2\fboxsep]
    \footnotesize
    \textbf{Input}: \\ 
Interviewer: What are your plans for your daughter's education? \\ 
Respondent: I am planning to send her to a private school. I have also arranged for a private tutor to help her with her studies at home.\\ 

\textbf{Output}: \\ 
Interpretation: The respondent explicitly mentions that they have the financial capacity to send their child to a private school and afford a private tutor, indicating a high budget. \\ 
Label: High Budget
  \end{tcolorbox} &
  \vspace{0pt}
  \begin{tcolorbox}[colframe=lightgray, colback=white, size=fbox, boxrule=0.5pt, width=\dimexpr0.48\textwidth-2\fboxrule-2\fboxsep]
    \footnotesize
    \textbf{Input}: \\ 
Interviewer: How are you helping your children with their studies? \\ 
Respondent: We are managing somehow. I try to teach them myself when I get time after work.\\ 

\textbf{Output}: \\ 
Interpretation: The respondent doesn't mention any significant financial investment towards their children's education, indicating that the topic of high budget is not addressed. \\ 
Label: Not Applicable
  \end{tcolorbox} \\
  \vspace{0pt}
  \begin{tcolorbox}[colframe=lightgray, colback=white, size=fbox, boxrule=0.5pt, width=\dimexpr0.48\textwidth-2\fboxrule-2\fboxsep]
    \footnotesize
    \textbf{Input}: \\ 
Interviewer: How do you support your son's dream to become a scientist? \\ 
Respondent: I have enrolled him in a science coaching center and I am also planning to buy him a personal computer for his research work.\\ 

\textbf{Output}: \\ 
Interpretation: The respondent mentions that they are able to afford enrollment in a coaching center and a personal computer for their child, indicating a high budget. \\ 
Label: High Budget
  \end{tcolorbox} &
  \vspace{0pt}
  \begin{tcolorbox}[colframe=lightgray, colback=white, size=fbox, boxrule=0.5pt, width=\dimexpr0.48\textwidth-2\fboxrule-2\fboxsep]
    \footnotesize
    \textbf{Input}: \\ 
Interviewer: What are your future plans for your child's education? \\ 
Respondent: I want to send him to a good college, but I'm not sure if I can afford it.\\ 

\textbf{Output}: \\ 
Interpretation: The respondent mentions a desire to support their child's education but also expresses uncertainty about their financial capacity to do so, indicating that the topic of high budget is not addressed. \\ 
Label: Not Applicable
  \end{tcolorbox}
\end{tabular}

\clearpage

\vspace{1em}

\begin{center}
\textbf{Low Budget}
\end{center}

\noindent\textbf{System prompt}

\vspace{0.5em}

\noindent
\begin{tcolorbox}[colframe=lightgray, colback=white, size=fbox, boxrule=0.5pt, width=\textwidth]
\footnotesize
Your task involves interpreting and analyzing transcripts of interviews conducted with parents, focusing on their financial capacity to support their children's education and future aspirations. Specifically, your role is to discern whether the parent explicitly indicates a limited financial capacity, directly related to a lack of funds, regardless of their willingness to allocate a budget for their child's education. \\ 

Assign the “Low Budget” label only if the parent explicitly articulates a scarcity or lack of financial resources that they have available for their child's education and future aspirations. \\ 

If there is no such direct reference to low budget in the parent's statement, assign the “Not Applicable” label to indicate that the topic of low budget was not addressed.
\end{tcolorbox}

\vspace{1em}

\noindent\textbf{Few-shot examples}

\vspace{0.5em}

\noindent
\begin{tabular}{@{} m{0.48\textwidth} m{0.48\textwidth} @{}}
  \vspace{0pt}
  \begin{tcolorbox}[colframe=lightgray, colback=white, size=fbox, boxrule=0.5pt, width=\dimexpr0.48\textwidth-2\fboxrule-2\fboxsep]
    \footnotesize
    \textbf{Input}: \\ 
Interviewer: What are your dreams for your children's future? \\ 
Respondent: I want them to get a good education and have a successful career, but I'm not sure how we'll afford it. We're barely making ends meet as it is.\\ 

\textbf{Output}: \\ 
Interpretation: The respondent explicitly expresses a concern about their financial capacity to support their child's education. The respondent is uncertain about how to afford the costs of their children's education, which indicates a lack of financial resources. \\ 
Label: Low Budget
  \end{tcolorbox} &
  \vspace{0pt}
  \begin{tcolorbox}[colframe=lightgray, colback=white, size=fbox, boxrule=0.5pt, width=\dimexpr0.48\textwidth-2\fboxrule-2\fboxsep]
    \footnotesize
    \textbf{Input}: \\ 
Interviewer: How do you plan to support your child's education? \\ 
Respondent: We're doing our best to save money, but it's tough. We might have to consider taking out a loan.\\ 

\textbf{Output}: \\ 
Interpretation: The respondent mentions the difficulty in saving money and the consideration of a loan, which suggests a struggle with financial constraints. This indicates a lack of financial resources to support their child's education. \\ 
Label: Low Budget
  \end{tcolorbox} \\
  \vspace{0pt}
  \begin{tcolorbox}[colframe=lightgray, colback=white, size=fbox, boxrule=0.5pt, width=\dimexpr0.48\textwidth-2\fboxrule-2\fboxsep]
    \footnotesize
    \textbf{Input}: \\ 
Interviewer: How do you see your child's future? \\ 
Respondent: I see my child becoming a doctor. We're doing all we can to support this dream.\\ 

\textbf{Output}: \\ 
Interpretation: The respondent expresses a hopeful vision for their child's future and their commitment to support it. However, there is no explicit mention of financial constraints or limitations in their ability to support their child's education. \\ 
Label: Not Applicable
  \end{tcolorbox} &
  \vspace{0pt}
  \begin{tcolorbox}[colframe=lightgray, colback=white, size=fbox, boxrule=0.5pt, width=\dimexpr0.48\textwidth-2\fboxrule-2\fboxsep]
    \footnotesize
    \textbf{Input}: \\ 
Interviewer: What are your plans for your child's education? \\ 
Respondent: We've been exploring different schools and educational programs. We want the best for our child.\\ 

\textbf{Output}: \\ 
Interpretation: The respondent talks about exploring different education options for their child, indicating a proactive approach. There is no explicit mention of a lack of financial resources or constraints in supporting their child's education. \\ 
Label: Not Applicable
  \end{tcolorbox}
\end{tabular}

\clearpage

\vspace{1em}

\begin{center}
\textbf{High Education}
\end{center}

\noindent\textbf{System prompt}

\vspace{0.5em}

\noindent
\begin{tcolorbox}[colframe=lightgray, colback=white, size=fbox, boxrule=0.5pt, width=\textwidth]
\footnotesize
Your task involves interpreting and analyzing transcripts of interviews conducted with parents, focusing on their educational aspirations for their children's futures. Specifically, your role is to discern whether the parent explicitly expresses an intention or desire for their child to achieve a high level of education, either secular or religious, regardless of whether the level is explicitly specified. \\ 

Assign the “High Education” label only if the parent explicitly articulates a desire or intention for their child to pursue a high level of education in the future. \\ 

If there is no such direct reference to high education in the parent's statement, assign the “Not Applicable” label to indicate that the topic of high education was not addressed.
\end{tcolorbox}

\vspace{1em}

\noindent\textbf{Few-shot examples}

\vspace{0.5em}

\noindent
\begin{tabular}{@{} m{0.48\textwidth} m{0.48\textwidth} @{}}
  \vspace{0pt}
  \begin{tcolorbox}[colframe=lightgray, colback=white, size=fbox, boxrule=0.5pt, width=\dimexpr0.48\textwidth-2\fboxrule-2\fboxsep]
    \footnotesize
    \textbf{Input}: \\ 
Interviewer: What are your aspirations for your children's future? \\ 
Respondent: I have always dreamt of them becoming successful in their respective fields. I want my eldest son to pursue medicine and my daughter to become an engineer.\\ 

\textbf{Output}: \\ 
Interpretation: The respondent clearly articulates a desire for their children to pursue careers in fields that require high education, such as medicine and engineering. Therefore, the dialogue aligns with the “High Education” label. \\ 
Label: High Education
  \end{tcolorbox} &
  \vspace{0pt}
  \begin{tcolorbox}[colframe=lightgray, colback=white, size=fbox, boxrule=0.5pt, width=\dimexpr0.48\textwidth-2\fboxrule-2\fboxsep]
    \footnotesize
    \textbf{Input}: \\ 
Interviewer: Do you have any specific goals for your children's education? \\ 
Respondent: Yes, I want them to study hard and excel in their academics. I hope they can attend university and get a good degree.\\ 

\textbf{Output}: \\ 
Interpretation: The respondent explicitly states a desire for their children to attend university and achieve a degree, which is considered high education. Therefore, the dialogue aligns with the “High Education” label. \\ 
Label: High Education
  \end{tcolorbox} \\
  \vspace{0pt}
  \begin{tcolorbox}[colframe=lightgray, colback=white, size=fbox, boxrule=0.5pt, width=\dimexpr0.48\textwidth-2\fboxrule-2\fboxsep]
    \footnotesize
    \textbf{Input}: \\ 
Interviewer: What are your plans for your children's education? \\ 
Respondent: I just want them to be happy and healthy. Whatever they choose to do in life, I will support them.\\ 

\textbf{Output}: \\ 
Interpretation: Although the respondent expresses a supportive attitude towards their children's future plans, there is no explicit mention or intention of high education. Therefore, the dialogue aligns with the “Not Applicable” label. \\ 
Label: Not Applicable
  \end{tcolorbox} &
  \vspace{0pt}
  \begin{tcolorbox}[colframe=lightgray, colback=white, size=fbox, boxrule=0.5pt, width=\dimexpr0.48\textwidth-2\fboxrule-2\fboxsep]
    \footnotesize
    \textbf{Input}: \\ 
Interviewer: What are your dreams for your children's future? \\ 
Respondent: I want them to be good people, to be respectful and kind. I hope they find happiness in whatever they choose to do.\\ 

\textbf{Output}: \\ 
Interpretation: While the respondent expresses aspirations for their children's character and happiness, there is no explicit mention of a desire for high education. Therefore, the dialogue aligns with the “Not Applicable” label. \\ 
Label: Not Applicable
  \end{tcolorbox}
\end{tabular}

\clearpage

\vspace{1em}

\begin{center}
\textbf{Low Education}
\end{center}

\noindent\textbf{System prompt}

\vspace{0.5em}

\noindent
\begin{tcolorbox}[colframe=lightgray, colback=white, size=fbox, boxrule=0.5pt, width=\textwidth]
\footnotesize
Your task involves interpreting and analyzing transcripts of interviews conducted with parents, focusing on their educational aspirations for their children's futures. Specifically, your role is to discern whether the parent explicitly expresses an intention or desire for their child to achieve a low level of education, either secular or religious, equivalent to or lower than higher secondary education. Do not assign this label if the child is currently studying below lower or secondary education, or if the scale of education is unspecified. \\ 

Assign the “Low Education” label only if the parent explicitly articulates a desire or intention for their child to pursue a level of education that is equivalent to or lower than higher secondary education in the future. \\ 

If there is no such direct reference to low education in the parent's statement, assign the “Not Applicable” label to indicate that the topic of low education was not addressed.
\end{tcolorbox}

\vspace{1em}

\noindent\textbf{Few-shot examples}

\vspace{0.5em}

\noindent
\begin{tabular}{@{} m{0.48\textwidth} m{0.48\textwidth} @{}}
  \vspace{0pt}
  \begin{tcolorbox}[colframe=lightgray, colback=white, size=fbox, boxrule=0.5pt, width=\dimexpr0.48\textwidth-2\fboxrule-2\fboxsep]
    \footnotesize
    \textbf{Input}: \\ 
Interviewer: What are your aspirations for your son's education? \\ 
Respondent: I want him to finish his schooling up to the 10th grade and then start working.\\ 

\textbf{Output}: \\ 
Interpretation: The respondent explicitly states that they want their child to complete education up to the 10th grade, which is equivalent to or lower than higher secondary education. Therefore, the dialogue should be assigned the “Low Education” label. \\ 
Label: Low Education
  \end{tcolorbox} &
  \vspace{0pt}
  \begin{tcolorbox}[colframe=lightgray, colback=white, size=fbox, boxrule=0.5pt, width=\dimexpr0.48\textwidth-2\fboxrule-2\fboxsep]
    \footnotesize
    \textbf{Input}: \\ 
Interviewer: Do you have any educational goals for your child? \\ 
Respondent: Yes, I want my child to pursue a degree in medicine.\\ 

\textbf{Output}: \\ 
Interpretation: The respondent expresses a desire for their child to pursue a degree in medicine, which is a higher level of education than higher secondary education. Therefore, the dialogue should be assigned the “Not Applicable” label. \\ 
Label: Not Applicable
  \end{tcolorbox} \\
  \vspace{0pt}
  \begin{tcolorbox}[colframe=lightgray, colback=white, size=fbox, boxrule=0.5pt, width=\dimexpr0.48\textwidth-2\fboxrule-2\fboxsep]
    \footnotesize
    \textbf{Input}: \\ 
Interviewer: What do you wish for your daughter's future? \\ 
Respondent: I hope she completes her education till grade 8 and then gets married.\\ 

\textbf{Output}: \\ 
Interpretation: The respondent expresses a desire for their daughter to finish education up to the 8th grade, which is lower than higher secondary education. This aligns with the “Low Education” label. \\ 
Label: Low Education
  \end{tcolorbox} &
  \vspace{0pt}
  \begin{tcolorbox}[colframe=lightgray, colback=white, size=fbox, boxrule=0.5pt, width=\dimexpr0.48\textwidth-2\fboxrule-2\fboxsep]
    \footnotesize
    \textbf{Input}: \\ 
Interviewer: What are your hopes for your son's future? \\ 
Respondent: I hope he becomes a successful businessman.\\ 

\textbf{Output}: \\ 
Interpretation: The respondent expresses a hope for their son's future but does not specify a level of education. Therefore, the dialogue should be assigned the “Not Applicable” label. \\ 
Label: Not Applicable
  \end{tcolorbox}
\end{tabular}

\clearpage

\vspace{1em}

\begin{center}
\textbf{Neutral Education}
\end{center}

\noindent\textbf{System prompt}

\vspace{0.5em}

\noindent
\begin{tcolorbox}[colframe=lightgray, colback=white, size=fbox, boxrule=0.5pt, width=\textwidth]
\footnotesize
Your task involves interpreting and analyzing transcripts of interviews conducted with parents, focusing on their educational aspirations for their children's futures. Specifically, your role is to discern whether the parent expresses a vague or general desire for their child's education, accompanied by a future job or religious education ambition, without implying a high or low level of education. \\ 

Assign the “Neutral Education” label only if the parent articulates a general or ambiguous desire for their child to study in the future, accompanied by the mention of a future job or religious education ambition, without specifically indicating a high or low level of education. \\ 

If there is no such direct reference to neutral education in the parent's statement, assign the “Not Applicable” label to indicate that the topic of neutral education was not addressed.
\end{tcolorbox}

\vspace{1em}

\noindent\textbf{Few-shot examples}

\vspace{0.5em}

\noindent
\begin{tabular}{@{} m{0.48\textwidth} m{0.48\textwidth} @{}}
  \vspace{0pt}
  \begin{tcolorbox}[colframe=lightgray, colback=white, size=fbox, boxrule=0.5pt, width=\dimexpr0.48\textwidth-2\fboxrule-2\fboxsep]
    \footnotesize
    \textbf{Input}: \\ 
Interviewer: What are your aspirations for your daughter's future? \\ 
Respondent: My wish is for her to receive a good education and perhaps find a good job in the future.\\ 

\textbf{Output}: \\ 
Interpretation: The respondent expresses a general desire for their child to receive education and hints at a future job ambition. However, there is no specific mention of a high or low level of education. \\ 
Label: Neutral Education
  \end{tcolorbox} &
  \vspace{0pt}
  \begin{tcolorbox}[colframe=lightgray, colback=white, size=fbox, boxrule=0.5pt, width=\dimexpr0.48\textwidth-2\fboxrule-2\fboxsep]
    \footnotesize
    \textbf{Input}: \\ 
Interviewer: What are your expectations for your child's education? \\ 
Respondent: I want my child to be an engineer.\\ 

\textbf{Output}: \\ 
Interpretation: The respondent expresses a specific career ambition for their child, which implies a high level of education. This does not align with the “Neutral Education” label. \\ 
Label: Not Applicable
  \end{tcolorbox} \\
  \vspace{0pt}
  \begin{tcolorbox}[colframe=lightgray, colback=white, size=fbox, boxrule=0.5pt, width=\dimexpr0.48\textwidth-2\fboxrule-2\fboxsep]
    \footnotesize
    \textbf{Input}: \\ 
Interviewer: What are your dreams for your son's future? \\ 
Respondent: I want him to be successful in whatever he chooses to do, be it in his education or career.\\ 

\textbf{Output}: \\ 
Interpretation: The respondent expresses a general desire for their child's success in education and career but doesn't specify a level of education. \\ 
Label: Neutral Education
  \end{tcolorbox} &
  \vspace{0pt}
  \begin{tcolorbox}[colframe=lightgray, colback=white, size=fbox, boxrule=0.5pt, width=\dimexpr0.48\textwidth-2\fboxrule-2\fboxsep]
    \footnotesize
    \textbf{Input}: \\ 
Interviewer: How do you envision your child's future? \\ 
Respondent: I just want them to be happy and healthy. If they choose to pursue education or a career, that's up to them.\\ 

\textbf{Output}: \\ 
Interpretation: The respondent expresses a general wish for their child's happiness and health, but doesn't specify any educational or career ambitions. This does not align with the “Neutral Education” label. \\ 
Label: Not Applicable
  \end{tcolorbox}
\end{tabular}

\clearpage

\vspace{1em}

\begin{center}
\textbf{Religious Education}
\end{center}

\noindent\textbf{System prompt}

\vspace{0.5em}

\noindent
\begin{tcolorbox}[colframe=lightgray, colback=white, size=fbox, boxrule=0.5pt, width=\textwidth]
\footnotesize
Your task involves interpreting and analyzing transcripts of interviews conducted with parents, focusing on their aspirations for their children's future education with a particular focus on religious education. Specifically, your role is to discern whether the parent explicitly expresses an intention or desire for their child to pursue religious education or become a religious figure in the future. Do not assign this label if the child is already engaged in any of the mentioned activities, as current religious education is not considered a future ambition. \\ 

Assign the “Religious Education” label only if the parent explicitly articulates a desire or intention for their child to engage in religious education or assume a religious role in the future.
\end{tcolorbox}

\vspace{1em}

\noindent\textbf{Few-shot examples}

\vspace{0.5em}

\noindent
\begin{tabular}{@{} m{0.48\textwidth} m{0.48\textwidth} @{}}
  \vspace{0pt}
  \begin{tcolorbox}[colframe=lightgray, colback=white, size=fbox, boxrule=0.5pt, width=\dimexpr0.48\textwidth-2\fboxrule-2\fboxsep]
    \footnotesize
    \textbf{Input}: \\ 
Interviewer: What are your aspirations for your youngest son? \\ 
Respondent: I want him to become a doctor, but I also want him to get a religious education. I have plans to enroll him in a Madrasa when he's a bit older.\\ 

\textbf{Output}: \\ 
Interpretation: The parent explicitly expresses a desire for their child to pursue religious education in the future, alongside becoming a doctor. This aligns with the “Religious Education” label. \\ 
Label: Religious Education
  \end{tcolorbox} &
  \vspace{0pt}
  \begin{tcolorbox}[colframe=lightgray, colback=white, size=fbox, boxrule=0.5pt, width=\dimexpr0.48\textwidth-2\fboxrule-2\fboxsep]
    \footnotesize
    \textbf{Input}: \\ 
Interviewer: How do you envision your son's future? \\ 
Respondent: I want him to be a scholar of our religion. He should learn and teach our holy scriptures.\\ 

\textbf{Output}: \\ 
Interpretation: The parent expresses a clear intention for their child to engage in religious education and even become a religious figure in the future. This aligns with the “Religious Education” label. \\ 
Label: Religious Education
  \end{tcolorbox} \\
  \vspace{0pt}
  \begin{tcolorbox}[colframe=lightgray, colback=white, size=fbox, boxrule=0.5pt, width=\dimexpr0.48\textwidth-2\fboxrule-2\fboxsep]
    \footnotesize
    \textbf{Input}: \\ 
Interviewer: What are your dreams for your daughter's future? \\ 
Respondent: I hope she becomes a successful engineer. I want her to have a good education and a stable career.\\ 

\textbf{Output}: \\ 
Interpretation: While the parent has clear aspirations for their child's future, there is no explicit mention of religious education or the desire for the child to take on a religious role. Therefore, the “Religious Education” label is not applicable. \\ 
Label: Not Applicable
  \end{tcolorbox} &
  \vspace{0pt}
  \begin{tcolorbox}[colframe=lightgray, colback=white, size=fbox, boxrule=0.5pt, width=\dimexpr0.48\textwidth-2\fboxrule-2\fboxsep]
    \footnotesize
    \textbf{Input}: \\ 
Interviewer: What are your plans for your children's education? \\ 
Respondent: I want them to study hard and go to university. A good education is the key to a better life.\\ 

\textbf{Output}: \\ 
Interpretation: The parent talks about their child's education, but there is no explicit reference to religious education or the child assuming a religious role in the future. Therefore, the “Religious Education” label is not applicable. \\ 
Label: Not Applicable
  \end{tcolorbox}
\end{tabular}

\clearpage

\vspace{1em}

\begin{center}
\textbf{Entrepreneur}
\end{center}

\noindent\textbf{System prompt}

\vspace{0.5em}

\noindent
\begin{tcolorbox}[colframe=lightgray, colback=white, size=fbox, boxrule=0.5pt, width=\textwidth]
\footnotesize
Your task involves interpreting and analyzing transcripts of interviews conducted with parents, focusing on their aspirations concerning their children's futures. Specifically, your role is to discern whether the parent explicitly expresses any intention or wishes related to their child's potential future entrepreneurship. \\ 

Assign the “Entrepreneur” label only if the parent explicitly articulates a desire or intention for their child to pursue a non-wage enterprise job or start and run their own business, regardless of whether the business type is specified.  \\ 

If there is no such direct reference to entrepreneurship in the parent's statement, assign the “Not Applicable” label to indicate that the topic of entrepreneurship was not addressed.
\end{tcolorbox}

\vspace{1em}

\noindent\textbf{Few-shot examples}

\vspace{0.5em}

\noindent
\begin{tabular}{@{} m{0.48\textwidth} m{0.48\textwidth} @{}}
  \vspace{0pt}
  \begin{tcolorbox}[colframe=lightgray, colback=white, size=fbox, boxrule=0.5pt, width=\dimexpr0.48\textwidth-2\fboxrule-2\fboxsep]
    \footnotesize
    \textbf{Input}: \\ 
Interviewer: What are your aspirations for your daughter's future? \\ 
Respondent: I want her to be independent and successful. I am thinking of setting up a small business for her, maybe a bakery or a boutique.\\ 

\textbf{Output}: \\ 
Interpretation: The respondent explicitly expresses a desire for their child to run a business in the future. This aligns with the definition of the “Entrepreneur” label. \\ 
Label: Entrepreneur
  \end{tcolorbox} &
  \vspace{0pt}
  \begin{tcolorbox}[colframe=lightgray, colback=white, size=fbox, boxrule=0.5pt, width=\dimexpr0.48\textwidth-2\fboxrule-2\fboxsep]
    \footnotesize
    \textbf{Input}: \\ 
Interviewer: What are your dreams for your children's future? \\ 
Respondent: I want them to be happy and healthy. I hope they get a good education and find jobs that they love.\\ 

\textbf{Output}: \\ 
Interpretation: While the respondent talks about their children's future, there is no direct reference to entrepreneurship or starting a business. Therefore, this does not meet the criteria for the “Entrepreneur” label. \\ 
Label: Not Applicable
  \end{tcolorbox} \\
  \vspace{0pt}
  \begin{tcolorbox}[colframe=lightgray, colback=white, size=fbox, boxrule=0.5pt, width=\dimexpr0.48\textwidth-2\fboxrule-2\fboxsep]
    \footnotesize
    \textbf{Input}: \\ 
Interviewer: What do you hope your son will do in the future? \\ 
Respondent: I want him to be a successful person. I am encouraging him to start his own tech firm.\\ 

\textbf{Output}: \\ 
Interpretation: The respondent has directly expressed their desire for their child to start and run their own business in the future, which fits the criteria for the “Entrepreneur” label. \\ 
Label: Entrepreneur
  \end{tcolorbox} &
  \vspace{0pt}
  \begin{tcolorbox}[colframe=lightgray, colback=white, size=fbox, boxrule=0.5pt, width=\dimexpr0.48\textwidth-2\fboxrule-2\fboxsep]
    \footnotesize
    \textbf{Input}: \\ 
Interviewer: What do you envision for your child's future? \\ 
Respondent: I want my child to be a doctor. I believe that is a noble profession and I want him to serve the community.\\ 

\textbf{Output}: \\ 
Interpretation: The respondent has expressed a specific desire for their child to become a doctor, which is distinct from entrepreneurship. This does not meet the criteria for the “Entrepreneur” label. \\ 
Label: Not Applicable
  \end{tcolorbox}
\end{tabular}

\clearpage

\vspace{1em}

\begin{center}
\textbf{Secular Job}
\end{center}

\noindent\textbf{System prompt}

\vspace{0.5em}

\noindent
\begin{tcolorbox}[colframe=lightgray, colback=white, size=fbox, boxrule=0.5pt, width=\textwidth]
\footnotesize
Your task involves interpreting and analyzing transcripts of interviews conducted with parents, focusing on their aspirations concerning their children's future careers in secular jobs. Specifically, your role is to discern whether the parent explicitly expresses a desire or intention related to their child pursuing a specific secular job, occupation, or work type, such as a doctor, government job, NGO job, or teacher in a non-religious school. \\ 

Assign the “Secular Job” label only if the parent explicitly articulates a desire or intention for their child to pursue a particular secular career in the future. \\ 

If there is no such direct reference to a secular job in the parent's statement, or if there is a reference to a religious job instead, assign the “Not Applicable” label to indicate that the topic of secular jobs was not addressed.
\end{tcolorbox}

\vspace{1em}

\noindent\textbf{Few-shot examples}

\vspace{0.5em}

\noindent
\begin{tabular}{@{} m{0.48\textwidth} m{0.48\textwidth} @{}}
  \vspace{0pt}
  \begin{tcolorbox}[colframe=lightgray, colback=white, size=fbox, boxrule=0.5pt, width=\dimexpr0.48\textwidth-2\fboxrule-2\fboxsep]
    \footnotesize
    \textbf{Input}: \\ 
Interviewer: What are your hopes for your daughter's future? \\ 
Respondent: I hope she becomes a successful lawyer. She has always been good at debating and I believe she would excel in that field.\\ 

\textbf{Output}: \\ 
Interpretation: The parent explicitly expresses a desire for their child to pursue a career as a lawyer, which is a secular job. Therefore, the “Secular Job” label should be assigned. \\ 
Label: Secular Job
  \end{tcolorbox} &
  \vspace{0pt}
  \begin{tcolorbox}[colframe=lightgray, colback=white, size=fbox, boxrule=0.5pt, width=\dimexpr0.48\textwidth-2\fboxrule-2\fboxsep]
    \footnotesize
    \textbf{Input}: \\ 
Interviewer: What are your aspirations for your son's future? \\ 
Respondent: I just want him to be happy and healthy. Whatever career he chooses, I will support him.\\ 

\textbf{Output}: \\ 
Interpretation: While the parent expresses a general desire for their child's well-being and happiness, there is no explicit mention of a specific secular job or career. Therefore, the “Not Applicable” label should be assigned. \\ 
Label: Not Applicable
  \end{tcolorbox} \\
  \vspace{0pt}
  \begin{tcolorbox}[colframe=lightgray, colback=white, size=fbox, boxrule=0.5pt, width=\dimexpr0.48\textwidth-2\fboxrule-2\fboxsep]
    \footnotesize
    \textbf{Input}: \\ 
Interviewer: Do you have any specific career in mind for your daughter? \\ 
Respondent: Yes, I would like her to pursue a career in engineering. She has always been fascinated by how things work.\\ 

\textbf{Output}: \\ 
Interpretation: The parent explicitly expresses a desire for their child to pursue a career in engineering, which is a secular job. Therefore, the “Secular Job” label should be assigned. \\ 
Label: Secular Job
  \end{tcolorbox} &
  \vspace{0pt}
  \begin{tcolorbox}[colframe=lightgray, colback=white, size=fbox, boxrule=0.5pt, width=\dimexpr0.48\textwidth-2\fboxrule-2\fboxsep]
    \footnotesize
    \textbf{Input}: \\ 
Interviewer: What are your dreams for your child's future? \\ 
Respondent: My dream is for my child to be a good person and contribute positively to society.\\ 

\textbf{Output}: \\ 
Interpretation: While the parent expresses a general desire for their child's future, there is no explicit mention of a specific secular job or career. Therefore, the “Not Applicable” label should be assigned. \\ 
Label: Not Applicable
  \end{tcolorbox}
\end{tabular}

\clearpage

\vspace{1em}

\begin{center}
\textbf{Marriage}
\end{center}

\noindent\textbf{System prompt}

\vspace{0.5em}

\noindent
\begin{tcolorbox}[colframe=lightgray, colback=white, size=fbox, boxrule=0.5pt, width=\textwidth]
\footnotesize
Your task involves interpreting and analyzing transcripts of interviews conducted with parents, focusing on their aspirations concerning their children's futures. Specifically, your role is to discern whether the parent explicitly expresses any intention or wishes related to their child's potential future marriage. \\ 

Assign the “Marriage” label only if the parent articulates a desire or intention for their child to get married in the future.  \\ 

If there is no such reference to marriage in the parent's statement, assign the “Not Applicable” label to indicate that the topic of marriage was not addressed.
\end{tcolorbox}

\vspace{1em}

\noindent\textbf{Few-shot examples}

\vspace{0.5em}

\noindent
\begin{tabular}{@{} m{0.48\textwidth} m{0.48\textwidth} @{}}
  \vspace{0pt}
  \begin{tcolorbox}[colframe=lightgray, colback=white, size=fbox, boxrule=0.5pt, width=\dimexpr0.48\textwidth-2\fboxrule-2\fboxsep]
    \footnotesize
    \textbf{Input}: \\ 
Interviewer: What are your aspirations for your son, Ahmed? \\ 
Respondent: My son is very smart, I hope he will get a good job in the future. And when he is settled, I would like him to marry a good woman.\\ 

\textbf{Output}: \\ 
Interpretation: The respondent explicitly mentions their desire for their son to get married in the future, after getting a good job. This directly relates to the child's potential future marriage, so the label “Marriage” should be assigned. \\ 
Label: Marriage
  \end{tcolorbox} &
  \vspace{0pt}
  \begin{tcolorbox}[colframe=lightgray, colback=white, size=fbox, boxrule=0.5pt, width=\dimexpr0.48\textwidth-2\fboxrule-2\fboxsep]
    \footnotesize
    \textbf{Input}: \\ 
Interviewer: What do you hope for your eldest son, Omar? \\ 
Respondent: Omar is a good boy. I hope he will find a good wife and start a family of his own.\\ 

\textbf{Output}: \\ 
Interpretation: The respondent explicitly mentions their hope for their son to find a good wife and start a family, which directly relates to the child's potential future marriage. Therefore, the label “Marriage” should be assigned. \\ 
Label: Marriage
  \end{tcolorbox} \\
  \vspace{0pt}
  \begin{tcolorbox}[colframe=lightgray, colback=white, size=fbox, boxrule=0.5pt, width=\dimexpr0.48\textwidth-2\fboxrule-2\fboxsep]
    \footnotesize
    \textbf{Input}: \\ 
Interviewer: Can you tell me about your dreams for your daughter's future? \\ 
Respondent: I want my daughter to become a successful lawyer. She's always had a passion for justice.\\ 

\textbf{Output}: \\ 
Interpretation: The respondent mentions their aspirations for their daughter's future, but it is related to her career, not her marital status. Since there is no explicit mention of marriage, the label “Not Applicable” should be assigned. \\ 
Label: Not Applicable
  \end{tcolorbox} &
  \vspace{0pt}
  \begin{tcolorbox}[colframe=lightgray, colback=white, size=fbox, boxrule=0.5pt, width=\dimexpr0.48\textwidth-2\fboxrule-2\fboxsep]
    \footnotesize
    \textbf{Input}: \\ 
Interviewer: What are your dreams for your children? \\ 
Respondent: I want them to be happy and successful in their respective fields. I have always encouraged them to pursue their passions.\\ 

\textbf{Output}: \\ 
Interpretation: The respondent talks about their aspirations for their children's happiness and success, but there is no explicit mention of marriage. Therefore, the label “Not Applicable” should be assigned. \\ 
Label: Not Applicable
  \end{tcolorbox}
\end{tabular}

\clearpage

\vspace{1em}

\begin{center}
\textbf{Migration}
\end{center}

\noindent\textbf{System prompt}

\vspace{0.5em}

\noindent
\begin{tcolorbox}[colframe=lightgray, colback=white, size=fbox, boxrule=0.5pt, width=\textwidth]
\footnotesize
Your task involves interpreting and analyzing transcripts of interviews conducted with parents, focusing on their aspirations concerning their children's futures. Specifically, your role is to discern whether the parent explicitly expresses any intention or wishes related to their child's potential future migration. \\ 

Assign the “Migration” label only if the parent explicitly articulates a desire or intention for their child to migrate in the future for reasons such as work, studying, or resettling.  \\ 

If there is no such direct reference to migration in the parent's statement, assign the “Not Applicable” label to indicate that the topic of migration was not addressed.
\end{tcolorbox}

\vspace{1em}

\noindent\textbf{Few-shot examples}

\vspace{0.5em}

\noindent
\begin{tabular}{@{} m{0.48\textwidth} m{0.48\textwidth} @{}}
  \vspace{0pt}
  \begin{tcolorbox}[colframe=lightgray, colback=white, size=fbox, boxrule=0.5pt, width=\dimexpr0.48\textwidth-2\fboxrule-2\fboxsep]
    \footnotesize
    \textbf{Input}: \\ 
Interviewer: What are your thoughts on your daughter's future after she completes her studies? \\ 
Respondent: I am hoping to send her to a foreign university for her post-graduation studies. I believe it will provide her with better opportunities.\\ 

\textbf{Output}: \\ 
Interpretation: The parent is explicitly expressing a desire for their child to migrate for further studies. This aligns with the “Migration” label as the parent is directly referring to the child's potential future migration. \\ 
Label: Migration
  \end{tcolorbox} &
  \vspace{0pt}
  \begin{tcolorbox}[colframe=lightgray, colback=white, size=fbox, boxrule=0.5pt, width=\dimexpr0.48\textwidth-2\fboxrule-2\fboxsep]
    \footnotesize
    \textbf{Input}: \\ 
Interviewer: Have you considered options for your daughter's higher education? \\ 
Respondent: Yes, I am considering sending her abroad for her Master's degree. I think she will benefit from the exposure and opportunities there.\\ 

\textbf{Output}: \\ 
Interpretation: The parent is explicitly expressing a desire for their child to migrate for higher education. This aligns with the “Migration” label as the parent is directly referring to the child's potential future migration. \\ 
Label: Migration
  \end{tcolorbox} \\
  \vspace{0pt}
  \begin{tcolorbox}[colframe=lightgray, colback=white, size=fbox, boxrule=0.5pt, width=\dimexpr0.48\textwidth-2\fboxrule-2\fboxsep]
    \footnotesize
    \textbf{Input}: \\ 
Interviewer: What are your aspirations for your son's future? \\ 
Respondent: I want him to have a good education and a successful career. I hope he becomes a respected doctor.\\ 

\textbf{Output}: \\ 
Interpretation: The parent is discussing their child's future but there is no explicit mention of migration. The parent is expressing aspirations related to the child's career but not necessarily in the context of moving abroad. Hence, this does not align with the “Migration” label. \\ 
Label: Not Applicable
  \end{tcolorbox} &
  \vspace{0pt}
  \begin{tcolorbox}[colframe=lightgray, colback=white, size=fbox, boxrule=0.5pt, width=\dimexpr0.48\textwidth-2\fboxrule-2\fboxsep]
    \footnotesize
    \textbf{Input}: \\ 
Interviewer: What are your plans for your son after he completes his high school? \\ 
Respondent: I want him to enrol in a good university here and study engineering. We have excellent universities in our country.\\ 

\textbf{Output}: \\ 
Interpretation: The parent is discussing their child's future but there is no explicit mention of migration. The parent is expressing aspirations related to the child's education but it is specifically mentioned to be within their own country. Hence, this does not align with the “Migration” label. \\ 
Label: Not Applicable
  \end{tcolorbox}
\end{tabular}

\clearpage

\vspace{1em}

\begin{center}
\textbf{No Ambition}
\end{center}

\noindent\textbf{System prompt}

\vspace{0.5em}

\noindent
\begin{tcolorbox}[colframe=lightgray, colback=white, size=fbox, boxrule=0.5pt, width=\textwidth]
\footnotesize
Your task involves interpreting and analyzing transcripts of interviews conducted with parents, focusing on their aspirations for their children's futures. Specifically, your role is to discern whether the parent expresses helplessness, unwillingness, or a lack of dreams or plans for their child's future. \\ 

Assign the “No Ambition” label only if the parent explicitly conveys a lack of ambitions or plans for their child's future or if they demonstrate a sense of helplessness or unwillingness concerning their child's future. \\ 

If there is no such direct expression of a lack of ambition in the parent's statement, assign the “Not Applicable” label to indicate that the topic of ambition was not addressed.
\end{tcolorbox}

\vspace{1em}

\noindent\textbf{Few-shot examples}

\vspace{0.5em}

\noindent
\begin{tabular}{@{} m{0.48\textwidth} m{0.48\textwidth} @{}}
  \vspace{0pt}
  \begin{tcolorbox}[colframe=lightgray, colback=white, size=fbox, boxrule=0.5pt, width=\dimexpr0.48\textwidth-2\fboxrule-2\fboxsep]
    \footnotesize
    \textbf{Input}: \\ 
Interviewer: What are your hopes for your daughter's future? \\ 
Respondent: What future? We are just surviving. She will just have to get married and that's it.\\ 

\textbf{Output}: \\ 
Interpretation: The respondent expresses a sense of helplessness regarding their daughter's future and does not express any dreams or plans for her future beyond getting married. This aligns with the “No Ambition” label. \\ 
Label: No Ambition
  \end{tcolorbox} &
  \vspace{0pt}
  \begin{tcolorbox}[colframe=lightgray, colback=white, size=fbox, boxrule=0.5pt, width=\dimexpr0.48\textwidth-2\fboxrule-2\fboxsep]
    \footnotesize
    \textbf{Input}: \\ 
Interviewer: What are your plans for your child's education? \\ 
Respondent: I can't afford to send him to school. He'll have to work like the rest of us.\\ 

\textbf{Output}: \\ 
Interpretation: The respondent expresses a sense of helplessness and lack of plans for their child's education, indicating a lack of ambition for the child's future. This aligns with the “No Ambition” label. \\ 
Label: No Ambition
  \end{tcolorbox} \\
  \vspace{0pt}
  \begin{tcolorbox}[colframe=lightgray, colback=white, size=fbox, boxrule=0.5pt, width=\dimexpr0.48\textwidth-2\fboxrule-2\fboxsep]
    \footnotesize
    \textbf{Input}: \\ 
Interviewer: What are your dreams for your son's future? \\ 
Respondent: I want him to become a doctor and serve the community.\\ 

\textbf{Output}: \\ 
Interpretation: The respondent clearly expresses a dream for their child's future, indicating that they have ambitions for the child. This does not align with the “No Ambition” label. \\ 
Label: Not Applicable
  \end{tcolorbox} &
  \vspace{0pt}
  \begin{tcolorbox}[colframe=lightgray, colback=white, size=fbox, boxrule=0.5pt, width=\dimexpr0.48\textwidth-2\fboxrule-2\fboxsep]
    \footnotesize
    \textbf{Input}: \\ 
Interviewer: How do you see your child's future? \\ 
Respondent: I hope he will be able to get a good education and have a successful career.\\ 

\textbf{Output}: \\ 
Interpretation: The respondent expresses hopes for their child's future, indicating that they have ambitions for the child. This does not align with the “No Ambition” label. \\ 
Label: Not Applicable
  \end{tcolorbox}
\end{tabular}

\clearpage

\vspace{1em}

\begin{center}
\textbf{Reliance on God}
\end{center}

\noindent\textbf{System prompt}

\vspace{0.5em}

\noindent
\begin{tcolorbox}[colframe=lightgray, colback=white, size=fbox, boxrule=0.5pt, width=\textwidth]
\footnotesize
Your task involves interpreting and analyzing transcripts of interviews conducted with parents, focusing on their reliance on divine intervention or faith in God to support their children's education and future aspirations. Specifically, your role is to discern whether the parent explicitly expresses full or partial reliance on God to fulfill their dreams for their children or if they currently rely on God's will in their lives. \\ 

Assign the “Reliance on God” label only if the parent explicitly articulates a full or partial dependence on God to accomplish their aspirations for their children's future or if they demonstrate a current reliance on divine will. \\ 

If there is no such direct reference to reliance on God in the parent's statement, assign the “Not Applicable” label to indicate that the topic of reliance on divine intervention was not addressed.
\end{tcolorbox}

\vspace{1em}

\noindent\textbf{Few-shot examples}

\vspace{0.5em}

\noindent
\begin{tabular}{@{} m{0.48\textwidth} m{0.48\textwidth} @{}}
  \vspace{0pt}
  \begin{tcolorbox}[colframe=lightgray, colback=white, size=fbox, boxrule=0.5pt, width=\dimexpr0.48\textwidth-2\fboxrule-2\fboxsep]
    \footnotesize
    \textbf{Input}: \\ 
Interviewer: What are your hopes for your children's future? \\ 
Respondent: I hope they will do well in their studies and become successful. But ultimately, it's all up to God's will.\\ 

\textbf{Output}: \\ 
Interpretation: The respondent clearly states their hopes for their children's future, but also expresses a reliance on God's will to determine the outcome. This demonstrates a partial reliance on divine intervention for their children's future. \\ 
Label: Reliance on God
  \end{tcolorbox} &
  \vspace{0pt}
  \begin{tcolorbox}[colframe=lightgray, colback=white, size=fbox, boxrule=0.5pt, width=\dimexpr0.48\textwidth-2\fboxrule-2\fboxsep]
    \footnotesize
    \textbf{Input}: \\ 
Interviewer: Do you have any dreams for your daughter's future? \\ 
Respondent: Yes, I dream that she will become a doctor. But we leave everything in God's hands. If it's His will, it will happen.\\ 

\textbf{Output}: \\ 
Interpretation: The respondent expresses a dream for their daughter's future but also mentions that the fulfillment of this dream is dependent on God's will. This shows a reliance on divine intervention for their child's future. \\ 
Label: Reliance on God
  \end{tcolorbox} \\
  \vspace{0pt}
  \begin{tcolorbox}[colframe=lightgray, colback=white, size=fbox, boxrule=0.5pt, width=\dimexpr0.48\textwidth-2\fboxrule-2\fboxsep]
    \footnotesize
    \textbf{Input}: \\ 
Interviewer: How do you plan to support your children's education? \\ 
Respondent: I work hard every day to provide for their needs. I believe in the power of hard work and dedication.\\ 

\textbf{Output}: \\ 
Interpretation: The respondent talks about supporting their children's education through their own efforts and does not mention any reliance on God or divine intervention. Therefore, the “Reliance on God” label does not apply in this case. \\ 
Label: Not Applicable
  \end{tcolorbox} &
  \vspace{0pt}
  \begin{tcolorbox}[colframe=lightgray, colback=white, size=fbox, boxrule=0.5pt, width=\dimexpr0.48\textwidth-2\fboxrule-2\fboxsep]
    \footnotesize
    \textbf{Input}: \\ 
Interviewer: What are your plans for your son's education? \\ 
Respondent: I just want him to do his best. The rest is up to him.\\ 

\textbf{Output}: \\ 
Interpretation: The respondent expresses a desire for their son to do well in his education, but does not mention any reliance on God or divine intervention. Therefore, the “Reliance on God” label does not apply in this case. \\ 
Label: Not Applicable
  \end{tcolorbox}
\end{tabular}

\clearpage

\vspace{1em}

\begin{center}
\textbf{Religious Aspirations}
\end{center}

\noindent\textbf{System prompt}

\vspace{0.5em}

\noindent
\begin{tcolorbox}[colframe=lightgray, colback=white, size=fbox, boxrule=0.5pt, width=\textwidth]
\footnotesize
Your task involves interpreting and analyzing transcripts of interviews conducted with parents, focusing on their aspirations concerning their children's future education and careers solely in the context of religion, without implying any monetary incentives. Specifically, your role is to discern whether the parent explicitly expresses a desire for their child to pursue a religious path. \\ 

Assign the “Religious Aspiration” label only if the parent explicitly articulates aspirations for their child that include becoming a religious scholar, Hafiz, attending a madrassa, learning Arabic, Quran reading, Islamic covering, regular prayer, working in Islamic banks, or being a great Islamic scholar. It is crucial not to assign this label if the child is already engaged in any of the mentioned activities, as current religious activities are not considered future aspirations. \\ 

If there is no such direct reference to religious aspirations in the parent's statement, or if there is a reference to secular aspirations, assign the “Not Applicable” label to indicate that the topic of religious aspirations was not addressed.
\end{tcolorbox}

\vspace{1em}

\noindent\textbf{Few-shot examples}

\vspace{0.5em}

\noindent
\begin{tabular}{@{} m{0.48\textwidth} m{0.48\textwidth} @{}}
  \vspace{0pt}
  \begin{tcolorbox}[colframe=lightgray, colback=white, size=fbox, boxrule=0.5pt, width=\dimexpr0.48\textwidth-2\fboxrule-2\fboxsep]
    \footnotesize
    \textbf{Input}: \\ 
Interviewer: What are your aspirations for your son's future? \\ 
Respondent: I hope that he will become a great Islamic scholar. I want him to learn Arabic and read the Quran regularly.\\ 

\textbf{Output}: \\ 
Interpretation: The parent explicitly expresses a desire for their child to pursue a path in religious education, specifically mentioning becoming an Islamic scholar, learning Arabic, and reading the Quran regularly. Therefore, this conversation aligns with the “Religious Aspiration” label. \\ 
Label: Religious Aspiration
  \end{tcolorbox} &
  \vspace{0pt}
  \begin{tcolorbox}[colframe=lightgray, colback=white, size=fbox, boxrule=0.5pt, width=\dimexpr0.48\textwidth-2\fboxrule-2\fboxsep]
    \footnotesize
    \textbf{Input}: \\ 
Interviewer: What do you envision for your child's future? \\ 
Respondent: I would like my son to become a Hafiz. It's a great honor in our religion.\\ 

\textbf{Output}: \\ 
Interpretation: The parent explicitly expresses a desire for their child to become a Hafiz, which is a religious aspiration as per the instruction. Therefore, this conversation aligns with the “Religious Aspiration” label. \\ 
Label: Religious Aspiration
  \end{tcolorbox} \\
  \vspace{0pt}
  \begin{tcolorbox}[colframe=lightgray, colback=white, size=fbox, boxrule=0.5pt, width=\dimexpr0.48\textwidth-2\fboxrule-2\fboxsep]
    \footnotesize
    \textbf{Input}: \\ 
Interviewer: Do you have any specific dreams for your daughter? \\ 
Respondent: I want her to excel in her studies and become a successful engineer.\\ 

\textbf{Output}: \\ 
Interpretation: While the parent does express aspirations for their child's future, these aspirations are not religious in nature. The parent hopes for their child to become an engineer, which doesn't involve any religious activities as per the instruction. Therefore, this conversation should be labeled as “Not Applicable”. \\ 
Label: Not Applicable
  \end{tcolorbox} &
  \vspace{0pt}
  \begin{tcolorbox}[colframe=lightgray, colback=white, size=fbox, boxrule=0.5pt, width=\dimexpr0.48\textwidth-2\fboxrule-2\fboxsep]
    \footnotesize
    \textbf{Input}: \\ 
Interviewer: What are your hopes for your children's future careers? \\ 
Respondent: I want them to have successful careers. I hope my daughter becomes a doctor and my son becomes a lawyer.\\ 

\textbf{Output}: \\ 
Interpretation: While the parent does express aspirations for their children's future, these aspirations do not involve any religious activities or pursuits as per the instruction. Therefore, this conversation should be labeled as “Not Applicable”. \\ 
Label: Not Applicable
  \end{tcolorbox}
\end{tabular}

\clearpage

\vspace{1em}

\begin{center}
\textbf{Secular Aspirations}
\end{center}

\noindent\textbf{System prompt}

\vspace{0.5em}

\noindent
\begin{tcolorbox}[colframe=lightgray, colback=white, size=fbox, boxrule=0.5pt, width=\textwidth]
\footnotesize
Your task involves interpreting and analyzing transcripts of interviews conducted with parents, focusing on their aspirations concerning their children's future education, career, and personal development without religious context. Specifically, your role is to discern whether the parent explicitly expresses a desire for their child to develop positive character traits, moralistic values, or general well-being. \\ 

Assign the “Secular Aspiration” label only if the parent explicitly articulates aspirations for their child that include becoming a good person, being responsible, having good manners, living a healthy life, achieving professional success, or having a good family. \\ 

If there is no such direct reference to secular aspirations in the parent's statement, or if there is a reference to religious aspirations, assign the “Not Applicable” label to indicate that the topic of secular aspirations was not addressed.
\end{tcolorbox}

\vspace{1em}

\noindent\textbf{Few-shot examples}

\vspace{0.5em}

\noindent
\begin{tabular}{@{} m{0.48\textwidth} m{0.48\textwidth} @{}}
  \vspace{0pt}
  \begin{tcolorbox}[colframe=lightgray, colback=white, size=fbox, boxrule=0.5pt, width=\dimexpr0.48\textwidth-2\fboxrule-2\fboxsep]
    \footnotesize
    \textbf{Input}: \\ 
Interviewer: Can you share your aspirations for your children? \\ 
Respondent: I want them to be kind, respectful and educated individuals. I hope they lead a healthy life and contribute positively to society.\\ 

\textbf{Output}: \\ 
Interpretation: The respondent explicitly expresses a desire for their children to develop positive traits such as kindness and respectfulness. They also mention the importance of education and leading a healthy life, which are secular aspirations for their children. Therefore, the dialogue should be assigned the “Secular Aspiration” label. \\ 
Label: Secular Aspiration
  \end{tcolorbox} &
  \vspace{0pt}
  \begin{tcolorbox}[colframe=lightgray, colback=white, size=fbox, boxrule=0.5pt, width=\dimexpr0.48\textwidth-2\fboxrule-2\fboxsep]
    \footnotesize
    \textbf{Input}: \\ 
Interviewer: What are your dreams for your children? \\ 
Respondent: My dream for my children is that they grow up to be responsible, hardworking, and successful in their careers. I also hope they have a happy family life.\\ 

\textbf{Output}: \\ 
Interpretation: The respondent clearly articulates secular aspirations for their children, which include becoming responsible and hardworking individuals, achieving professional success, and having a good family life. Therefore, the dialogue should be assigned the “Secular Aspiration” label. \\ 
Label: Secular Aspiration
  \end{tcolorbox} \\
  \vspace{0pt}
  \begin{tcolorbox}[colframe=lightgray, colback=white, size=fbox, boxrule=0.5pt, width=\dimexpr0.48\textwidth-2\fboxrule-2\fboxsep]
    \footnotesize
    \textbf{Input}: \\ 
Interviewer: What kind of future do you envision for your children? \\ 
Respondent: I just hope they're happy, whatever they choose to do.\\ 

\textbf{Output}: \\ 
Interpretation: While the respondent expresses a desire for their children's happiness, they do not explicitly state any secular aspirations such as developing positive character traits or achieving professional success. Therefore, the dialogue should be assigned the “Not Applicable” label. \\ 
Label: Not Applicable
  \end{tcolorbox} &
  \vspace{0pt}
  \begin{tcolorbox}[colframe=lightgray, colback=white, size=fbox, boxrule=0.5pt, width=\dimexpr0.48\textwidth-2\fboxrule-2\fboxsep]
    \footnotesize
    \textbf{Input}: \\ 
Interviewer: What kind of person would you like your child to become and how do you want them to live their life? \\ 
Respondent: I hope they will follow our religious teachings, pray regularly, and become a great Islamic scholar.\\ 

\textbf{Output}: \\ 
Interpretation: The respondent discusses faith-based aspirations for their child, such as following religious teachings, praying regularly, and becoming a great Islamic scholar. Therefore, the dialogue should be assigned the “Not Applicable” label. \\ 
Label: Not Applicable
  \end{tcolorbox}
\end{tabular}

\clearpage

\vspace{1em}

\begin{center}
\textbf{Vocational Training}
\end{center}

\noindent\textbf{System prompt}

\vspace{0.5em}

\noindent
\begin{tcolorbox}[colframe=lightgray, colback=white, size=fbox, boxrule=0.5pt, width=\textwidth]
\footnotesize
Your task involves interpreting and analyzing transcripts of interviews conducted with parents, focusing on their aspirations concerning their children's future vocational training. Specifically, your role is to discern whether the parent explicitly expresses a desire or intention related to their child pursuing vocational training or learning a specific skill in the context of ambition, such as tailoring, handicrafts, or sewing. The label should only be assigned if training, teaching, or learning is mentioned in the context of ambition. \\ 

Assign the “Vocational Training” label only if the parent explicitly articulates a desire or intention for their child to undergo vocational training or learn a particular skill in the future as part of their ambition. \\ 

If there is no such direct reference to vocational training in the parent's statement, assign the “Not Applicable” label to indicate that the topic of vocational training was not addressed.
\end{tcolorbox}

\vspace{1em}

\noindent\textbf{Few-shot examples}

\vspace{0.5em}

\noindent
\begin{tabular}{@{} m{0.48\textwidth} m{0.48\textwidth} @{}}
  \vspace{0pt}
  \begin{tcolorbox}[colframe=lightgray, colback=white, size=fbox, boxrule=0.5pt, width=\dimexpr0.48\textwidth-2\fboxrule-2\fboxsep]
    \footnotesize
    \textbf{Input}: \\ 
Interviewer: What are your aspirations for your youngest daughter? \\ 
Respondent: I want her to learn sewing. It's a skill that can help her in the future.\\ 

\textbf{Output}: \\ 
Interpretation: The parent explicitly mentions a desire for their daughter to learn sewing, a vocational skill. Therefore, the dialogue should be labeled as 'Vocational Training'. \\ 
Label: Vocational Training
  \end{tcolorbox} &
  \vspace{0pt}
  \begin{tcolorbox}[colframe=lightgray, colback=white, size=fbox, boxrule=0.5pt, width=\dimexpr0.48\textwidth-2\fboxrule-2\fboxsep]
    \footnotesize
    \textbf{Input}: \\ 
Interviewer: How do you envision your child's future? \\ 
Respondent: Well, I believe in letting him choose his own path. But I do encourage him to learn carpentry. It's a useful skill to have.\\ 

\textbf{Output}: \\ 
Interpretation: The parent expresses a desire for their child to learn carpentry, a vocational skill. Therefore, the dialogue should be labeled as 'Vocational Training'. \\ 
Label: Vocational Training
  \end{tcolorbox} \\
  \vspace{0pt}
  \begin{tcolorbox}[colframe=lightgray, colback=white, size=fbox, boxrule=0.5pt, width=\dimexpr0.48\textwidth-2\fboxrule-2\fboxsep]
    \footnotesize
    \textbf{Input}: \\ 
Interviewer: Do you have any specific plans for your son's future? \\ 
Respondent: I hope he continues his studies and becomes a doctor one day.\\ 

\textbf{Output}: \\ 
Interpretation: Although the parent expresses a desire for their child's future, there is no mention of vocational training or learning a specific skill. Therefore, the dialogue should be labeled as 'Not Applicable'. \\ 
Label: Not Applicable
  \end{tcolorbox} &
  \vspace{0pt}
  \begin{tcolorbox}[colframe=lightgray, colback=white, size=fbox, boxrule=0.5pt, width=\dimexpr0.48\textwidth-2\fboxrule-2\fboxsep]
    \footnotesize
    \textbf{Input}: \\ 
Interviewer: What are your dreams for your children? \\ 
Respondent: I just want them to be happy and healthy. That's all that matters.\\ 

\textbf{Output}: \\ 
Interpretation: The parent expresses a general wish for their children's well-being, but there is no explicit mention of vocational training or learning a specific skill. Therefore, the dialogue should be labeled as 'Not Applicable'. \\ 
Label: Not Applicable
  \end{tcolorbox}
\end{tabular}

\subsection{LLM prompts for Data Augmentation} \label{app:prompts_augmentation}

We directed the LLM with the following “system prompt”:
\begin{tcolorbox}[colframe=lightgray, colback=white, size=fbox, boxrule=0.5pt]
\footnotesize
Revise a given dialogue between an interviewer and respondent while maintaining the core meaning. \\
Additionally, you may alter names, places, and other identifying information while preserving the core meaning of the dialogue. \\ \\
Create a single revision of the dialogue in the following format: \\ 
\hspace{10mm}Interviewer: (question/statement by the interviewer) 
\hspace{10mm}Respondent: (answer by the respondent)
\end{tcolorbox}

\section{iQual Appendix} \label{app:iqual}

This Appendix summarises the key features of the iQual approach to supervised NLP for qualitative analysis. The method is set out in more detail in \cite{ashwin2022method} and available through a Python package at \href{https://github.com/worldbank/iQual}{https://github.com/worldbank/iQual}.

\subsection{Summary of Approach}

The overall goal of our iQual is to use our subset of annotated interviews to provide reliable annotations for the remainder of the sample. Broadly, we do this by training a series of classifier models on our annotated set and then using this model to predict annotations for the unannotated set. We train and predict for each  annotations separately, so the model for Religious Aspiration will be trained and make its predictions separately from the model for Secular Aspiration. As the qualitative annotations are defined at the level of question-answer pairs (QA), we can represent each annotation as a binary classification problem at the QA level.
 
\begin{figure}[H]
    \caption{Methodology}
    \label{fig:flowchart}
    \centering
    \includegraphics[trim=30.0 70.0 30.0 70.0, clip, width = \linewidth ]{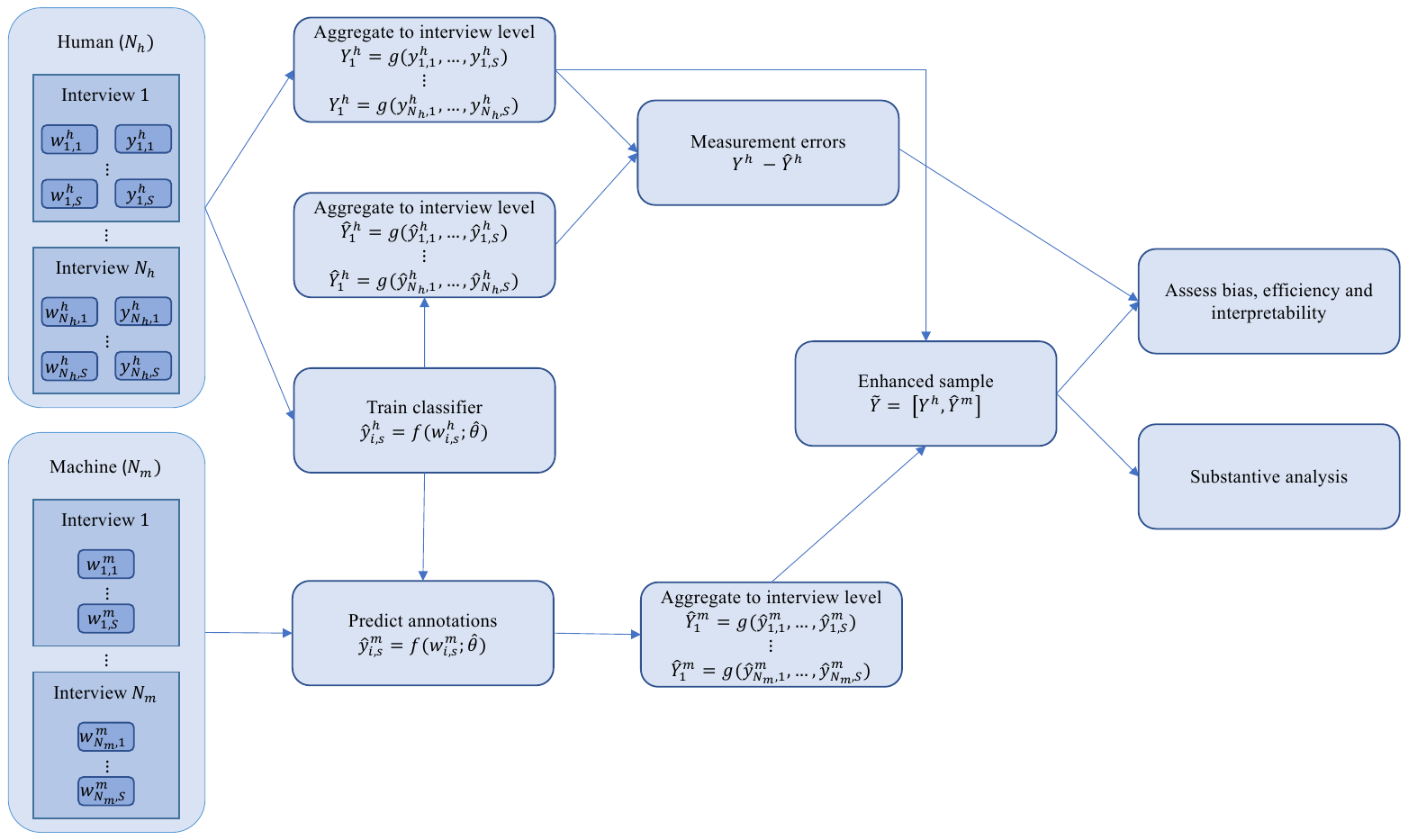}
\end{figure}

Figure \ref{fig:flowchart}, also shown in \cite{ashwin2022method}, illustrates our overall methodology for a single annotation. On the left hand side we see a "human" sample of size $N_h$, in which interviews include both text $w$ and annotations $y$, and a "machine" sample in which interviews include only the text. As annotations are defined at the QA (question-answer pair) level, so we represent $w^h_{i,s}$ as the $s$th QA in interview $i$ in the human sample, with $y^h_{i,s}$ being the binary annotation on that QA. In other words, if the annotation Religious Aspiration, $y^h_{i,s}$ will be equal to one if that QA has been annotated as displaying religious aspirations, and will be zero otherwise. 

We then train some classifier $f()$ parameterised by $\theta$ to predict $y^h_{i,s}$ based on the QA text $w^h_{i,s}$. As we will discuss below, there are many options for both the classifier we can use here, as well as how to represent the text numerically. A key point here is that the text representation must by full unsupervised - i.e. we do not use any information about $y$ or any further information about the interview subject when creating a numerical representation of the text. The text representation, classifier and a variety of hyperparameters are chosen using k-fold cross-validation. Given this trained classifier we can then predict annotations a the QA level for our unannotated "machine" sample. This gives us the predicted annotations $\hat{y}^m_{i,s}$. 
 
We then aggregate the QA level annotations to the interview level using aggregation function $g()$. The choice of this aggregation function is at least in part a substantive question that depends on the research question. For example, if we take the mean value of $y$ across QA pairs for each interview this gives us a measure of the intensity with which this concept comes up. On the other hand, if we take the maximum value across the interview this gives us a measure of interviews in which this concept comes up at least once. We perform this aggregation for both the observed human annotations $Y^h$, the "in-sample" predicted human annotations $\hat{Y}^h$ and the "out-of-sample" predicted machine annotations $\hat{Y}^m$. The predicted annotations for the human sample can then by used to assess the measurement errors introduced by the model. For the quantification of measurement errors, we also make use of bootstrapping. The observed human annotations and machine annotations are then combined to give an enhanced sample $\tilde{Y}$. Once we have verified that the enhancement does indeed add value, we proceed with substantive analysis.

\subsection{Choices}

There are many possible options for the numerical representation of the text representation $w$, the classifier $f()$ and the aggregation function $g()$. We fix the aggregation function as taking the mean across QA pairs and we use cross-validation to select the text representation and the classifier. As we train the classifier for each annotation independently, this allows for the fact that a different classification model of text representation may be optimal for different annotations. THe following gives an exhaustive list of the text representations, models and hyperparameters that are selected over during cross validation. In addition to these options we also cross-validate whether the question is included along with the answer in the text representation.

\begin{table}[H] \centering 
	\caption{Statistical methods for text vectorization}
	\scriptsize 
	\begin{tabularx}{\textwidth}{p{3cm}XXX}
		\hline
		\\
		\textbf{Method Name} & \textbf{Description} & \textbf{Hyperparameters (Options)} & \textbf{Hyperparameters (Used)} \\
		\\
		\hline
		\hline
\\
TfidfVectorizer & TfidfVectorizer is a method for converting text into numerical representations, specifically term frequency-inverse document frequency (TF-IDF) vectors. It counts the frequency of words in a document and down-weights the importance of commonly used words. This can be useful for text classification tasks, as it allows the classifier to focus on the words that are most distinctive to a particular document. &
		\begin{itemize}[leftmargin=*, noitemsep,topsep=0pt]
			\setlength\itemsep{0.5em}
\item \textbf{ngram\_range}: The range of n-grams to consider when creating the vocabulary.
\item \textbf{min\_df}: The minimum number of documents a word must be in to be included in the vocabulary.
\item \textbf{max\_df}: The maximum number of documents a word can be in to be included in the vocabulary.
\item \textbf{max\_features}: The maximum number of words to keep in the vocabulary, based on word frequency.
\item \textbf{use\_idf}: A boolean flag indicating whether to use the inverse-document-frequency weighting.
\item \textbf{norm}: The type of normalization to apply to the vectors.
\item \textbf{smooth\_idf}: A boolean flag indicating whether to smooth the idf values.
\item \textbf{sublinear\_tf}: A boolean flag indicating whether to apply sublinear scaling to the term frequency.
\end{itemize}
&
		\begin{itemize}[leftmargin=*, noitemsep,topsep=0pt]
			\setlength\itemsep{0.5em}
\item \textbf{max\_features}: The maximum number of words to keep in the vocabulary based on word frequency. \textbf{[1000, 10000]}
\item \textbf{ngram\_range}: The lower and upper boundary of the range of n-values for different word n-grams to be extracted. \textbf{ \{ (1,1), (1,2), (1,3) \}}
\end{itemize}
\\ 
\hline 
\\
CountVectorizer & CountVectorizer is a method for converting text into numerical representations, specifically a sparse matrix of word counts. It counts the frequency of words in a document and does not down-weight the importance of commonly used words. This can be useful for text classification tasks, as it allows the classifier to consider all words equally, rather than down-weighting the importance of commonly used words. &

		\begin{itemize}[leftmargin=*, noitemsep,topsep=0pt]
			\setlength\itemsep{0.5em}
\item \textbf{ngram\_range}: The range of n-grams to consider when creating the vocabulary.
\item \textbf{min\_df}: The minimum number of documents a word must be in to be included in the vocabulary.
\item \textbf{max\_df}: The maximum number of documents a word can be in to be included in the vocabulary.
\item \textbf{max\_features}: The maximum number of words to keep in the vocabulary, based on word frequency.
\item \textbf{binary}: A boolean flag indicating whether to create binary vectors, with 0/1 values indicating the presence/absence of a word in a document.
\end{itemize} & 
		\begin{itemize}[leftmargin=*, noitemsep,topsep=0pt]
			\setlength\itemsep{0.5em}
\item \textbf{max\_features}: The maximum number of words to keep in the vocabulary, based on word frequency. \textbf{[1000, 10000]}
\item \textbf{ngram\_range}: The lower and upper boundary of the range of n-values for different word n-grams to be extracted. \textbf{\{ (1,1), (1,2), (1,3) \}}
\item \textbf{binary}: Whether to use binary or frequency counts. \textbf{\{True, False\}}
\end{itemize}
\\
\hline
\\
\end{tabularx}
\end{table}

\begin{table}[H] \centering
  \caption{Pre-trained embeddings for text vectorization} 
  \scriptsize
\begin{tabularx}{\textwidth}{p{4cm}p{2cm}X}
\\
  \hline
  \\
\textbf{Model Name} & \textbf{Dimensions} & \textbf{Description} \\
\\ 
\hline 
\hline
\\
all-mpnet-base-v2 &  768 & This a pre-trained language understanding model that combines the advantages of masked language modeling (MLM) and permuted language modeling (PLM) to address the limitations of both methods. It leverages the dependency among predicted tokens through PLM and takes auxiliary position information as input to make the model see a full sentence, reducing the position discrepancy between pre-training and fine-tuning. This model was pre-trained on a large-scale dataset and generates a vector of 768 dimensions. \\
\\ 
\hline 
\\
all-roberta-large-v1 &  1024 & This is a pre-trained language understanding model with a vector representation of 1024 dimensions. It was developed as an improvement upon the BERT model and was trained using the masked language modeling (MLM) objective. It has achieved strong performance on natural language processing tasks and can be fine-tuned on labeled datasets for specific tasks such as classification or language translation. \\
\\ 
\hline 
\\
average\_word\_embeddings \_glove.6B.300d &  300 &  This is a method for converting text into numerical representations, specifically word embeddings. It uses a pre-trained GloVe model to generate 300-dimensional vector representations for each word in a document, and then averages these vectors to create a single representation for the entire document. This can be useful for text classification tasks, as it allows the classifier to consider the semantic relationships between words, rather than just their frequencies. \\
\\ 
\hline
\\
distiluse-base-multilingual-cased-v2 &  512 &  This is a pre-trained language understanding model that maps text into a 512-dimensional vector representation. It is a smaller and faster version of the popular transformer model, BERT, and has been trained on a large multilingual dataset, allowing it to process text in multiple languages. It has also been cased, meaning it can distinguish between upper and lower case letters. This model is useful for natural language processing tasks such as language translation and text classification, and can be fine-tuned on labeled datasets for specific tasks. \\
\\ 
\hline 
\\ 
\end{tabularx}
\end{table}

\begin{table}[H] \centering
\caption{Classifier Options I}
  \scriptsize
  \setlength{\tabcolsep}{0.1cm}
  \begin{tabularx}{\textwidth}{p{3cm}p{3cm}XX}
  \\
  \hline
  \\
\textbf{Method} & \textbf{Description} & \textbf{Hyperparameters (Options)} & \textbf{Hyperparameters (Used)} \\ 
\\ 
\hline 
\hline
\\

LogisticRegression & This is a linear classifier that uses a logistic function to predict the probability of a sample belonging to a particular class. It is commonly used for binary classification tasks, but can also be used for multi-class classification by implementing a one-versus-rest approach. & \begin{itemize}[leftmargin=*, noitemsep,topsep=0pt] \setlength\itemsep{0.1em}
\item 	C: The inverse of the regularization strength, with higher values indicating less regularization.
\item 	penalty: The type of regularization to use, either L1 or L2.
\item 	fit\_intercept: A boolean flag indicating whether to fit an intercept term.
\item 	tol: The tolerance for stopping criteria.
\item 	intercept\_scaling: The scaling of the intercept term, if it is being fitted.
\item 	class\_weight: The class weights to use for unbalanced classes.
\item 	max\_iter: The maximum number of iterations for the optimization algorithm.
\end{itemize} & \begin{itemize}[leftmargin=*, noitemsep,topsep=0pt] \setlength\itemsep{0.1em}

\item 	penalty: The type of regularization to use: L1 or L2.
\item 	C: Inverse of regularization strength. [0.00002, 10000]
\end{itemize}
\\ \hline \\
SGDClassifier & This is a linear classifier that uses stochastic gradient descent to learn the parameters of the model. The modified huber loss function is a smooth approximation of the hinge loss, which is commonly used for linear classification tasks. & \begin{itemize}[leftmargin=*, noitemsep,topsep=0pt] \setlength\itemsep{0.1em}

\item 	loss: The loss function to use, with options such as "hinge", "log", "modified\_huber", "squared\_hinge", and "perceptron".
\item 	penalty: The type of regularization to use, with options such as L1, L2, "elasticnet", and "none".
\item 	alpha: The regularization strength, with higher values indicating stronger regularization.
\item 	l1\_ratio: The proportion of L1 regularization to use in the elasticnet penalty.
\item 	tol: The tolerance for the stopping criteria.
\item 	learning\_rate: The learning rate for the optimization algorithm, with options such as "constant", "optimal", and "invscaling".
\item 	eta0: The initial learning rate for the "constant" and "invscaling" learning rate schedules.
\item 	power\_t: The exponent for the "invscaling" learning rate schedule.
\end{itemize} & \begin{itemize}[leftmargin=*, noitemsep,topsep=0pt] \setlength\itemsep{0.1em}

\item 	loss: The loss function to use. ("modified\_huber")	
\item 	penalty: The type of regularization to use: L1 or L2.
\item 	learning\_rate: The learning rate schedule to use. ("optimal")
\item 	alpha: The constant that multiplies the regularization term. [0.00002, 1000]
\end{itemize}
\\ \hline \\
\end{tabularx}
\end{table}

\begin{table}[H] \centering
\caption{Classifier Options II}
  \scriptsize
  \setlength{\tabcolsep}{0.1cm}
  \begin{tabularx}{\textwidth}{p{3cm}p{3cm}XX}
  \\
  \hline
  \\
\textbf{Method} & \textbf{Description} & \textbf{Hyperparameters (Options)} & \textbf{Hyperparameters (Used)} \\ 
\\ 
\hline 
\hline
\\
RandomForestClassifier & This is an ensemble classifier that uses multiple decision trees to make predictions. It randomly selects a subset of features to consider at each split in the tree, which helps to reduce overfitting and improve the generalization of the model. & \begin{itemize}[leftmargin=*, noitemsep,topsep=0pt] \setlength\itemsep{0.1em}

\item 	n\_estimators: The number of decision trees in the forest.
\item 	criterion: The function to measure the quality of a split, with options such as "gini" and "entropy".
\item 	max\_depth: The maximum depth of the decision tree.
\item 	min\_samples\_split: The minimum number of samples required to split an internal node.
\item 	min\_samples\_leaf: The minimum number of samples required to be at a leaf node.
\item 	min\_weight\_fraction\_leaf: The minimum weighted fraction of the sum total of weights required to be at a leaf node.
\item 	max\_features: The number of features to consider when looking for the best split.
\item 	max\_leaf\_nodes: The maximum number of leaf nodes in the tree.
\item 	min\_impurity\_decrease: The minimum decrease in impurity required to split the node.
\item 	bootstrap: A boolean flag indicating whether to use bootstrap samples when building the trees.
\item 	oob\_score: A boolean flag indicating whether to use out-of-bag samples to estimate the generalization error.
\end{itemize} & \begin{itemize}[leftmargin=*, noitemsep,topsep=0pt] \setlength\itemsep{0.1em}
\item 	n\_estimators: The number of trees in the forest. [100, 1000]
\item 	max\_depth: The maximum depth of the tree. [10, 100]
\end{itemize}
\\ \hline \\
DecisionTreeClassifier & This is a classifier that uses a tree structure to make decisions based on the features of a sample. At each node in the tree, the classifier considers a single feature and splits the data based on the value of that feature. The final decision is made based on the path taken through the tree. & \begin{itemize}[leftmargin=*, noitemsep,topsep=0pt] \setlength\itemsep{0.1em}

\item 	criterion: The function to measure the quality of a split, with options such as "gini" and "entropy".
\item 	splitter: The strategy to use when searching for a split, with options such as "best" and "random".
\item 	max\_depth: The maximum depth of the tree.
\item 	min\_samples\_split: The minimum number of samples required to split an internal node.
\item 	min\_samples\_leaf: The minimum number of samples required to be at a leaf node.
\item 	min\_weight\_fraction\_leaf: The minimum weighted fraction of the sum total of weights required to be at a leaf node.
\item 	max\_features: The number of features to consider when looking for the best split.
\item 	max\_leaf\_nodes: The maximum number of leaf nodes in the tree.
\item 	min\_impurity\_decrease: The minimum decrease in impurity required to split the node.
\end{itemize} & \begin{itemize}[leftmargin=*, noitemsep,topsep=0pt] \setlength\itemsep{0.1em}

\item 	max\_depth: The maximum depth of the tree. [5, 100]
\item 	min\_impurity\_decrease: A node will be split if this split induces a decrease of the impurity greater than or equal to this value. [0.00002,10000]
\end{itemize}
\\ \hline \\

\end{tabularx}
\end{table}

\begin{table}[H] \centering
\caption{Classifier Options III}
  \scriptsize
  \setlength{\tabcolsep}{0.1cm}
  \begin{tabularx}{\textwidth}{p{3cm}p{3cm}XX}
  \\
  \hline
  \\
\textbf{Method} & \textbf{Description} & \textbf{Hyperparameters (Options)} & \textbf{Hyperparameters (Used)} \\ 
\\ 
\hline 
\hline
\\
MLPClassifier & This is a classifier that uses a neural network with multiple layers to make predictions. It is commonly used for classification tasks and can handle both continuous and categorical data. The number of layers and the number of units in each layer can be adjusted to fit the complexity of the task. & \begin{itemize}[leftmargin=*, noitemsep,topsep=0pt] \setlength\itemsep{0.1em}

\item 	hidden\_layer\_sizes: The number of neurons in each hidden layer.
\item 	activation: The activation function to use, with options such as "identity", "logistic", "tanh", and "relu".
\item 	solver: The algorithm to use for optimization, with options such as "lbfgs", "sgd", and "adam".
\item 	alpha: The regularization strength, with higher values indicating stronger regularization.
\item 	batch\_size: The number of samples to use in each iteration of the optimization algorithm.
\item 	learning\_rate: The learning rate for the optimization algorithm, with options such as "constant", "invscaling", and "adaptive".
\item 	learning\_rate\_init: The initial learning rate for the "constant" and "invscaling" learning rate schedules.
\item 	power\_t: The exponent for the "invscaling" learning rate schedule.
\item 	max\_iter: The maximum number of iterations to run the optimization algorithm.
\item 	shuffle: A boolean flag indicating whether to shuffle the training data before each epoch.
\item 	tol: The tolerance for the stopping criteria.
\item 	warm\_start: A boolean flag indicating whether to reuse the solution of the previous call to fit.
\item 	momentum: The momentum for the optimization algorithm.
\item 	nesterovs\_momentum: A boolean flag indicating whether to use Nesterov's momentum.
\item 	early\_stopping: A boolean flag indicating whether to use early stopping to terminate the optimization early.
\item 	validation\_fraction: The fraction of the training data to use as validation data for early stopping.
\item 	beta\_1: The beta 1 parameter for the Adam optimization algorithm.
\end{itemize} & \begin{itemize}[leftmargin=*, noitemsep,topsep=0pt] \setlength\itemsep{0.1em}

\item 	hidden\_layer\_sizes: The ith element represents the number of neurons in the ith hidden layer. [(100,), (100, 100), (100, 100, 100)]
\item 	activation: Activation function for the hidden layer. ("tanh", "relu")
\item 	alpha: L2 penalty (regularization term) parameter. [0.01, 1]
\end{itemize}
\\    \hline 
\end{tabularx}
\end{table}

\begin{table}[H] \centering
\caption{Classifier Options III}
  \scriptsize
  \setlength{\tabcolsep}{0.1cm}
  \begin{tabularx}{\textwidth}{p{3cm}p{3cm}XX}
  \\
  \hline
  \\
\textbf{Method} & \textbf{Description} & \textbf{Hyperparameters (Options)} & \textbf{Hyperparameters (Used)} \\ 
\\ 
\hline 
\hline
\\
KNeighborsClassifier & This is a non-parametric classifier that uses the K nearest neighbors of a sample to make a prediction. It is commonly used for classification tasks and can handle both continuous and categorical data. The number of neighbors to consider (K) is a hyperparameter that can be adjusted to fit the complexity of the task. & \begin{itemize}[leftmargin=*, noitemsep,topsep=0pt] \setlength\itemsep{0.1em}

\item 	n\_neighbors: The number of neighbors to use when making a prediction.
\item 	weights: The weight function to use when making a prediction, with options such as "uniform" and "distance".
\item 	algorithm: The algorithm to use for finding the nearest neighbors, with options such as "brute" and "kd\_tree".
\item 	leaf\_size: The number of points at which to switch to a brute force search for the nearest neighbors.
\item 	p: The power parameter for the Minkowski distance metric.
\item 	metric: The distance metric to use, with options such as "euclidean", "manhattan", and "minkowski".
\item 	metric\_params: Additional parameters for the distance metric.
\end{itemize} & \begin{itemize}[leftmargin=*, noitemsep,topsep=0pt] \setlength\itemsep{0.1em}

\item 	n\_neighbors: Number of neighbors to use by default for kneighbors queries. [10,10000]
\item 	weights: weight function used in prediction. ("uniform", "distance")
\end{itemize}
\\ \hline \\
SVC & This is a classifier that uses a support vector machine (SVM) to find the optimal hyperplane to separate the different classes. It is commonly used for classification tasks and can handle both continuous and categorical data. The kernel function used to project the data into a higher dimensional space can be adjusted to fit the complexity of the task. & \begin{itemize}[leftmargin=*, noitemsep,topsep=0pt] \setlength\itemsep{0.1em}

\item 	C: The regularization strength, with higher values indicating stronger regularization.
\item 	kernel: The kernel to use for the decision function, with options such as "linear", "poly", "rbf", "sigmoid", and "precomputed".
\item 	degree: The degree of the polynomial kernel.
\item 	gamma: The kernel coefficient for the rbf, poly, and sigmoid kernels.
\item 	coef0: The independent term in the polynomial and sigmoid kernels.
\item 	shrinking: A boolean flag indicating whether to use the shrinking heuristic.
\item 	probability: A boolean flag indicating whether to enable probability estimates.
\item 	tol: The tolerance for the stopping criteria.
\item 	class\_weight: The class weights to use for unbalanced classes.
\item 	verbose: The level of verbosity in the output.
\item 	decision\_function\_shape: The shape of the decision function, with options such as "ovo" and "ovr".
\end{itemize} & \begin{itemize}[leftmargin=*, noitemsep,topsep=0pt] \setlength\itemsep{0.1em}

\item 	C: Penalty parameter C of the error term. [0.00001, -00]
\end{itemize}
\\    \hline 
\end{tabularx}
\end{table}

\section{Bias tests} \label{app:bias_tests}

\begin{table}[H]
   \caption{Bias tests for Aspiration: Religious}
   \centering
   \resizebox{\linewidth}{!}{%
   \begin{tabular}{lccccc}
      \tabularnewline \midrule \midrule
      Dependent Variables: & ChatGPT                & Llama213Bchat   & Llama213B     & iQualChatGPTaug & iQual\\  
      Model:               & (1)                    & (2)             & (3)           & (4)             & (5)\\  
      \midrule
      \emph{Variables}\\
      (Intercept)          & -0.0347$^{***}$        & -0.0716$^{***}$ & -0.0139       & -0.0032         & $-1.55\times 10^{-5}$\\    
                           & (0.0132)               & (0.0185)        & (0.0126)      & (0.0104)        & (0.0107)\\   
      data\_roundR3        & 0.0059                 & -0.0206         & -0.0233$^{*}$ & 0.0116          & 0.0248$^{**}$\\   
                           & (0.0127)               & (0.0178)        & (0.0121)      & (0.0100)        & (0.0103)\\   
      refugee              & -0.0101                & -0.0167         & -0.0005       & 0.0039          & -0.0044\\   
                           & (0.0083)               & (0.0116)        & (0.0079)      & (0.0065)        & (0.0067)\\   
      num\_child           & 0.0007                 & -0.0008         & 0.0008        & -0.0003         & -0.0018\\   
                           & (0.0021)               & (0.0029)        & (0.0020)      & (0.0017)        & (0.0017)\\   
      hh\_head\_sex        & -0.0076                & 0.0115          & 0.0003        & -0.0091         & -0.0009\\   
                           & (0.0075)               & (0.0105)        & (0.0072)      & (0.0059)        & (0.0061)\\   
      hh\_head\_age        & 0.0004                 & 0.0006          & -0.0004       & 0.0004          & 0.0004\\   
                           & (0.0003)               & (0.0005)        & (0.0003)      & (0.0003)        & (0.0003)\\   
      parent\_eduyears     & $1.25\times 10^{-5}$   & -0.0020$^{*}$   & -0.0004       & -0.0004         & $1.67\times 10^{-5}$\\    
                           & (0.0008)               & (0.0012)        & (0.0008)      & (0.0007)        & (0.0007)\\   
      parent\_reledu       & -0.0088                & -0.0231         & -0.0039       & 0.0073          & 0.0063\\   
                           & (0.0142)               & (0.0199)        & (0.0136)      & (0.0112)        & (0.0115)\\   
      eld\_sex             & -0.0052                & -0.0113         & -0.0012       & -0.0074         & -0.0087$^{*}$\\   
                           & (0.0058)               & (0.0081)        & (0.0055)      & (0.0046)        & (0.0047)\\   
      eld\_age             & $-9.15\times 10^{-5}$  & 0.0003          & 0.0007        & -0.0003         & -0.0006$^{*}$\\   
                           & (0.0005)               & (0.0007)        & (0.0004)      & (0.0004)        & (0.0004)\\   
      hh\_asset\_index     & -0.0022                & -0.0041         & 0.0002        & -0.0016         & -0.0015\\   
                           & (0.0024)               & (0.0033)        & (0.0023)      & (0.0019)        & (0.0019)\\   
      hh\_income           & 0.0008                 & 0.0027          & 0.0010        & -0.0009         & -0.0006\\   
                           & (0.0017)               & (0.0024)        & (0.0016)      & (0.0014)        & (0.0014)\\   
      int\_trauma\_exp     & 0.0019                 & 0.0010          & 0.0023$^{*}$  & -0.0004         & 0.0012\\   
                           & (0.0012)               & (0.0017)        & (0.0012)      & (0.0010)        & (0.0010)\\   
      \midrule
      \emph{Fit statistics}\\
      Observations         & 667                    & 667             & 667           & 667             & 667\\  
      R$^2$                & 0.01430                & 0.02935         & 0.01555       & 0.02448         & 0.02312\\  
      Adjusted R$^2$       & -0.00378               & 0.01154         & -0.00252      & 0.00659         & 0.00520\\  
      F-test               & 0.79090                & 1.6477          & 0.86071       & 1.3679          & 1.2898\\  
      F-test, p-value      & 0.66022                & 0.07445         & 0.58739       & 0.17634         & 0.21956\\  
      \midrule \midrule
      \multicolumn{6}{l}{\emph{IID standard-errors in parentheses}}\\
      \multicolumn{6}{l}{\emph{Signif. Codes: ***: 0.01, **: 0.05, *: 0.1}}\\
   \end{tabular}
    }
\end{table}

\begin{table}[H]
   \caption{Bias tests for Aspiration: Secular}
   \centering
   \resizebox{\linewidth}{!}{%
   \begin{tabular}{lccccc}
      \tabularnewline \midrule \midrule
      Dependent Variables: & ChatGPT               & Llama213Bchat         & Llama213B             & iQualChatGPTaug        & iQual\\  
      Model:               & (1)                   & (2)                   & (3)                   & (4)                    & (5)\\  
      \midrule
      \emph{Variables}\\
      (Intercept)          & -0.1790$^{***}$       & -0.0602$^{**}$        & -0.0812$^{***}$       & -0.0247                & -0.0113\\   
                           & (0.0301)              & (0.0243)              & (0.0242)              & (0.0152)               & (0.0137)\\   
      data\_roundR3        & 0.0877$^{***}$        & 0.0308                & -0.0205               & 0.0064                 & 0.0075\\   
                           & (0.0289)              & (0.0233)              & (0.0233)              & (0.0146)               & (0.0132)\\   
      refugee              & 0.0243                & 0.0054                & 0.0166                & 0.0137                 & 0.0052\\   
                           & (0.0188)              & (0.0152)              & (0.0152)              & (0.0095)               & (0.0086)\\   
      num\_child           & 0.0052                & 0.0054                & 0.0013                & 0.0039                 & 0.0012\\   
                           & (0.0048)              & (0.0039)              & (0.0038)              & (0.0024)               & (0.0022)\\   
      hh\_head\_sex        & 0.0130                & 0.0018                & -0.0039               & -0.0026                & -0.0008\\   
                           & (0.0171)              & (0.0138)              & (0.0138)              & (0.0086)               & (0.0078)\\   
      hh\_head\_age        & -0.0006               & $8.01\times 10^{-5}$  & $3.67\times 10^{-5}$  & 0.0002                 & $-4.79\times 10^{-6}$\\    
                           & (0.0007)              & (0.0006)              & (0.0006)              & (0.0004)               & (0.0003)\\   
      parent\_eduyears     & -0.0039$^{**}$        & -0.0030$^{*}$         & -0.0012               & -0.0018$^{*}$          & -0.0004\\   
                           & (0.0019)              & (0.0015)              & (0.0015)              & (0.0010)               & (0.0009)\\   
      parent\_reledu       & 0.0294                & 0.0085                & -0.0036               & -0.0015                & 0.0127\\   
                           & (0.0323)              & (0.0261)              & (0.0261)              & (0.0163)               & (0.0147)\\   
      eld\_sex             & 0.0043                & -0.0086               & 0.0100                & -0.0008                & -0.0012\\   
                           & (0.0132)              & (0.0106)              & (0.0106)              & (0.0067)               & (0.0060)\\   
      eld\_age             & $-9.1\times 10^{-5}$  & -0.0006               & 0.0004                & $-6.99\times 10^{-5}$  & -0.0001\\   
                           & (0.0011)              & (0.0009)              & (0.0009)              & (0.0005)               & (0.0005)\\   
      hh\_asset\_index     & -0.0026               & -0.0024               & -0.0008               & 0.0046$^{*}$           & 0.0028\\   
                           & (0.0054)              & (0.0044)              & (0.0044)              & (0.0027)               & (0.0025)\\   
      hh\_income           & -0.0022               & 0.0012                & 0.0036                & -0.0008                & -0.0008\\   
                           & (0.0039)              & (0.0032)              & (0.0032)              & (0.0020)               & (0.0018)\\   
      int\_trauma\_exp     & -0.0021               & -0.0010               & -0.0012               & 0.0003                 & $-7.47\times 10^{-5}$\\    
                           & (0.0028)              & (0.0023)              & (0.0023)              & (0.0014)               & (0.0013)\\   
      \midrule
      \emph{Fit statistics}\\
      Observations         & 667                   & 667                   & 667                   & 667                    & 667\\  
      R$^2$                & 0.09448               & 0.02157               & 0.01093               & 0.01966                & 0.00522\\  
      Adjusted R$^2$       & 0.07786               & 0.00362               & -0.00721              & 0.00167                & -0.01303\\  
      F-test               & 5.6861                & 1.2017                & 0.60246               & 1.0930                 & 0.28618\\  
      F-test, p-value      & $2.3\times 10^{-9}$   & 0.27754               & 0.84103               & 0.36292                & 0.99143\\  
      \midrule \midrule
      \multicolumn{6}{l}{\emph{IID standard-errors in parentheses}}\\
      \multicolumn{6}{l}{\emph{Signif. Codes: ***: 0.01, **: 0.05, *: 0.1}}\\
   \end{tabular}
    }
\end{table}

\begin{table}[H]
   \caption{Bias tests for Ambition: No Ambition}
   \centering
   \resizebox{\linewidth}{!}{%
   \begin{tabular}{lccccc}
      \tabularnewline \midrule \midrule
      Dependent Variables: & ChatGPT               & Llama213Bchat         & Llama213B              & iQualChatGPTaug      & iQual\\  
      Model:               & (1)                   & (2)                   & (3)                    & (4)                  & (5)\\  
      \midrule
      \emph{Variables}\\
      (Intercept)          & -0.1563$^{***}$       & -0.4332$^{***}$       & -0.5035$^{***}$        & -0.0085              & -0.0087\\   
                           & (0.0219)              & (0.0336)              & (0.0352)               & (0.0075)             & (0.0081)\\   
      data\_roundR3        & 0.0406$^{*}$          & 0.0864$^{***}$        & -0.0430                & -0.0044              & -0.0061\\   
                           & (0.0210)              & (0.0323)              & (0.0338)               & (0.0072)             & (0.0078)\\   
      refugee              & 0.0385$^{***}$        & 0.0284                & 0.0613$^{***}$         & 0.0136$^{***}$       & 0.0150$^{***}$\\   
                           & (0.0137)              & (0.0210)              & (0.0221)               & (0.0047)             & (0.0051)\\   
      num\_child           & -0.0032               & 0.0014                & -0.0087                & -0.0002              & -0.0008\\   
                           & (0.0035)              & (0.0053)              & (0.0056)               & (0.0012)             & (0.0013)\\   
      hh\_head\_sex        & -0.0436$^{***}$       & -0.0442$^{**}$        & -0.0607$^{***}$        & -0.0054              & -0.0107$^{**}$\\   
                           & (0.0124)              & (0.0191)              & (0.0200)               & (0.0043)             & (0.0046)\\   
      hh\_head\_age        & $3.79\times 10^{-5}$  & -0.0012               & -0.0011                & 0.0002               & 0.0002\\   
                           & (0.0005)              & (0.0008)              & (0.0009)               & (0.0002)             & (0.0002)\\   
      parent\_eduyears     & 0.0061$^{***}$        & 0.0052$^{**}$         & 0.0085$^{***}$         & -0.0001              & -0.0004\\   
                           & (0.0014)              & (0.0021)              & (0.0022)               & (0.0005)             & (0.0005)\\   
      parent\_reledu       & 0.0138                & 0.0115                & 0.0124                 & -0.0104              & -0.0161$^{*}$\\   
                           & (0.0235)              & (0.0362)              & (0.0379)               & (0.0081)             & (0.0087)\\   
      eld\_sex             & -0.0030               & 0.0005                & -0.0112                & -0.0021              & $9.85\times 10^{-5}$\\    
                           & (0.0096)              & (0.0147)              & (0.0154)               & (0.0033)             & (0.0036)\\   
      eld\_age             & -0.0004               & -0.0011               & $-3.39\times 10^{-6}$  & $7.4\times 10^{-5}$  & $7.2\times 10^{-5}$\\    
                           & (0.0008)              & (0.0012)              & (0.0012)               & (0.0003)             & (0.0003)\\   
      hh\_asset\_index     & 0.0042                & 0.0111$^{*}$          & 0.0148$^{**}$          & -0.0010              & -0.0012\\   
                           & (0.0039)              & (0.0060)              & (0.0063)               & (0.0014)             & (0.0015)\\   
      hh\_income           & 0.0011                & 0.0025                & -0.0009                & 0.0004               & 0.0006\\   
                           & (0.0028)              & (0.0044)              & (0.0046)               & (0.0010)             & (0.0011)\\   
      int\_trauma\_exp     & 0.0004                & -0.0002               & -0.0028                & -0.0002              & -0.0009\\   
                           & (0.0021)              & (0.0032)              & (0.0033)               & (0.0007)             & (0.0008)\\   
      \midrule
      \emph{Fit statistics}\\
      Observations         & 667                   & 667                   & 667                    & 667                  & 667\\  
      R$^2$                & 0.08578               & 0.06986               & 0.08313                & 0.04273              & 0.05054\\  
      Adjusted R$^2$       & 0.06900               & 0.05280               & 0.06630                & 0.02516              & 0.03312\\  
      F-test               & 5.1135                & 4.0935                & 4.9410                 & 2.4325               & 2.9013\\  
      F-test, p-value      & $3.3\times 10^{-8}$   & $3.49\times 10^{-6}$  & $7.31\times 10^{-8}$   & 0.00426              & 0.00063\\  
      \midrule \midrule
      \multicolumn{6}{l}{\emph{IID standard-errors in parentheses}}\\
      \multicolumn{6}{l}{\emph{Signif. Codes: ***: 0.01, **: 0.05, *: 0.1}}\\
   \end{tabular}
    }
\end{table}

\begin{table}[H]
   \caption{Bias tests for Ambition: Job Secular}
   \centering
   \resizebox{\linewidth}{!}{%
   \begin{tabular}{lccccc}
      \tabularnewline \midrule \midrule
      Dependent Variables: & ChatGPT        & Llama213Bchat   & Llama213B       & iQualChatGPTaug        & iQual\\  
      Model:               & (1)            & (2)             & (3)             & (4)                    & (5)\\  
      \midrule
      \emph{Variables}\\
      (Intercept)          & -0.0379$^{**}$ & -0.1023$^{***}$ & -0.1063$^{***}$ & -0.0030                & -0.0041\\   
                           & (0.0178)       & (0.0229)        & (0.0254)        & (0.0165)               & (0.0159)\\   
      data\_roundR3        & 0.0177         & 0.0197          & 0.0031          & -0.0024                & -0.0190\\   
                           & (0.0171)       & (0.0220)        & (0.0244)        & (0.0159)               & (0.0152)\\   
      refugee              & 0.0260$^{**}$  & 0.0205          & 0.0135          & 0.0078                 & 0.0071\\   
                           & (0.0112)       & (0.0143)        & (0.0159)        & (0.0103)               & (0.0099)\\   
      num\_child           & 0.0061$^{**}$  & 0.0056          & 0.0053          & 0.0036                 & 0.0031\\   
                           & (0.0028)       & (0.0036)        & (0.0040)        & (0.0026)               & (0.0025)\\   
      hh\_head\_sex        & 0.0002         & 0.0121          & 0.0085          & -0.0055                & -0.0021\\   
                           & (0.0101)       & (0.0130)        & (0.0144)        & (0.0094)               & (0.0090)\\   
      hh\_head\_age        & 0.0008$^{*}$   & 0.0009          & 0.0003          & $-5.25\times 10^{-5}$  & -0.0003\\   
                           & (0.0004)       & (0.0006)        & (0.0006)        & (0.0004)               & (0.0004)\\   
      parent\_eduyears     & -0.0010        & -0.0009         & -0.0007         & $-5.75\times 10^{-5}$  & 0.0001\\   
                           & (0.0011)       & (0.0014)        & (0.0016)        & (0.0010)               & (0.0010)\\   
      parent\_reledu       & -0.0202        & -0.0249         & -0.0732$^{***}$ & -0.0137                & -0.0120\\   
                           & (0.0192)       & (0.0246)        & (0.0273)        & (0.0178)               & (0.0171)\\   
      eld\_sex             & 0.0022         & -0.0004         & -0.0010         & -0.0095                & -0.0106\\   
                           & (0.0078)       & (0.0100)        & (0.0111)        & (0.0072)               & (0.0070)\\   
      eld\_age             & -0.0009        & -0.0004         & -0.0002         & $-3.47\times 10^{-5}$  & 0.0006\\   
                           & (0.0006)       & (0.0008)        & (0.0009)        & (0.0006)               & (0.0006)\\   
      hh\_asset\_index     & -0.0010        & 0.0002          & 0.0012          & 0.0009                 & 0.0003\\   
                           & (0.0032)       & (0.0041)        & (0.0046)        & (0.0030)               & (0.0029)\\   
      hh\_income           & 0.0004         & 0.0034          & 0.0029          & -0.0011                & -0.0018\\   
                           & (0.0023)       & (0.0030)        & (0.0033)        & (0.0022)               & (0.0021)\\   
      int\_trauma\_exp     & -0.0024        & -0.0008         & -0.0050$^{**}$  & -0.0005                & -0.0007\\   
                           & (0.0017)       & (0.0022)        & (0.0024)        & (0.0016)               & (0.0015)\\   
      \midrule
      \emph{Fit statistics}\\
      Observations         & 667            & 667             & 667             & 667                    & 667\\  
      R$^2$                & 0.03887        & 0.01983         & 0.02149         & 0.01108                & 0.01631\\  
      Adjusted R$^2$       & 0.02123        & 0.00185         & 0.00353         & -0.00706               & -0.00174\\  
      F-test               & 2.2039         & 1.1029          & 1.1967          & 0.61066                & 0.90381\\  
      F-test, p-value      & 0.01031        & 0.35453         & 0.28112         & 0.83416                & 0.54279\\  
      \midrule \midrule
      \multicolumn{6}{l}{\emph{IID standard-errors in parentheses}}\\
      \multicolumn{6}{l}{\emph{Signif. Codes: ***: 0.01, **: 0.05, *: 0.1}}\\
   \end{tabular}
    }
\end{table}

\begin{table}[H]
   \caption{Bias tests for Ambition: Vocational Training}
   \centering
   \resizebox{\linewidth}{!}{%
   \begin{tabular}{lccccc}
      \tabularnewline \midrule \midrule
      Dependent Variables: & ChatGPT                & Llama213Bchat         & Llama213B      & iQualChatGPTaug & iQual\\  
      Model:               & (1)                    & (2)                   & (3)            & (4)             & (5)\\  
      \midrule
      \emph{Variables}\\
      (Intercept)          & 0.0016                 & -0.1262$^{***}$       & -0.0516$^{**}$ & 0.0120$^{**}$   & 0.0104$^{*}$\\   
                           & (0.0051)               & (0.0233)              & (0.0207)       & (0.0058)        & (0.0055)\\   
      data\_roundR3        & -0.0040                & -0.0118               & -0.0328$^{*}$  & -0.0085         & -0.0063\\   
                           & (0.0049)               & (0.0223)              & (0.0199)       & (0.0056)        & (0.0053)\\   
      refugee              & -0.0021                & -0.0014               & 0.0028         & 0.0012          & -0.0019\\   
                           & (0.0032)               & (0.0146)              & (0.0130)       & (0.0036)        & (0.0035)\\   
      num\_child           & -0.0010                & -0.0023               & -0.0073$^{**}$ & -0.0001         & 0.0004\\   
                           & (0.0008)               & (0.0037)              & (0.0033)       & (0.0009)        & (0.0009)\\   
      hh\_head\_sex        & -0.0028                & 0.0245$^{*}$          & 0.0033         & -0.0014         & 0.0015\\   
                           & (0.0029)               & (0.0132)              & (0.0118)       & (0.0033)        & (0.0032)\\   
      hh\_head\_age        & $9.34\times 10^{-5}$   & $6.68\times 10^{-5}$  & -0.0004        & -0.0002         & -0.0001\\   
                           & (0.0001)               & (0.0006)              & (0.0005)       & (0.0001)        & (0.0001)\\   
      parent\_eduyears     & -0.0004                & -0.0038$^{***}$       & -0.0016        & -0.0007$^{*}$   & -0.0007$^{**}$\\   
                           & (0.0003)               & (0.0015)              & (0.0013)       & (0.0004)        & (0.0003)\\   
      parent\_reledu       & 0.0026                 & -0.0235               & -0.0186        & 0.0071          & 0.0043\\   
                           & (0.0054)               & (0.0250)              & (0.0223)       & (0.0062)        & (0.0060)\\   
      eld\_sex             & 0.0026                 & 0.0113                & 0.0038         & 0.0034          & 0.0009\\   
                           & (0.0022)               & (0.0102)              & (0.0091)       & (0.0025)        & (0.0024)\\   
      eld\_age             & $-1.58\times 10^{-5}$  & 0.0009                & 0.0008         & 0.0001          & $7.07\times 10^{-5}$\\    
                           & (0.0002)               & (0.0008)              & (0.0007)       & (0.0002)        & (0.0002)\\   
      hh\_asset\_index     & $3.77\times 10^{-5}$   & -0.0010               & -0.0036        & 0.0009          & 0.0002\\   
                           & (0.0009)               & (0.0042)              & (0.0037)       & (0.0010)        & (0.0010)\\   
      hh\_income           & 0.0002                 & 0.0016                & 0.0016         & 0.0005          & 0.0005\\   
                           & (0.0007)               & (0.0030)              & (0.0027)       & (0.0008)        & (0.0007)\\   
      int\_trauma\_exp     & -0.0001                & -0.0010               & -0.0007        & -0.0006         & -0.0002\\   
                           & (0.0005)               & (0.0022)              & (0.0019)       & (0.0005)        & (0.0005)\\   
      \midrule
      \emph{Fit statistics}\\
      Observations         & 667                    & 667                   & 667            & 667             & 667\\  
      R$^2$                & 0.01689                & 0.02743               & 0.01987        & 0.02429         & 0.01802\\  
      Adjusted R$^2$       & -0.00115               & 0.00958               & 0.00189        & 0.00638         & $6.82\times 10^{-6}$\\   
      F-test               & 0.93641                & 1.5368                & 1.1049         & 1.3565          & 1.0004\\  
      F-test, p-value      & 0.50962                & 0.10618               & 0.35279        & 0.18220         & 0.44678\\  
      \midrule \midrule
      \multicolumn{6}{l}{\emph{IID standard-errors in parentheses}}\\
      \multicolumn{6}{l}{\emph{Signif. Codes: ***: 0.01, **: 0.05, *: 0.1}}\\
   \end{tabular}
    }
\end{table}

\begin{table}[H]
   \caption{Bias tests for Ambition: Entrepreneur}
   \centering
   \resizebox{\linewidth}{!}{%
   \begin{tabular}{lccccc}
      \tabularnewline \midrule \midrule
      Dependent Variables: & ChatGPT         & Llama213Bchat & Llama213B              & iQualChatGPTaug        & iQual\\  
      Model:               & (1)             & (2)           & (3)                    & (4)                    & (5)\\  
      \midrule
      \emph{Variables}\\
      (Intercept)          & -0.0273$^{**}$  & 0.0001        & -0.0034                & -0.0136$^{*}$          & -0.0092\\   
                           & (0.0119)        & (0.0112)      & (0.0096)               & (0.0070)               & (0.0073)\\   
      data\_roundR3        & -0.0035         & -0.0025       & 0.0032                 & 0.0014                 & 0.0058\\   
                           & (0.0114)        & (0.0108)      & (0.0092)               & (0.0068)               & (0.0070)\\   
      refugee              & -0.0112         & -0.0105       & -0.0016                & -0.0057                & -0.0015\\   
                           & (0.0075)        & (0.0070)      & (0.0060)               & (0.0044)               & (0.0046)\\   
      num\_child           & 0.0002          & 0.0017        & 0.0008                 & 0.0024$^{**}$          & 0.0019$^{*}$\\   
                           & (0.0019)        & (0.0018)      & (0.0015)               & (0.0011)               & (0.0012)\\   
      hh\_head\_sex        & -0.0186$^{***}$ & 0.0074        & 0.0138$^{**}$          & 0.0004                 & 0.0036\\   
                           & (0.0068)        & (0.0064)      & (0.0055)               & (0.0040)               & (0.0041)\\   
      hh\_head\_age        & -0.0003         & 0.0005$^{*}$  & 0.0004                 & 0.0001                 & $4.05\times 10^{-5}$\\    
                           & (0.0003)        & (0.0003)      & (0.0002)               & (0.0002)               & (0.0002)\\   
      parent\_eduyears     & 0.0007          & -0.0006       & -0.0003                & 0.0007                 & 0.0003\\   
                           & (0.0007)        & (0.0007)      & (0.0006)               & (0.0004)               & (0.0005)\\   
      parent\_reledu       & 0.0047          & -0.0092       & -0.0183$^{*}$          & -0.0005                & 0.0003\\   
                           & (0.0128)        & (0.0121)      & (0.0104)               & (0.0076)               & (0.0078)\\   
      eld\_sex             & 0.0111$^{**}$   & -0.0086$^{*}$ & -0.0111$^{***}$        & 0.0007                 & -0.0018\\   
                           & (0.0052)        & (0.0049)      & (0.0042)               & (0.0031)               & (0.0032)\\   
      eld\_age             & 0.0003          & -0.0003       & -0.0003                & $-2.76\times 10^{-5}$  & -0.0001\\   
                           & (0.0004)        & (0.0004)      & (0.0003)               & (0.0002)               & (0.0003)\\   
      hh\_asset\_index     & -0.0025         & 0.0004        & 0.0016                 & -0.0024$^{*}$          & -0.0015\\   
                           & (0.0021)        & (0.0020)      & (0.0017)               & (0.0013)               & (0.0013)\\   
      hh\_income           & 0.0003          & -0.0007       & $-6.87\times 10^{-5}$  & 0.0006                 & 0.0013\\   
                           & (0.0016)        & (0.0015)      & (0.0013)               & (0.0009)               & (0.0009)\\   
      int\_trauma\_exp     & 0.0024$^{**}$   & 0.0001        & -0.0004                & $6.71\times 10^{-5}$   & -0.0004\\   
                           & (0.0011)        & (0.0011)      & (0.0009)               & (0.0007)               & (0.0007)\\   
      \midrule
      \emph{Fit statistics}\\
      Observations         & 667             & 667           & 667                    & 667                    & 667\\  
      R$^2$                & 0.02655         & 0.02647       & 0.02927                & 0.01755                & 0.01095\\  
      Adjusted R$^2$       & 0.00869         & 0.00861       & 0.01146                & -0.00047               & -0.00719\\  
      F-test               & 1.4864          & 1.4819        & 1.6436                 & 0.97374                & 0.60359\\  
      F-test, p-value      & 0.12408         & 0.12581       & 0.07548                & 0.47254                & 0.84009\\  
      \midrule \midrule
      \multicolumn{6}{l}{\emph{IID standard-errors in parentheses}}\\
      \multicolumn{6}{l}{\emph{Signif. Codes: ***: 0.01, **: 0.05, *: 0.1}}\\
   \end{tabular}
    }
\end{table}

\begin{table}[H]
   \caption{Bias tests for Ambition: Education High}
   \centering
   \resizebox{\linewidth}{!}{%
   \begin{tabular}{lccccc}
      \tabularnewline \midrule \midrule
      Dependent Variables: & ChatGPT                & Llama213Bchat   & Llama213B       & iQualChatGPTaug & iQual\\  
      Model:               & (1)                    & (2)             & (3)             & (4)             & (5)\\  
      \midrule
      \emph{Variables}\\
      (Intercept)          & -0.1557$^{***}$        & -0.2259$^{***}$ & -0.3257$^{***}$ & 0.0112          & 0.0268\\   
                           & (0.0233)               & (0.0294)        & (0.0298)        & (0.0172)        & (0.0171)\\   
      data\_roundR3        & 0.0246                 & 0.0196          & -0.0146         & -0.0020         & -0.0192\\   
                           & (0.0224)               & (0.0283)        & (0.0286)        & (0.0165)        & (0.0164)\\   
      refugee              & 0.0162                 & -0.0075         & 0.0063          & 0.0095          & 0.0119\\   
                           & (0.0146)               & (0.0184)        & (0.0186)        & (0.0108)        & (0.0107)\\   
      num\_child           & 0.0056                 & -0.0043         & 0.0096$^{**}$   & 0.0009          & -0.0005\\   
                           & (0.0037)               & (0.0047)        & (0.0047)        & (0.0027)        & (0.0027)\\   
      hh\_head\_sex        & 0.0154                 & 0.0404$^{**}$   & 0.0077          & 0.0128          & 0.0211$^{**}$\\   
                           & (0.0133)               & (0.0167)        & (0.0169)        & (0.0098)        & (0.0097)\\   
      hh\_head\_age        & 0.0007                 & 0.0005          & 0.0010          & -0.0004         & -0.0007$^{*}$\\   
                           & (0.0006)               & (0.0007)        & (0.0007)        & (0.0004)        & (0.0004)\\   
      parent\_eduyears     & -0.0036$^{**}$         & -0.0044$^{**}$  & -0.0007         & -0.0008         & -0.0010\\   
                           & (0.0015)               & (0.0018)        & (0.0019)        & (0.0011)        & (0.0011)\\   
      parent\_reledu       & -0.0241                & 0.0226          & 0.0018          & -0.0001         & 0.0163\\   
                           & (0.0251)               & (0.0317)        & (0.0320)        & (0.0185)        & (0.0184)\\   
      eld\_sex             & 0.0074                 & 0.0143          & 0.0337$^{***}$  & 0.0023          & 0.0025\\   
                           & (0.0102)               & (0.0129)        & (0.0131)        & (0.0075)        & (0.0075)\\   
      eld\_age             & $-5.05\times 10^{-5}$  & 0.0002          & 0.0007          & -0.0003         & 0.0003\\   
                           & (0.0008)               & (0.0010)        & (0.0011)        & (0.0006)        & (0.0006)\\   
      hh\_asset\_index     & 0.0027                 & -0.0039         & -0.0009         & 0.0067$^{**}$   & 0.0054$^{*}$\\   
                           & (0.0042)               & (0.0053)        & (0.0054)        & (0.0031)        & (0.0031)\\   
      hh\_income           & 0.0017                 & 0.0065$^{*}$    & 0.0061          & 0.0006          & 0.0006\\   
                           & (0.0030)               & (0.0038)        & (0.0039)        & (0.0022)        & (0.0022)\\   
      int\_trauma\_exp     & 0.0013                 & 0.0007          & 0.0022          & 0.0009          & -0.0020\\   
                           & (0.0022)               & (0.0028)        & (0.0028)        & (0.0016)        & (0.0016)\\   
      \midrule
      \emph{Fit statistics}\\
      Observations         & 667                    & 667             & 667             & 667             & 667\\  
      R$^2$                & 0.03744                & 0.03659         & 0.03663         & 0.01590         & 0.02704\\  
      Adjusted R$^2$       & 0.01978                & 0.01891         & 0.01896         & -0.00215        & 0.00919\\  
      F-test               & 2.1198                 & 2.0700          & 2.0724          & 0.88080         & 1.5148\\  
      F-test, p-value      & 0.01415                & 0.01702         & 0.01687         & 0.56652         & 0.11370\\  
      \midrule \midrule
      \multicolumn{6}{l}{\emph{IID standard-errors in parentheses}}\\
      \multicolumn{6}{l}{\emph{Signif. Codes: ***: 0.01, **: 0.05, *: 0.1}}\\
   \end{tabular}
    }
\end{table}

\begin{table}[H]
   \caption{Bias tests for Ambition: Education Neutral}
   \centering
   \resizebox{\linewidth}{!}{%
   \begin{tabular}{lccccc}
      \tabularnewline \midrule \midrule
      Dependent Variables: & ChatGPT         & Llama213Bchat   & Llama213B       & iQualChatGPTaug       & iQual\\  
      Model:               & (1)             & (2)             & (3)             & (4)                   & (5)\\  
      \midrule
      \emph{Variables}\\
      (Intercept)          & -0.3200$^{***}$ & -0.3254$^{***}$ & -0.1237$^{***}$ & -0.0377$^{*}$         & -0.0144\\   
                           & (0.0331)        & (0.0350)        & (0.0354)        & (0.0223)              & (0.0220)\\   
      data\_roundR3        & 0.0043          & 0.0721$^{**}$   & -0.0076         & 0.0498$^{**}$         & 0.0312\\   
                           & (0.0318)        & (0.0336)        & (0.0340)        & (0.0214)              & (0.0211)\\   
      refugee              & 0.0227          & 0.0118          & 0.0608$^{***}$  & 0.0268$^{*}$          & 0.0181\\   
                           & (0.0207)        & (0.0219)        & (0.0222)        & (0.0140)              & (0.0138)\\   
      num\_child           & 0.0039          & 0.0021          & -0.0010         & -0.0070$^{**}$        & -0.0060$^{*}$\\   
                           & (0.0053)        & (0.0056)        & (0.0056)        & (0.0035)              & (0.0035)\\   
      hh\_head\_sex        & 0.0383$^{**}$   & 0.0027          & 0.0084          & 0.0155                & 0.0166\\   
                           & (0.0188)        & (0.0199)        & (0.0201)        & (0.0127)              & (0.0125)\\   
      hh\_head\_age        & 0.0007          & 0.0017$^{**}$   & -0.0004         & $-7.1\times 10^{-6}$  & $-8.23\times 10^{-5}$\\    
                           & (0.0008)        & (0.0009)        & (0.0009)        & (0.0005)              & (0.0005)\\   
      parent\_eduyears     & -0.0020         & -0.0077$^{***}$ & -0.0022         & -0.0014               & -0.0012\\   
                           & (0.0021)        & (0.0022)        & (0.0022)        & (0.0014)              & (0.0014)\\   
      parent\_reledu       & 0.0487          & -0.0836$^{**}$  & 0.0042          & 0.0187                & 0.0229\\   
                           & (0.0356)        & (0.0377)        & (0.0381)        & (0.0240)              & (0.0236)\\   
      eld\_sex             & 0.0318$^{**}$   & -0.0033         & -0.0124         & 0.0022                & -0.0044\\   
                           & (0.0145)        & (0.0154)        & (0.0155)        & (0.0098)              & (0.0096)\\   
      eld\_age             & 0.0005          & -0.0023$^{*}$   & -0.0015         & -0.0006               & -0.0003\\   
                           & (0.0012)        & (0.0012)        & (0.0013)        & (0.0008)              & (0.0008)\\   
      hh\_asset\_index     & -0.0079         & -0.0116$^{*}$   & 0.0076          & 0.0074$^{*}$          & 0.0048\\   
                           & (0.0060)        & (0.0063)        & (0.0064)        & (0.0040)              & (0.0040)\\   
      hh\_income           & 0.0031          & 0.0029          & 0.0036          & -0.0012               & 0.0014\\   
                           & (0.0043)        & (0.0046)        & (0.0046)        & (0.0029)              & (0.0029)\\   
      int\_trauma\_exp     & -0.0048         & -0.0039         & -0.0062$^{*}$   & 0.0020                & 0.0011\\   
                           & (0.0031)        & (0.0033)        & (0.0033)        & (0.0021)              & (0.0021)\\   
      \midrule
      \emph{Fit statistics}\\
      Observations         & 667             & 667             & 667             & 667                   & 667\\  
      R$^2$                & 0.04520         & 0.05661         & 0.03922         & 0.04675               & 0.02593\\  
      Adjusted R$^2$       & 0.02768         & 0.03930         & 0.02159         & 0.02926               & 0.00806\\  
      F-test               & 2.5800          & 3.2705          & 2.2248          & 2.6727                & 1.4507\\  
      F-test, p-value      & 0.00236         & 0.00013         & 0.00952         & 0.00162               & 0.13825\\  
      \midrule \midrule
      \multicolumn{6}{l}{\emph{IID standard-errors in parentheses}}\\
      \multicolumn{6}{l}{\emph{Signif. Codes: ***: 0.01, **: 0.05, *: 0.1}}\\
   \end{tabular}
    }
\end{table}

\begin{table}[H]
   \caption{Bias tests for Ambition: Education Low}
   \centering
   \resizebox{\linewidth}{!}{%
   \begin{tabular}{lccccc}
      \tabularnewline \midrule \midrule
      Dependent Variables: & ChatGPT        & Llama213Bchat          & Llama213B       & iQualChatGPTaug        & iQual\\  
      Model:               & (1)            & (2)                    & (3)             & (4)                    & (5)\\  
      \midrule
      \emph{Variables}\\
      (Intercept)          & -0.0171        & -0.1612$^{***}$        & -0.0657$^{***}$ & 0.0112                 & 0.0150$^{*}$\\   
                           & (0.0118)       & (0.0224)               & (0.0188)        & (0.0089)               & (0.0090)\\   
      data\_roundR3        & -0.0123        & 0.0068                 & 0.0050          & -0.0032                & -0.0062\\   
                           & (0.0113)       & (0.0215)               & (0.0181)        & (0.0086)               & (0.0086)\\   
      refugee              & 0.0157$^{**}$  & 0.0230                 & 0.0022          & -0.0042                & -0.0117$^{**}$\\   
                           & (0.0074)       & (0.0140)               & (0.0118)        & (0.0056)               & (0.0056)\\   
      num\_child           & -0.0016        & -0.0014                & 0.0036          & 0.0010                 & 0.0004\\   
                           & (0.0019)       & (0.0035)               & (0.0030)        & (0.0014)               & (0.0014)\\   
      hh\_head\_sex        & -0.0075        & -0.0048                & -0.0066         & -0.0038                & -0.0077\\   
                           & (0.0067)       & (0.0127)               & (0.0107)        & (0.0051)               & (0.0051)\\   
      hh\_head\_age        & -0.0003        & $3.83\times 10^{-6}$   & -0.0002         & -0.0002                & -0.0002\\   
                           & (0.0003)       & (0.0005)               & (0.0005)        & (0.0002)               & (0.0002)\\   
      parent\_eduyears     & -0.0002        & 0.0004                 & -0.0014         & 0.0003                 & -0.0007\\   
                           & (0.0007)       & (0.0014)               & (0.0012)        & (0.0006)               & (0.0006)\\   
      parent\_reledu       & -0.0051        & -0.0289                & -0.0347$^{*}$   & -0.0024                & -0.0063\\   
                           & (0.0126)       & (0.0240)               & (0.0203)        & (0.0096)               & (0.0097)\\   
      eld\_sex             & 0.0173$^{***}$ & 0.0011                 & 0.0048          & 0.0048                 & 0.0032\\   
                           & (0.0052)       & (0.0098)               & (0.0083)        & (0.0039)               & (0.0039)\\   
      eld\_age             & 0.0008$^{*}$   & 0.0003                 & -0.0007         & $-9.99\times 10^{-5}$  & $2.42\times 10^{-5}$\\    
                           & (0.0004)       & (0.0008)               & (0.0007)        & (0.0003)               & (0.0003)\\   
      hh\_asset\_index     & 0.0006         & -0.0021                & -0.0009         & -0.0011                & -0.0014\\   
                           & (0.0021)       & (0.0040)               & (0.0034)        & (0.0016)               & (0.0016)\\   
      hh\_income           & 0.0002         & -0.0001                & 0.0010          & 0.0002                 & -0.0003\\   
                           & (0.0015)       & (0.0029)               & (0.0025)        & (0.0012)               & (0.0012)\\   
      int\_trauma\_exp     & 0.0003         & $-8.98\times 10^{-5}$  & 0.0006          & 0.0008                 & 0.0014\\   
                           & (0.0011)       & (0.0021)               & (0.0018)        & (0.0008)               & (0.0008)\\   
      \midrule
      \emph{Fit statistics}\\
      Observations         & 667            & 667                    & 667             & 667                    & 667\\  
      R$^2$                & 0.03226        & 0.01698                & 0.01891         & 0.01307                & 0.01801\\  
      Adjusted R$^2$       & 0.01450        & -0.00106               & 0.00091         & -0.00504               & $-3.96\times 10^{-6}$\\   
      F-test               & 1.8168         & 0.94150                & 1.0507          & 0.72175                & 0.99978\\  
      F-test, p-value      & 0.04214        & 0.50451                & 0.40003         & 0.73085                & 0.44735\\  
      \midrule \midrule
      \multicolumn{6}{l}{\emph{IID standard-errors in parentheses}}\\
      \multicolumn{6}{l}{\emph{Signif. Codes: ***: 0.01, **: 0.05, *: 0.1}}\\
   \end{tabular}
    }
\end{table}

\begin{table}[H]
   \caption{Bias tests for Ambition: Education Religious}
   \centering
   \resizebox{\linewidth}{!}{%
   \begin{tabular}{lccccc}
      \tabularnewline \midrule \midrule
      Dependent Variables: & ChatGPT               & Llama213Bchat          & Llama213B     & iQualChatGPTaug        & iQual\\  
      Model:               & (1)                   & (2)                    & (3)           & (4)                    & (5)\\  
      \midrule
      \emph{Variables}\\
      (Intercept)          & -0.0076               & -0.0266$^{*}$          & -0.0251$^{*}$ & 0.0147                 & 0.0256$^{**}$\\   
                           & (0.0135)              & (0.0146)               & (0.0141)      & (0.0107)               & (0.0109)\\   
      data\_roundR3        & -0.0209               & -0.0424$^{***}$        & -0.0260$^{*}$ & -0.0037                & -0.0028\\   
                           & (0.0129)              & (0.0140)               & (0.0136)      & (0.0102)               & (0.0105)\\   
      refugee              & 0.0096                & 0.0154$^{*}$           & 0.0031        & 0.0080                 & 0.0075\\   
                           & (0.0084)              & (0.0091)               & (0.0089)      & (0.0067)               & (0.0069)\\   
      num\_child           & -0.0017               & -0.0007                & -0.0021       & -0.0012                & -0.0030$^{*}$\\   
                           & (0.0021)              & (0.0023)               & (0.0022)      & (0.0017)               & (0.0017)\\   
      hh\_head\_sex        & -0.0055               & 0.0094                 & -0.0105       & -0.0032                & -0.0045\\   
                           & (0.0077)              & (0.0083)               & (0.0080)      & (0.0061)               & (0.0062)\\   
      hh\_head\_age        & $1.68\times 10^{-6}$  & 0.0001                 & 0.0003        & $3.15\times 10^{-6}$   & $-3.36\times 10^{-5}$\\    
                           & (0.0003)              & (0.0004)               & (0.0003)      & (0.0003)               & (0.0003)\\   
      parent\_eduyears     & -0.0012               & -0.0005                & -0.0006       & -0.0005                & -0.0012$^{*}$\\   
                           & (0.0008)              & (0.0009)               & (0.0009)      & (0.0007)               & (0.0007)\\   
      parent\_reledu       & 0.0048                & -0.0313$^{**}$         & -0.0153       & 0.0127                 & 0.0157\\   
                           & (0.0145)              & (0.0157)               & (0.0152)      & (0.0115)               & (0.0118)\\   
      eld\_sex             & 0.0067                & 0.0032                 & 0.0053        & $-4.65\times 10^{-5}$  & -0.0019\\   
                           & (0.0059)              & (0.0064)               & (0.0062)      & (0.0047)               & (0.0048)\\   
      eld\_age             & 0.0002                & 0.0007                 & 0.0004        & -0.0002                & -0.0003\\   
                           & (0.0005)              & (0.0005)               & (0.0005)      & (0.0004)               & (0.0004)\\   
      hh\_asset\_index     & 0.0039                & 0.0070$^{***}$         & 0.0015        & 0.0025                 & 0.0037$^{*}$\\   
                           & (0.0024)              & (0.0026)               & (0.0025)      & (0.0019)               & (0.0020)\\   
      hh\_income           & 0.0002                & -0.0010                & -0.0003       & -0.0015                & -0.0010\\   
                           & (0.0018)              & (0.0019)               & (0.0018)      & (0.0014)               & (0.0014)\\   
      int\_trauma\_exp     & 0.0014                & $-9.22\times 10^{-5}$  & 0.0022$^{*}$  & $9.23\times 10^{-5}$   & 0.0009\\   
                           & (0.0013)              & (0.0014)               & (0.0013)      & (0.0010)               & (0.0010)\\   
      \midrule
      \emph{Fit statistics}\\
      Observations         & 667                   & 667                    & 667           & 667                    & 667\\  
      R$^2$                & 0.02348               & 0.05358                & 0.02626       & 0.01853                & 0.03037\\  
      Adjusted R$^2$       & 0.00556               & 0.03621                & 0.00839       & 0.00052                & 0.01258\\  
      F-test               & 1.3103                & 3.0853                 & 1.4698        & 1.0289                 & 1.7073\\  
      F-test, p-value      & 0.20751               & 0.00029                & 0.13050       & 0.41997                & 0.06115\\  
      \midrule \midrule
      \multicolumn{6}{l}{\emph{IID standard-errors in parentheses}}\\
      \multicolumn{6}{l}{\emph{Signif. Codes: ***: 0.01, **: 0.05, *: 0.1}}\\
   \end{tabular}
    }
\end{table}

\begin{table}[H]
   \caption{Bias tests for Ambition: Marriage}
   \centering
   \resizebox{\linewidth}{!}{%
   \begin{tabular}{lccccc}
      \tabularnewline \midrule \midrule
      Dependent Variables: & ChatGPT               & Llama213Bchat         & Llama213B            & iQualChatGPTaug       & iQual\\  
      Model:               & (1)                   & (2)                   & (3)                  & (4)                   & (5)\\  
      \midrule
      \emph{Variables}\\
      (Intercept)          & 0.0031                & 0.0204                & 0.0462$^{***}$       & 0.0029                & -0.0052\\   
                           & (0.0091)              & (0.0126)              & (0.0153)             & (0.0082)              & (0.0078)\\   
      data\_roundR3        & 0.0036                & -0.0231$^{*}$         & -0.0479$^{***}$      & 0.0106                & 0.0136$^{*}$\\   
                           & (0.0087)              & (0.0121)              & (0.0147)             & (0.0078)              & (0.0075)\\   
      refugee              & -0.0091               & -0.0068               & -0.0032              & -0.0008               & -0.0002\\   
                           & (0.0057)              & (0.0079)              & (0.0096)             & (0.0051)              & (0.0049)\\   
      num\_child           & -0.0002               & 0.0005                & 0.0036               & -0.0029$^{**}$        & -0.0028$^{**}$\\   
                           & (0.0014)              & (0.0020)              & (0.0024)             & (0.0013)              & (0.0012)\\   
      hh\_head\_sex        & 0.0003                & -0.0018               & 0.0013               & 0.0006                & $-4.3\times 10^{-5}$\\    
                           & (0.0052)              & (0.0072)              & (0.0087)             & (0.0046)              & (0.0044)\\   
      hh\_head\_age        & $4.94\times 10^{-5}$  & $-2.9\times 10^{-5}$  & -0.0004              & 0.0003                & 0.0004$^{*}$\\   
                           & (0.0002)              & (0.0003)              & (0.0004)             & (0.0002)              & (0.0002)\\   
      parent\_eduyears     & -0.0007               & -0.0025$^{***}$       & -0.0006              & -0.0004               & -0.0003\\   
                           & (0.0006)              & (0.0008)              & (0.0010)             & (0.0005)              & (0.0005)\\   
      parent\_reledu       & -0.0162$^{*}$         & -0.0172               & -0.0178              & -0.0103               & -0.0093\\   
                           & (0.0098)              & (0.0136)              & (0.0164)             & (0.0088)              & (0.0084)\\   
      eld\_sex             & -0.0010               & 0.0077                & 0.0256$^{***}$       & -0.0072$^{**}$        & -0.0079$^{**}$\\   
                           & (0.0040)              & (0.0055)              & (0.0067)             & (0.0036)              & (0.0034)\\   
      eld\_age             & -0.0005$^{*}$         & 0.0001                & 0.0005               & -0.0004               & -0.0004\\   
                           & (0.0003)              & (0.0004)              & (0.0005)             & (0.0003)              & (0.0003)\\   
      hh\_asset\_index     & -0.0008               & -0.0034               & -0.0061$^{**}$       & -0.0014               & -0.0023$^{*}$\\   
                           & (0.0016)              & (0.0023)              & (0.0028)             & (0.0015)              & (0.0014)\\   
      hh\_income           & 0.0016                & 0.0026                & 0.0025               & 0.0010                & 0.0009\\   
                           & (0.0012)              & (0.0016)              & (0.0020)             & (0.0011)              & (0.0010)\\   
      int\_trauma\_exp     & 0.0003                & -0.0015               & -0.0007              & $-9.1\times 10^{-5}$  & 0.0002\\   
                           & (0.0009)              & (0.0012)              & (0.0014)             & (0.0008)              & (0.0007)\\   
      \midrule
      \emph{Fit statistics}\\
      Observations         & 667                   & 667                   & 667                  & 667                   & 667\\  
      R$^2$                & 0.02664               & 0.05043               & 0.08818              & 0.02193               & 0.03117\\  
      Adjusted R$^2$       & 0.00878               & 0.03301               & 0.07144              & 0.00398               & 0.01339\\  
      F-test               & 1.4918                & 2.8945                & 5.2702               & 1.2217                & 1.7535\\  
      F-test, p-value      & 0.12203               & 0.00065               & $1.6\times 10^{-8}$  & 0.26352               & 0.05235\\  
      \midrule \midrule
      \multicolumn{6}{l}{\emph{IID standard-errors in parentheses}}\\
      \multicolumn{6}{l}{\emph{Signif. Codes: ***: 0.01, **: 0.05, *: 0.1}}\\
   \end{tabular}
    }
\end{table}

\begin{table}[H]
   \caption{Bias tests for Ambition: Migration}
   \centering
   \resizebox{\linewidth}{!}{%
   \begin{tabular}{lccccc}
      \tabularnewline \midrule \midrule
      Dependent Variables: & ChatGPT                & Llama213Bchat & Llama213B              & iQualChatGPTaug        & iQual\\  
      Model:               & (1)                    & (2)           & (3)                    & (4)                    & (5)\\  
      \midrule
      \emph{Variables}\\
      (Intercept)          & 0.0020                 & -0.0100       & 0.0057                 & 0.0024                 & 0.0082\\   
                           & (0.0081)               & (0.0091)      & (0.0080)               & (0.0057)               & (0.0067)\\   
      data\_roundR3        & 0.0044                 & -0.0111       & -0.0099                & -0.0051                & -0.0076\\   
                           & (0.0078)               & (0.0088)      & (0.0077)               & (0.0054)               & (0.0064)\\   
      refugee              & -0.0099$^{*}$          & 0.0009        & 0.0031                 & -0.0001                & -0.0009\\   
                           & (0.0051)               & (0.0057)      & (0.0050)               & (0.0036)               & (0.0042)\\   
      num\_child           & 0.0009                 & -0.0004       & 0.0018                 & $4.24\times 10^{-6}$   & 0.0006\\   
                           & (0.0013)               & (0.0014)      & (0.0013)               & (0.0009)               & (0.0011)\\   
      hh\_head\_sex        & -0.0080$^{*}$          & 0.0050        & $8.42\times 10^{-5}$   & -0.0011                & -0.0021\\   
                           & (0.0046)               & (0.0052)      & (0.0045)               & (0.0032)               & (0.0038)\\   
      hh\_head\_age        & -0.0004$^{**}$         & -0.0003       & $-1.95\times 10^{-5}$  & $3.56\times 10^{-6}$   & -0.0001\\   
                           & (0.0002)               & (0.0002)      & (0.0002)               & (0.0001)               & (0.0002)\\   
      parent\_eduyears     & $-8.93\times 10^{-5}$  & 0.0003        & 0.0003                 & -0.0002                & -0.0007$^{*}$\\   
                           & (0.0005)               & (0.0006)      & (0.0005)               & (0.0004)               & (0.0004)\\   
      parent\_reledu       & 0.0062                 & 0.0088        & 0.0081                 & 0.0054                 & 0.0071\\   
                           & (0.0088)               & (0.0098)      & (0.0086)               & (0.0061)               & (0.0072)\\   
      eld\_sex             & 0.0067$^{*}$           & 0.0014        & -0.0043                & 0.0003                 & 0.0007\\   
                           & (0.0036)               & (0.0040)      & (0.0035)               & (0.0025)               & (0.0029)\\   
      eld\_age             & $5.58\times 10^{-5}$   & 0.0004        & $7.59\times 10^{-5}$   & $6.47\times 10^{-5}$   & 0.0001\\   
                           & (0.0003)               & (0.0003)      & (0.0003)               & (0.0002)               & (0.0002)\\   
      hh\_asset\_index     & -0.0014                & -0.0001       & -0.0004                & $1.38\times 10^{-5}$   & 0.0009\\   
                           & (0.0015)               & (0.0016)      & (0.0014)               & (0.0010)               & (0.0012)\\   
      hh\_income           & 0.0003                 & -0.0002       & -0.0002                & $-2.67\times 10^{-5}$  & -0.0008\\   
                           & (0.0011)               & (0.0012)      & (0.0010)               & (0.0007)               & (0.0009)\\   
      int\_trauma\_exp     & -0.0009                & -0.0004       & $-9.94\times 10^{-5}$  & 0.0005                 & 0.0004\\   
                           & (0.0008)               & (0.0009)      & (0.0008)               & (0.0005)               & (0.0006)\\   
      \midrule
      \emph{Fit statistics}\\
      Observations         & 667                    & 667           & 667                    & 667                    & 667\\  
      R$^2$                & 0.03138                & 0.00724       & 0.02165                & 0.00823                & 0.01693\\  
      Adjusted R$^2$       & 0.01361                & -0.01098      & 0.00370                & -0.00997               & -0.00111\\  
      F-test               & 1.7658                 & 0.39750       & 1.2059                 & 0.45233                & 0.93836\\  
      F-test, p-value      & 0.05019                & 0.96466       & 0.27459                & 0.94137                & 0.50766\\  
      \midrule \midrule
      \multicolumn{6}{l}{\emph{IID standard-errors in parentheses}}\\
      \multicolumn{6}{l}{\emph{Signif. Codes: ***: 0.01, **: 0.05, *: 0.1}}\\
   \end{tabular}
    }
\end{table}

\begin{table}[H]
   \caption{Bias tests for Capacity: Reliance On God}
   \centering
   \resizebox{\linewidth}{!}{%
   \begin{tabular}{lccccc}
      \tabularnewline \midrule \midrule
      Dependent Variables: & ChatGPT                & Llama213Bchat   & Llama213B       & iQualChatGPTaug & iQual\\  
      Model:               & (1)                    & (2)             & (3)             & (4)             & (5)\\  
      \midrule
      \emph{Variables}\\
      (Intercept)          & -0.0735$^{***}$        & -0.1134$^{***}$ & -0.0993$^{***}$ & 0.0008          & 0.0066\\   
                           & (0.0158)               & (0.0217)        & (0.0198)        & (0.0120)        & (0.0117)\\   
      data\_roundR3        & -0.0198                & -0.0501$^{**}$  & -0.0186         & -0.0073         & -0.0080\\   
                           & (0.0151)               & (0.0208)        & (0.0190)        & (0.0115)        & (0.0112)\\   
      refugee              & 0.0215$^{**}$          & 0.0242$^{*}$    & 0.0325$^{***}$  & 0.0103          & 0.0070\\   
                           & (0.0099)               & (0.0136)        & (0.0124)        & (0.0075)        & (0.0073)\\   
      num\_child           & 0.0004                 & 0.0014          & 0.0004          & 0.0012          & -0.0006\\   
                           & (0.0025)               & (0.0034)        & (0.0031)        & (0.0019)        & (0.0019)\\   
      hh\_head\_sex        & -0.0002                & -0.0085         & -0.0230$^{**}$  & 0.0097          & 0.0062\\   
                           & (0.0090)               & (0.0123)        & (0.0112)        & (0.0068)        & (0.0067)\\   
      hh\_head\_age        & $-6.25\times 10^{-5}$  & -0.0003         & -0.0003         & -0.0005$^{*}$   & -0.0004\\   
                           & (0.0004)               & (0.0005)        & (0.0005)        & (0.0003)        & (0.0003)\\   
      parent\_eduyears     & -0.0010                & -0.0015         & -0.0008         & -0.0005         & -0.0009\\   
                           & (0.0010)               & (0.0014)        & (0.0012)        & (0.0007)        & (0.0007)\\   
      parent\_reledu       & -0.0275                & -0.0125         & -0.0085         & -0.0181         & -0.0071\\   
                           & (0.0170)               & (0.0233)        & (0.0213)        & (0.0129)        & (0.0126)\\   
      eld\_sex             & 0.0074                 & 0.0095          & 0.0077          & -0.0012         & -0.0020\\   
                           & (0.0069)               & (0.0095)        & (0.0087)        & (0.0052)        & (0.0051)\\   
      eld\_age             & 0.0007                 & 0.0018$^{**}$   & 0.0007          & 0.0002          & 0.0003\\   
                           & (0.0006)               & (0.0008)        & (0.0007)        & (0.0004)        & (0.0004)\\   
      hh\_asset\_index     & 0.0066$^{**}$          & 0.0082$^{**}$   & 0.0075$^{**}$   & 0.0042$^{**}$   & 0.0037$^{*}$\\   
                           & (0.0028)               & (0.0039)        & (0.0036)        & (0.0022)        & (0.0021)\\   
      hh\_income           & 0.0006                 & -0.0018         & 0.0002          & 0.0014          & 0.0018\\   
                           & (0.0021)               & (0.0028)        & (0.0026)        & (0.0016)        & (0.0015)\\   
      int\_trauma\_exp     & 0.0013                 & -0.0007         & 0.0007          & -0.0002         & 0.0004\\   
                           & (0.0015)               & (0.0020)        & (0.0019)        & (0.0011)        & (0.0011)\\   
      \midrule
      \emph{Fit statistics}\\
      Observations         & 667                    & 667             & 667             & 667             & 667\\  
      R$^2$                & 0.02232                & 0.02407         & 0.02126         & 0.01891         & 0.01410\\  
      Adjusted R$^2$       & 0.00438                & 0.00617         & 0.00331         & 0.00091         & -0.00398\\  
      F-test               & 1.2443                 & 1.3443          & 1.1841          & 1.0507          & 0.77974\\  
      F-test, p-value      & 0.24830                & 0.18863         & 0.29035         & 0.40002         & 0.67179\\  
      \midrule \midrule
      \multicolumn{6}{l}{\emph{IID standard-errors in parentheses}}\\
      \multicolumn{6}{l}{\emph{Signif. Codes: ***: 0.01, **: 0.05, *: 0.1}}\\
   \end{tabular}
    }
\end{table}

\begin{table}[H]
   \caption{Bias tests for Capacity: Ability High}
   \centering
   \resizebox{\linewidth}{!}{%
   \begin{tabular}{lccccc}
      \tabularnewline \midrule \midrule
      Dependent Variables: & ChatGPT                & Llama213Bchat   & Llama213B      & iQualChatGPTaug       & iQual\\  
      Model:               & (1)                    & (2)             & (3)            & (4)                   & (5)\\  
      \midrule
      \emph{Variables}\\
      (Intercept)          & -0.1597$^{***}$        & -0.1068$^{***}$ & -0.0103        & -0.0190               & -0.0156\\   
                           & (0.0253)               & (0.0215)        & (0.0167)       & (0.0135)              & (0.0138)\\   
      data\_roundR3        & 0.0807$^{***}$         & 0.0153          & -0.0145        & 0.0050                & 0.0120\\   
                           & (0.0243)               & (0.0206)        & (0.0160)       & (0.0129)              & (0.0132)\\   
      refugee              & -0.0042                & -0.0002         & 0.0038         & -0.0008               & 0.0021\\   
                           & (0.0159)               & (0.0134)        & (0.0105)       & (0.0085)              & (0.0086)\\   
      num\_child           & 0.0071$^{*}$           & 0.0048          & 0.0002         & 0.0009                & 0.0014\\   
                           & (0.0040)               & (0.0034)        & (0.0027)       & (0.0021)              & (0.0022)\\   
      hh\_head\_sex        & 0.0045                 & 0.0061          & -0.0218$^{**}$ & 0.0001                & -0.0043\\   
                           & (0.0144)               & (0.0122)        & (0.0095)       & (0.0077)              & (0.0078)\\   
      hh\_head\_age        & $-3.34\times 10^{-5}$  & 0.0003          & -0.0006        & 0.0001                & -0.0002\\   
                           & (0.0006)               & (0.0005)        & (0.0004)       & (0.0003)              & (0.0003)\\   
      parent\_eduyears     & -0.0024                & -0.0010         & 0.0017         & 0.0005                & -0.0002\\   
                           & (0.0016)               & (0.0013)        & (0.0010)       & (0.0008)              & (0.0009)\\   
      parent\_reledu       & -0.0200                & -0.0648$^{***}$ & 0.0138         & 0.0102                & 0.0130\\   
                           & (0.0272)               & (0.0231)        & (0.0180)       & (0.0145)              & (0.0148)\\   
      eld\_sex             & 0.0175                 & -0.0014         & 0.0117         & 0.0046                & 0.0061\\   
                           & (0.0111)               & (0.0094)        & (0.0073)       & (0.0059)              & (0.0060)\\   
      eld\_age             & -0.0003                & -0.0002         & 0.0005         & $5.98\times 10^{-5}$  & 0.0001\\   
                           & (0.0009)               & (0.0008)        & (0.0006)       & (0.0005)              & (0.0005)\\   
      hh\_asset\_index     & 0.0013                 & -0.0035         & -0.0008        & 0.0015                & 0.0038\\   
                           & (0.0046)               & (0.0039)        & (0.0030)       & (0.0024)              & (0.0025)\\   
      hh\_income           & 0.0026                 & 0.0016          & 0.0018         & 0.0021                & $2.23\times 10^{-5}$\\    
                           & (0.0033)               & (0.0028)        & (0.0022)       & (0.0018)              & (0.0018)\\   
      int\_trauma\_exp     & 0.0017                 & 0.0016          & 0.0011         & 0.0018                & 0.0014\\   
                           & (0.0024)               & (0.0020)        & (0.0016)       & (0.0013)              & (0.0013)\\   
      \midrule
      \emph{Fit statistics}\\
      Observations         & 667                    & 667             & 667            & 667                   & 667\\  
      R$^2$                & 0.08009                & 0.02463         & 0.01936        & 0.01452               & 0.01975\\  
      Adjusted R$^2$       & 0.06321                & 0.00673         & 0.00137        & -0.00356              & 0.00177\\  
      F-test               & 4.7448                 & 1.3763          & 1.0762         & 0.80300               & 1.0982\\  
      F-test, p-value      & $1.8\times 10^{-7}$    & 0.17214         & 0.37742        & 0.64763               & 0.35848\\  
      \midrule \midrule
      \multicolumn{6}{l}{\emph{IID standard-errors in parentheses}}\\
      \multicolumn{6}{l}{\emph{Signif. Codes: ***: 0.01, **: 0.05, *: 0.1}}\\
   \end{tabular}
    }
\end{table}

\begin{table}[H]
   \caption{Bias tests for Capacity: Ability Low}
   \centering
   \resizebox{\linewidth}{!}{%
   \begin{tabular}{lccccc}
      \tabularnewline \midrule \midrule
      Dependent Variables: & ChatGPT                & Llama213Bchat         & Llama213B       & iQualChatGPTaug        & iQual\\  
      Model:               & (1)                    & (2)                   & (3)             & (4)                    & (5)\\  
      \midrule
      \emph{Variables}\\
      (Intercept)          & -0.1072$^{***}$        & -0.1025$^{***}$       & -0.0482$^{***}$ & -0.0090                & -0.0170\\   
                           & (0.0214)               & (0.0221)              & (0.0180)        & (0.0140)               & (0.0138)\\   
      data\_roundR3        & 0.0694$^{***}$         & 0.0658$^{***}$        & 0.0352$^{**}$   & 0.0001                 & 0.0058\\   
                           & (0.0205)               & (0.0212)              & (0.0173)        & (0.0134)               & (0.0133)\\   
      refugee              & 0.0103                 & -0.0070               & -0.0106         & 0.0018                 & 0.0014\\   
                           & (0.0134)               & (0.0138)              & (0.0113)        & (0.0088)               & (0.0087)\\   
      num\_child           & -0.0024                & 0.0012                & 0.0025          & 0.0043$^{*}$           & 0.0036\\   
                           & (0.0034)               & (0.0035)              & (0.0029)        & (0.0022)               & (0.0022)\\   
      hh\_head\_sex        & -0.0369$^{***}$        & -0.0229$^{*}$         & -0.0154         & -0.0049                & -0.0024\\   
                           & (0.0122)               & (0.0126)              & (0.0102)        & (0.0080)               & (0.0079)\\   
      hh\_head\_age        & $6.31\times 10^{-5}$   & 0.0002                & 0.0003          & $-3.59\times 10^{-5}$  & 0.0002\\   
                           & (0.0005)               & (0.0005)              & (0.0004)        & (0.0003)               & (0.0003)\\   
      parent\_eduyears     & 0.0027$^{**}$          & 0.0025$^{*}$          & 0.0015          & -0.0009                & -0.0004\\   
                           & (0.0013)               & (0.0014)              & (0.0011)        & (0.0009)               & (0.0009)\\   
      parent\_reledu       & -0.0160                & -0.0109               & -0.0292         & -0.0159                & -0.0373$^{**}$\\   
                           & (0.0230)               & (0.0238)              & (0.0194)        & (0.0151)               & (0.0149)\\   
      eld\_sex             & 0.0096                 & 0.0097                & -0.0016         & 0.0029                 & -0.0011\\   
                           & (0.0094)               & (0.0097)              & (0.0079)        & (0.0061)               & (0.0061)\\   
      eld\_age             & $7.46\times 10^{-5}$   & -0.0005               & -0.0004         & $9.65\times 10^{-5}$   & $-8.11\times 10^{-5}$\\    
                           & (0.0008)               & (0.0008)              & (0.0006)        & (0.0005)               & (0.0005)\\   
      hh\_asset\_index     & -0.0027                & -0.0075$^{*}$         & -0.0016         & -0.0017                & -0.0005\\   
                           & (0.0039)               & (0.0040)              & (0.0032)        & (0.0025)               & (0.0025)\\   
      hh\_income           & 0.0037                 & 0.0016                & 0.0005          & 0.0007                 & 0.0002\\   
                           & (0.0028)               & (0.0029)              & (0.0023)        & (0.0018)               & (0.0018)\\   
      int\_trauma\_exp     & -0.0004                & -0.0033               & -0.0009         & -0.0019                & -0.0012\\   
                           & (0.0020)               & (0.0021)              & (0.0017)        & (0.0013)               & (0.0013)\\   
      \midrule
      \emph{Fit statistics}\\
      Observations         & 667                    & 667                   & 667             & 667                    & 667\\  
      R$^2$                & 0.11546                & 0.07145               & 0.03663         & 0.01581                & 0.01706\\  
      Adjusted R$^2$       & 0.09923                & 0.05441               & 0.01895         & -0.00225               & -0.00097\\  
      F-test               & 7.1142                 & 4.1934                & 2.0723          & 0.87541                & 0.94597\\  
      F-test, p-value      & $2.85\times 10^{-12}$  & $2.22\times 10^{-6}$  & 0.01688         & 0.57210                & 0.50002\\  
      \midrule \midrule
      \multicolumn{6}{l}{\emph{IID standard-errors in parentheses}}\\
      \multicolumn{6}{l}{\emph{Signif. Codes: ***: 0.01, **: 0.05, *: 0.1}}\\
   \end{tabular}
    }
\end{table}

\begin{table}[H]
   \caption{Bias tests for Capacity: Budget High}
   \centering
   \resizebox{\linewidth}{!}{%
   \begin{tabular}{lccccc}
      \tabularnewline \midrule \midrule
      Dependent Variables: & ChatGPT       & Llama213Bchat          & Llama213B       & iQualChatGPTaug      & iQual\\  
      Model:               & (1)           & (2)                    & (3)             & (4)                  & (5)\\  
      \midrule
      \emph{Variables}\\
      (Intercept)          & -0.0092       & -0.0277                & -0.0472$^{***}$ & 0.0070               & -0.0051\\   
                           & (0.0109)      & (0.0169)               & (0.0164)        & (0.0105)             & (0.0109)\\   
      data\_roundR3        & 0.0140        & -0.0153                & 0.0346$^{**}$   & 0.0046               & 0.0027\\   
                           & (0.0105)      & (0.0162)               & (0.0158)        & (0.0101)             & (0.0105)\\   
      refugee              & -0.0067       & 0.0035                 & -0.0076         & 0.0021               & -0.0036\\   
                           & (0.0069)      & (0.0106)               & (0.0103)        & (0.0066)             & (0.0068)\\   
      num\_child           & 0.0008        & -0.0013                & -0.0043$^{*}$   & -0.0003              & 0.0003\\   
                           & (0.0017)      & (0.0027)               & (0.0026)        & (0.0017)             & (0.0017)\\   
      hh\_head\_sex        & -0.0030       & -0.0003                & -0.0166$^{*}$   & 0.0110$^{*}$         & -0.0026\\   
                           & (0.0062)      & (0.0096)               & (0.0093)        & (0.0060)             & (0.0062)\\   
      hh\_head\_age        & 0.0001        & -0.0001                & 0.0005          & $1.9\times 10^{-5}$  & $-6.18\times 10^{-5}$\\    
                           & (0.0003)      & (0.0004)               & (0.0004)        & (0.0003)             & (0.0003)\\   
      parent\_eduyears     & 0.0013$^{*}$  & 0.0005                 & 0.0007          & 0.0002               & 0.0006\\   
                           & (0.0007)      & (0.0011)               & (0.0010)        & (0.0007)             & (0.0007)\\   
      parent\_reledu       & 0.0018        & -0.0113                & -0.0253         & 0.0002               & 0.0042\\   
                           & (0.0118)      & (0.0182)               & (0.0177)        & (0.0113)             & (0.0117)\\   
      eld\_sex             & 0.0029        & 0.0068                 & -0.0158$^{**}$  & -0.0107$^{**}$       & -0.0018\\   
                           & (0.0048)      & (0.0074)               & (0.0072)        & (0.0046)             & (0.0048)\\   
      eld\_age             & -0.0007$^{*}$ & $-1.18\times 10^{-5}$  & -0.0007         & -0.0004              & $-5.27\times 10^{-5}$\\    
                           & (0.0004)      & (0.0006)               & (0.0006)        & (0.0004)             & (0.0004)\\   
      hh\_asset\_index     & -0.0021       & -0.0007                & 0.0004          & 0.0033$^{*}$         & 0.0018\\   
                           & (0.0020)      & (0.0030)               & (0.0030)        & (0.0019)             & (0.0020)\\   
      hh\_income           & 0.0036$^{**}$ & 0.0062$^{***}$         & 0.0011          & 0.0031$^{**}$        & 0.0011\\   
                           & (0.0014)      & (0.0022)               & (0.0021)        & (0.0014)             & (0.0014)\\   
      int\_trauma\_exp     & 0.0018$^{*}$  & 0.0014                 & -0.0004         & 0.0004               & 0.0011\\   
                           & (0.0010)      & (0.0016)               & (0.0015)        & (0.0010)             & (0.0010)\\   
      \midrule
      \emph{Fit statistics}\\
      Observations         & 667           & 667                    & 667             & 667                  & 667\\  
      R$^2$                & 0.03223       & 0.02420                & 0.04615         & 0.03678              & 0.01432\\  
      Adjusted R$^2$       & 0.01447       & 0.00629                & 0.02865         & 0.01910              & -0.00377\\  
      F-test               & 1.8149        & 1.3515                 & 2.6371          & 2.0807               & 0.79180\\  
      F-test, p-value      & 0.04242       & 0.18482                & 0.00188         & 0.01636              & 0.65928\\  
      \midrule \midrule
      \multicolumn{6}{l}{\emph{IID standard-errors in parentheses}}\\
      \multicolumn{6}{l}{\emph{Signif. Codes: ***: 0.01, **: 0.05, *: 0.1}}\\
   \end{tabular}
    }
\end{table}

\begin{table}[H]
   \caption{Bias tests for Capacity: Budget Low}
   \centering
   \resizebox{\linewidth}{!}{%
   \begin{tabular}{lccccc}
      \tabularnewline \midrule \midrule
      Dependent Variables: & ChatGPT                & Llama213Bchat   & Llama213B      & iQualChatGPTaug & iQual\\  
      Model:               & (1)                    & (2)             & (3)            & (4)             & (5)\\  
      \midrule
      \emph{Variables}\\
      (Intercept)          & 0.0064                 & -0.1465$^{***}$ & -0.0080        & 0.0140          & 0.0073\\   
                           & (0.0145)               & (0.0227)        & (0.0208)       & (0.0177)        & (0.0170)\\   
      data\_roundR3        & -0.0225                & 0.0435$^{**}$   & -0.0352$^{*}$  & -0.0153         & -0.0080\\   
                           & (0.0139)               & (0.0218)        & (0.0200)       & (0.0170)        & (0.0163)\\   
      refugee              & -0.0102                & -0.0141         & -0.0314$^{**}$ & -0.0223$^{**}$  & -0.0230$^{**}$\\   
                           & (0.0091)               & (0.0142)        & (0.0130)       & (0.0111)        & (0.0106)\\   
      num\_child           & 0.0025                 & 0.0023          & -0.0014        & 0.0050$^{*}$    & 0.0059$^{**}$\\   
                           & (0.0023)               & (0.0036)        & (0.0033)       & (0.0028)        & (0.0027)\\   
      hh\_head\_sex        & -0.0144$^{*}$          & -0.0233$^{*}$   & -0.0098        & -0.0055         & -0.0008\\   
                           & (0.0083)               & (0.0129)        & (0.0118)       & (0.0100)        & (0.0096)\\   
      hh\_head\_age        & $-4.17\times 10^{-5}$  & 0.0004          & -0.0009$^{*}$  & -0.0005         & -0.0003\\   
                           & (0.0004)               & (0.0006)        & (0.0005)       & (0.0004)        & (0.0004)\\   
      parent\_eduyears     & -0.0009                & 0.0010          & 0.0007         & -0.0011         & -0.0006\\   
                           & (0.0009)               & (0.0014)        & (0.0013)       & (0.0011)        & (0.0011)\\   
      parent\_reledu       & $6.56\times 10^{-5}$   & -0.0070         & -0.0189        & -0.0134         & -0.0064\\   
                           & (0.0156)               & (0.0244)        & (0.0224)       & (0.0190)        & (0.0183)\\   
      eld\_sex             & 0.0079                 & 0.0261$^{***}$  & 0.0017         & 0.0120          & 0.0083\\   
                           & (0.0064)               & (0.0099)        & (0.0091)       & (0.0078)        & (0.0074)\\   
      eld\_age             & 0.0003                 & -0.0011         & 0.0001         & 0.0002          & $-3.12\times 10^{-5}$\\    
                           & (0.0005)               & (0.0008)        & (0.0007)       & (0.0006)        & (0.0006)\\   
      hh\_asset\_index     & 0.0006                 & -0.0012         & -0.0017        & -0.0016         & -0.0035\\   
                           & (0.0026)               & (0.0041)        & (0.0037)       & (0.0032)        & (0.0031)\\   
      hh\_income           & 0.0006                 & -0.0003         & -0.0002        & 0.0008          & $5.47\times 10^{-5}$\\    
                           & (0.0019)               & (0.0029)        & (0.0027)       & (0.0023)        & (0.0022)\\   
      int\_trauma\_exp     & 0.0015                 & 0.0028          & 0.0016         & 0.0015          & 0.0007\\   
                           & (0.0014)               & (0.0021)        & (0.0020)       & (0.0017)        & (0.0016)\\   
      \midrule
      \emph{Fit statistics}\\
      Observations         & 667                    & 667             & 667            & 667             & 667\\  
      R$^2$                & 0.02488                & 0.03241         & 0.04124        & 0.02117         & 0.01903\\  
      Adjusted R$^2$       & 0.00699                & 0.01465         & 0.02365        & 0.00321         & 0.00103\\  
      F-test               & 1.3905                 & 1.8253          & 2.3443         & 1.1790          & 1.0575\\  
      F-test, p-value      & 0.16518                & 0.04092         & 0.00601        & 0.29415         & 0.39394\\  
      \midrule \midrule
      \multicolumn{6}{l}{\emph{IID standard-errors in parentheses}}\\
      \multicolumn{6}{l}{\emph{Signif. Codes: ***: 0.01, **: 0.05, *: 0.1}}\\
   \end{tabular}
    }
\end{table}

\begin{table}[H]
   \caption{Bias tests for Capacity: Awareness Information High}
   \centering
   \resizebox{\linewidth}{!}{%
   \begin{tabular}{lccccc}
      \tabularnewline \midrule \midrule
      Dependent Variables: & ChatGPT         & Llama213Bchat & Llama213B       & iQualChatGPTaug        & iQual\\  
      Model:               & (1)             & (2)           & (3)             & (4)                    & (5)\\  
      \midrule
      \emph{Variables}\\
      (Intercept)          & -0.2264$^{***}$ & -0.0253       & -0.2251$^{***}$ & -0.0274                & -0.0159\\   
                           & (0.0324)        & (0.0232)      & (0.0318)        & (0.0185)               & (0.0182)\\   
      data\_roundR3        & 0.0049          & -0.0259       & 0.0386          & 0.0093                 & -0.0104\\   
                           & (0.0311)        & (0.0223)      & (0.0305)        & (0.0178)               & (0.0175)\\   
      refugee              & -0.0197         & 0.0072        & -0.0064         & 0.0175                 & 0.0133\\   
                           & (0.0203)        & (0.0146)      & (0.0199)        & (0.0116)               & (0.0114)\\   
      num\_child           & 0.0017          & -0.0035       & 0.0004          & -0.0011                & -0.0010\\   
                           & (0.0051)        & (0.0037)      & (0.0050)        & (0.0029)               & (0.0029)\\   
      hh\_head\_sex        & 0.0345$^{*}$    & 0.0060        & 0.0372$^{**}$   & -0.0074                & -0.0044\\   
                           & (0.0184)        & (0.0132)      & (0.0181)        & (0.0105)               & (0.0103)\\   
      hh\_head\_age        & 0.0006          & 0.0004        & 0.0011          & $-1.72\times 10^{-5}$  & -0.0002\\   
                           & (0.0008)        & (0.0006)      & (0.0008)        & (0.0005)               & (0.0004)\\   
      parent\_eduyears     & -0.0036$^{*}$   & -0.0004       & -0.0032         & 0.0022$^{*}$           & 0.0029$^{**}$\\   
                           & (0.0020)        & (0.0015)      & (0.0020)        & (0.0012)               & (0.0011)\\   
      parent\_reledu       & -0.0068         & -0.0314       & 0.0089          & 0.0089                 & -0.0032\\   
                           & (0.0349)        & (0.0250)      & (0.0342)        & (0.0199)               & (0.0196)\\   
      eld\_sex             & 0.0319$^{**}$   & 0.0013        & 0.0006          & 0.0128                 & 0.0147$^{*}$\\   
                           & (0.0142)        & (0.0102)      & (0.0139)        & (0.0081)               & (0.0080)\\   
      eld\_age             & 0.0001          & -0.0005       & -0.0004         & -0.0004                & $3.4\times 10^{-6}$\\    
                           & (0.0011)        & (0.0008)      & (0.0011)        & (0.0007)               & (0.0006)\\   
      hh\_asset\_index     & -0.0014         & 0.0010        & 0.0018          & 0.0048                 & 0.0041\\   
                           & (0.0058)        & (0.0042)      & (0.0057)        & (0.0033)               & (0.0033)\\   
      hh\_income           & 0.0019          & 0.0011        & 0.0037          & -0.0003                & 0.0004\\   
                           & (0.0042)        & (0.0030)      & (0.0041)        & (0.0024)               & (0.0024)\\   
      int\_trauma\_exp     & 0.0047          & 0.0021        & 0.0032          & 0.0004                 & 0.0011\\   
                           & (0.0031)        & (0.0022)      & (0.0030)        & (0.0017)               & (0.0017)\\   
      \midrule
      \emph{Fit statistics}\\
      Observations         & 667             & 667           & 667             & 667                    & 667\\  
      R$^2$                & 0.02910         & 0.02823       & 0.02511         & 0.01993                & 0.02743\\  
      Adjusted R$^2$       & 0.01128         & 0.01040       & 0.00723         & 0.00194                & 0.00959\\  
      F-test               & 1.6334          & 1.5831        & 1.4040          & 1.1081                 & 1.5372\\  
      F-test, p-value      & 0.07802         & 0.09173       & 0.15880         & 0.35014                & 0.10605\\  
      \midrule \midrule
      \multicolumn{6}{l}{\emph{IID standard-errors in parentheses}}\\
      \multicolumn{6}{l}{\emph{Signif. Codes: ***: 0.01, **: 0.05, *: 0.1}}\\
   \end{tabular}
    }
\end{table}

\begin{table}[H]
   \caption{Bias tests for Capacity: Awareness Information Low}
   \centering
   \resizebox{\linewidth}{!}{%
   \begin{tabular}{lccccc}
      \tabularnewline \midrule \midrule
      Dependent Variables: & ChatGPT              & Llama213Bchat         & Llama213B              & iQualChatGPTaug & iQual\\  
      Model:               & (1)                  & (2)                   & (3)                    & (4)             & (5)\\  
      \midrule
      \emph{Variables}\\
      (Intercept)          & -0.3240$^{***}$      & -0.6286$^{***}$       & -0.4136$^{***}$        & -0.0033         & -0.0087\\   
                           & (0.0300)             & (0.0318)              & (0.0340)               & (0.0076)        & (0.0088)\\   
      data\_roundR3        & 0.0630$^{**}$        & 0.1408$^{***}$        & 0.0198                 & -0.0005         & 0.0025\\   
                           & (0.0287)             & (0.0305)              & (0.0326)               & (0.0073)        & (0.0085)\\   
      refugee              & 0.0177               & 0.0436$^{**}$         & 0.0255                 & 0.0101$^{**}$   & 0.0071\\   
                           & (0.0188)             & (0.0199)              & (0.0213)               & (0.0048)        & (0.0055)\\   
      num\_child           & 0.0028               & 0.0036                & -0.0016                & 0.0003          & -0.0009\\   
                           & (0.0048)             & (0.0050)              & (0.0054)               & (0.0012)        & (0.0014)\\   
      hh\_head\_sex        & -0.0284$^{*}$        & -0.0319$^{*}$         & -0.0647$^{***}$        & -0.0033         & -0.0071\\   
                           & (0.0170)             & (0.0181)              & (0.0193)               & (0.0043)        & (0.0050)\\   
      hh\_head\_age        & -0.0005              & -0.0004               & -0.0007                & -0.0001         & 0.0001\\   
                           & (0.0007)             & (0.0008)              & (0.0008)               & (0.0002)        & (0.0002)\\   
      parent\_eduyears     & 0.0052$^{***}$       & 0.0063$^{***}$        & 0.0038$^{*}$           & -0.0007         & -0.0013$^{**}$\\   
                           & (0.0019)             & (0.0020)              & (0.0021)               & (0.0005)        & (0.0006)\\   
      parent\_reledu       & 0.0506               & 0.0068                & -0.0011                & -0.0082         & -0.0116\\   
                           & (0.0322)             & (0.0342)              & (0.0365)               & (0.0082)        & (0.0095)\\   
      eld\_sex             & -0.0089              & 0.0080                & -0.0116                & 0.0012          & 0.0006\\   
                           & (0.0131)             & (0.0139)              & (0.0149)               & (0.0033)        & (0.0039)\\   
      eld\_age             & -0.0002              & -0.0004               & -0.0004                & 0.0001          & $-3.79\times 10^{-6}$\\    
                           & (0.0011)             & (0.0011)              & (0.0012)               & (0.0003)        & (0.0003)\\   
      hh\_asset\_index     & 0.0053               & 0.0078                & 0.0094                 & 0.0013          & 0.0005\\   
                           & (0.0054)             & (0.0057)              & (0.0061)               & (0.0014)        & (0.0016)\\   
      hh\_income           & 0.0009               & 0.0024                & 0.0025                 & 0.0009          & 0.0012\\   
                           & (0.0039)             & (0.0041)              & (0.0044)               & (0.0010)        & (0.0011)\\   
      int\_trauma\_exp     & -0.0004              & -0.0005               & $-5.23\times 10^{-5}$  & 0.0004          & -0.0002\\   
                           & (0.0028)             & (0.0030)              & (0.0032)               & (0.0007)        & (0.0008)\\   
      \midrule
      \emph{Fit statistics}\\
      Observations         & 667                  & 667                   & 667                    & 667             & 667\\  
      R$^2$                & 0.05864              & 0.15437               & 0.04177                & 0.01835         & 0.01859\\  
      Adjusted R$^2$       & 0.04137              & 0.13885               & 0.02419                & 0.00034         & 0.00058\\  
      F-test               & 3.3950               & 9.9489                & 2.3757                 & 1.0187          & 1.0323\\  
      F-test, p-value      & $7.7\times 10^{-5}$  & $4.8\times 10^{-18}$  & 0.00532                & 0.42946         & 0.41682\\  
      \midrule \midrule
      \multicolumn{6}{l}{\emph{IID standard-errors in parentheses}}\\
      \multicolumn{6}{l}{\emph{Signif. Codes: ***: 0.01, **: 0.05, *: 0.1}}\\
   \end{tabular}
    }
\end{table}

\begin{figure}[H]
    \centering
    \caption{Refugee and Male eldest child coefficients for all annotations}
    \label{fig:coefs_ref_sex_all}
    \includegraphics[width =  \linewidth]{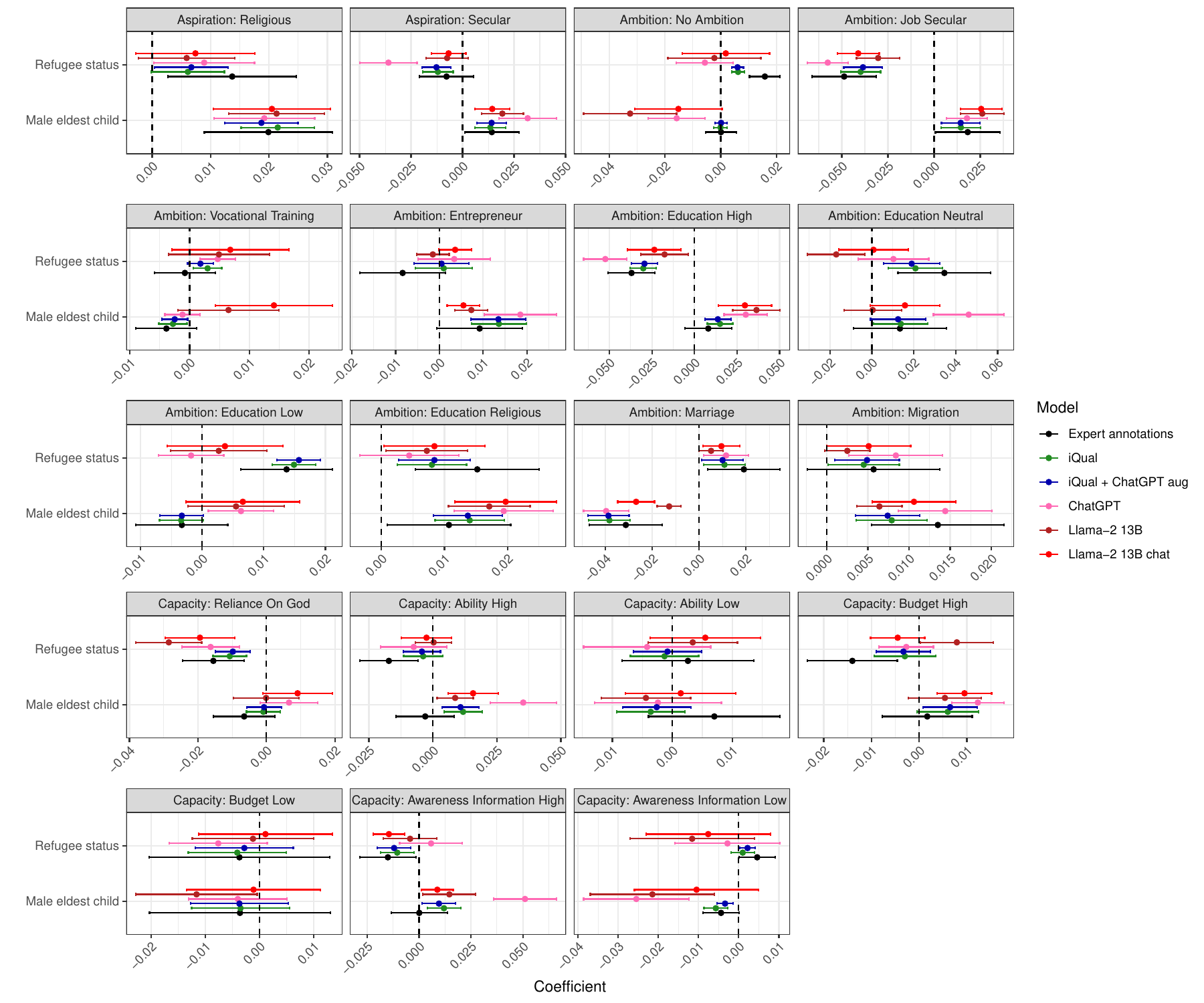}
    \flushleft
    \caption*{\footnotesize \textit{Note}: }
\end{figure}

\end{document}